\documentclass{aifrontiers}
\usepackage{aifrontiers}
\usepackage{booktabs}
\usepackage{tabularx}
\usepackage{makecell}
\usepackage{multirow}
\usepackage{array}
\usepackage{longtable}
\usepackage{comment}
\usepackage{colortbl}
\usepackage{caption}
\usepackage{subcaption}
\usepackage{multicol}
\usepackage{float}
\usepackage{enumitem}
\usepackage{xspace}
\usepackage{soul}
\usepackage{listings}
\usepackage[export]{adjustbox}
\usepackage{algorithm}
\usepackage{algorithmic}
\usepackage{pifont}
\usepackage{marvosym}

\usetikzlibrary{arrows.meta, positioning, calc}

\definecolor{ourcol}{RGB}{225,238,252}
\definecolor{grprule}{RGB}{200,200,200}
\newcommand{\yesmark}{\textcolor{green!60!black}{\ding{51}}}
\newcommand{\nomark}{\textcolor{red!45!black!75}{\ding{55}}}
\newcommand{\pon}{\textcolor{black!85}{\small\Gentsroom}}
\newcommand{\poff}{\textcolor{black!20}{\small\Gentsroom}}
\newcommand{\pn}[1]{\ifcase#1
  \poff\poff\poff\poff\poff\or\pon\poff\poff\poff\poff\or
  \pon\pon\poff\poff\poff\or\pon\pon\pon\poff\poff\or
  \pon\pon\pon\pon\poff\or\pon\pon\pon\pon\pon\fi}
\newcommand{\adh}[2]{\pn{#1}\,\pn{#2}}
\newcommand{\systemhead}[2]{\rotatebox{90}{\tiny #1 #2}}
\newcommand{\systemheadone}[1]{\rotatebox{90}{\tiny #1}}

\usepackage[T1]{fontenc}
\usepackage[utf8]{inputenc}

\newcommand{\tabref}[1]{Table~\ref{#1}}
\newcommand{\figref}[1]{Figure~\ref{#1}}
\newcommand{\secref}[1]{Section~\ref{#1}}

\newlength\savewidth
\newcommand\shline{\noalign{\global\savewidth\arrayrulewidth
  \global\arrayrulewidth 1pt}\hline\noalign{\global\arrayrulewidth\savewidth}}
\newcommand{\tablestyle}[2]{\setlength{\tabcolsep}{#1}\renewcommand{\arraystretch}{#2}\centering\footnotesize}
\newcolumntype{x}[1]{>{\centering\arraybackslash}p{#1pt}}
\newcolumntype{y}[1]{>{\raggedright\arraybackslash}p{#1pt}}

\definecolor{baselinecolor}{gray}{0.93}

\shorttitle{MatrAIx: Simulating the World with 8.3 Billion Persona Agents}

\edef\CasePreInterSf{\sfdefault}
\usepackage[nomath]{inter}
\edef\CaseInterFamily{\sfdefault}
\renewcommand{\sfdefault}{\CasePreInterSf}
\newcommand{\casefont}{\fontfamily{\CaseInterFamily}\selectfont}

\definecolor{casemuted}{HTML}{5B6472}
\definecolor{caseborder}{HTML}{AEBFD1}
\definecolor{casebackground}{HTML}{FBFCFE}
\definecolor{casefillleft}{HTML}{F8F9FB}
\definecolor{casefillright}{HTML}{F3F7FA}

\newtcolorbox{casefeature}[1][]{%
  enhanced,
  breakable,
  colback=casebackground,
  colframe=caseborder,
  boxrule=0.8pt,
  arc=3pt,
  left=8pt, right=8pt, top=8pt, bottom=8pt,
  boxsep=0pt,
  before skip=6pt, after skip=8pt,
  fontupper=\small,
  #1}

\newcommand{\caseopts}[1]{%
  \begin{itemize}[leftmargin=1.05em, itemsep=0.5pt, topsep=2pt, parsep=0pt,
    label={\color{aifrontiersblue}\scriptsize\textbullet}, font=\scriptsize]
  #1
  \end{itemize}\vspace{2.5pt}\noindent\ignorespaces}

\newtcolorbox{caseblock}[1][]{%
  enhanced,
  colback=casebackground,
  colframe=caseborder,
  boxrule=0.6pt,
  arc=3pt,
  left=7pt, right=7pt, top=7pt, bottom=7pt,
  boxsep=0pt,
  before skip=0pt, after skip=0pt,
  #1}

\newcounter{casetopgroup}

\newcommand{\caseheadingdark}[1]{%
  {\scriptsize\bfseries #1}\\[2pt]
  {\color{caseborder}\rule{\linewidth}{0.7pt}}\\[5pt]}

\newcommand{\casestudyrow}[5]{%
  \par\noindent
  \begingroup
  \casefont
  \noindent{\scriptsize\bfseries #1}\hfill
  {\scriptsize\color{casemuted}#2}\par
  \vspace{6pt}%
  \noindent
  \begin{minipage}[t]{0.30\linewidth}\raggedright\scriptsize #3\end{minipage}%
  \hfill
  \begin{minipage}[t]{0.30\linewidth}\raggedright\scriptsize #4\end{minipage}%
  \hfill
  \begin{minipage}[t]{0.30\linewidth}\raggedright\scriptsize #5\end{minipage}%
  \endgroup
  \par\vspace{3pt}}

\newsavebox{\caseboxleft}
\newsavebox{\caseboxright}
\newlength{\casewidth}
\usepackage{bm}

\def\figref#1{figure~\ref{#1}}

\def\secref#1{section~\ref{#1}}

\def\eqref#1{equation~\ref{#1}}

\def\1{\bm{1}}

\DeclareMathAlphabet{\mathsfit}{\encodingdefault}{\sfdefault}{m}{sl}
\SetMathAlphabet{\mathsfit}{bold}{\encodingdefault}{\sfdefault}{bx}{n}

\newcommand{\organizerauthors}{%
Xiaomin Li$^{1}$\thanks{Equal contribution. Correspondence: Xiaomin Li
(\texttt{xiaominli@g.harvard.edu}) and Yuexing Hao
(\texttt{yuexing@mit.edu}).\protect\\Code:
\url{https://github.com/MatrAIx-ai/MatrAIx-Persona-8B}. Project website:
\url{https://matraix.ai}.} \&
Yuexing Hao$^{2}$\footnotemark[2]
}
\newcommand{\advisoryauthors}{%
Paul Pu Liang$^{2}$,
Mitchell Gordon$^{2}$,
Yilun Du$^{1}$,
Marinka Zitnik$^{1,32}$,
James Zou$^{11}$,
Prasanna Tambe$^{24,36}$,
Philip Torr$^{37}$,
Emily Fox$^{11}$,
Asu Ozdaglar$^{2}$,
Dawn Song$^{21}$
}
\newcommand{\contributorauthors}{%
Jianheng Hou$^{3}$,
Jintao Huang$^{4}$,
Qianfeng Wen$^{5}$,
Shirley Huang$^{1,6}$,
Yifan Liu$^{5}$,
Xiaoyi Liu$^{7}$,
Yilan Fan$^{8}$,
Yijun Wang$^{1}$,
Koutian Wu$^{9}$,
Ruoqi Gao$^{11}$,
Muhammad Ahmed Mohsin$^{11}$,
Jing Tang$^{12}$,
Brihi Joshi$^{3}$,
Heming Liu$^{13}$,
Zheyuan Deng$^{7}$,
Zonglin Di$^{10}$,
Sankalp Jajee$^{14}$,
Jiuyao Lu$^{36}$,
Zhiwei Zhang$^{15}$,
Saksham Kapoor$^{16}$,
Ishan Gupta$^{17}$,
Yunhan Zhao$^{18}$,
Chanwoo Park$^{2}$,
Yucheng Lu$^{1,39}$,
Bing Hu$^{19}$,
Weihang Xiao$^{20}$,
Aravind Mohan$^{22}$,
Hanwen Xing$^{3}$,
Runyu Zhang$^{2}$,
Mihir Kulshreshtha$^{20}$,
Yuanda Xu$^{23}$,
Qianyu Zhu$^{2}$,
Dianzhuo Wang$^{1}$,
Yuxin Xiao$^{2}$,
Bowen Jiang$^{24}$,
Yongye Su$^{25}$,
Wenhao Chai$^{23}$,
Zuxin Liu$^{26}$,
Lawrence Yunliang Chen$^{21}$,
Xuandong Zhao$^{21}$,
Ethan Ye$^{26}$,
Shivam Patel$^{26}$,
Jason Xie$^{10}$,
Alex Martin Richmond$^{2}$,
Weixiang Ding$^{26}$,
Emre Okcular$^{27}$,
Diya Mathew$^{13}$,
Ziheng Wang$^{11}$,
Rana M. Shahroz Khan$^{28}$,
Zhejian Peng$^{13}$,
Fang Wu$^{11}$,
Fan Nie$^{11}$,
Xinyang Han$^{21}$,
Yubin Kim$^{2}$,
Jiawei Zhang$^{29}$,
Zhenting Qi$^{1}$,
Huangyuan Su$^{1}$,
Xu Pan$^{1}$,
Abinitha Gourabathina$^{2}$,
Hyewon Jeong$^{2}$,
Hemanth Neelgund Ramesh$^{30}$,
Kumail Alhamoud$^{2}$,
Kimia Hamidieh$^{2}$,
Zidi Xiong$^{1}$,
Samuel Schmidgall$^{31}$,
Pengrui Han$^{2,13}$,
Yepeng Huang$^{1,32}$,
Yongheng Wang$^{2}$,
Bowen Yang$^{33}$,
Alex Gu$^{2}$,
Yuchu Wang$^{34}$,
Akshay Paruchuri$^{11}$,
Brenna Li$^{11}$,
Hejie Cui$^{11}$,
Jiayuan Ding$^{11}$,
Chaosheng Dong$^{35}$,
Jiahao Wang$^{21}$,
Yixuan He$^{38}$,
Chi Wang$^{13}$,
Pamela Bhattacharya$^{19}$,
Tianyi Peng$^{33}$
}
\newcommand{\authoraffiliations}{%
\small
\begin{multicols}{2}
\begin{enumerate}[leftmargin=1.6em,itemsep=0.15em,topsep=0pt]
  \item Harvard University
  \item Massachusetts Institute of Technology
  \item University of Southern California
  \item The Ohio State University
  \item University of Toronto
  \item Harvard Business School
  \item Brown University
  \item Georgia Institute of Technology
  \item University of Texas at Austin
  \item University of California, Santa Cruz
  \item Stanford University
  \item Boston University
  \item University of Illinois Urbana-Champaign
  \item Medical University of South Carolina
  \item Pennsylvania State University
  \item University of Maryland, College Park
  \item University of California, San Diego
  \item University of California, Irvine
  \item University of California, Riverside
  \item Cornell University
  \item University of California, Berkeley
  \item University at Buffalo
  \item Princeton University
  \item University of Pennsylvania
  \item Purdue University
  \item Carnegie Mellon University
  \item University of San Francisco
  \item University of North Carolina at Chapel Hill
  \item University of Wisconsin-Madison
  \item University of Washington
  \item Johns Hopkins University
  \item Harvard Medical School
  \item Columbia University
  \item University of Michigan
  \item University of Pittsburgh
  \item The Wharton School of the University of Pennsylvania
  \item University of Oxford
  \item Arizona State University
  \item New York University
\end{enumerate}
\end{multicols}
}

\title{MatrAIx: Simulating the World with 8.3 Billion \\Persona Agents}

\author{
{\bfseries Organizers}\\[0.2em]
{\normalsize \organizerauthors}\\[0.55em]
{\bfseries Contributors}\\[0.2em]
{\normalsize \contributorauthors}\\[0.55em]
{\bfseries Advisory Committee}\\[0.2em]
{\normalsize \advisoryauthors}
}
\date{}

\begin{document}

\begin{abstract}
Human evaluation of Artificial Intelligence (AI) systems and digital products is costly, slow, and
difficult to scale. Offline evaluations are more scalable but often abstract
away human diversity and interactive behavior. We therefore introduce
\emph{MatrAIx}, a population-scale simulated-user evaluation infrastructure
for testing AI systems and digital products with heterogeneous users. MatrAIx
has three core components: First, Persona~8B contains $8.3$ billion
persona records represented through a schema of $1{,}290$ categorical
dimensions. Records are either sampled from a dependency graph that preserves
correlated attributes or derived from human-authored profiles. We release a
quality-filtered coreset of approximately 1 million personas,
comprising 599,847 human-grounded and 400,000 synthetic records. Second, the
MatrAIx Playground provides four environments in which diverse users evaluate
and interact with digital products: Survey, AI Chatbot, Web, and App. Third,
MatrAIx provides $1{,}010$ application tasks spanning more than 25 domains,
including Commerce, Software, Finance, and Healthcare. We conducted 18,189
evaluation trials across eight representative tasks. Persona agents were
powered by three large language models: Claude Opus 4.8, GPT 5.5, and
Claude Haiku 4.5. The resulting feedback captures how decisions and preferences
vary across persona backgrounds, including hesitation after a price increase,
willingness to continue after an AI assistant fails, and latency tolerance. We
conducted two main validation studies: First, a 400-trial controlled study
evaluated persona adherence across ten behavioral attributes and all four
environments. The declared behavior was expressed or correctly suppressed in
366 trials (91.5\%). Second, human and large language model (LLM) judges evaluated the extraction
quality of human-grounded personas. Overall, MatrAIx provides an end-to-end
infrastructure for evaluating AI systems and digital products with diverse
simulated human users.
\end{abstract}
\maketitle
\setcounter{footnote}{0}
\clearpage

\section{Introduction}
\label{sec:introduction}

Human evaluation remains essential for understanding how AI systems and
digital products perform for real users. However, its time and expense limit
the breadth and frequency of studies during development. Offline benchmarks
offer a scalable and reproducible alternative. However, they typically measure task
outcomes without modeling how diverse users formulate requests, interact with
a system \citep{chang2025chatbench}, and judge its results
\citep{santurkar2023whoseopinions,kirk2024prism}. For
example, a coding-agent benchmark may test whether the trajectory passes all unit tests
\citep{jimenez2024swebench,miserendino2025swelancer,
zan2025multiswebench,zhang2025swebenchlive}.
This establishes functional correctness, but it does not capture user needs
or preferences. A novice may want explanations, small edits, and frequent
confirmation. An expert may instead prefer terse responses, broader
refactoring, and greater autonomy. Some users provide detailed specifications
and inspect every change. Others begin with underspecified goals and expect
the agent to ask clarifying
questions. Such differences shape interaction trajectories, trust in the
result, and willingness to continue after a failure. Offline capability
benchmarks emphasize task completion
\citep{zhou2023webarena,yao2024taubench,xie2024osworld}. They provide less
evidence about how performance and experience vary across users
\citep{chang2025chatbench}. Aggregate scores can also hide problems
encountered by particular user groups.
The same limitation applies to any digital product whose outcome depends on
how users interact with it, not only to AI systems.

Simulated-user evaluation can help bridge the gap between offline benchmarks
and online evaluation. Persona-driven agents can interact with a system
\citep{yoon2024evaluating}, adapt their actions to its responses
\citep{zhou2023sotopia}, and report outcomes from different user perspectives
\citep{dou2025simulatorarena}. This
approach is increasingly practical because AI agents can now
reason and plan \citep{liu2023agentbench,mialon2023gaia}, browse the web
\citep{zhou2023webarena}, write code \citep{trivedi2024appworld}, call tools
\citep{qin2023toolllm}, and operate software \citep{xie2024osworld}. User simulation offers several practical
advantages. First, it reduces time and cost. Human studies may require weeks
for recruitment, scheduling, and data collection, whereas parallel agent
trials can return initial screening results within hours. Second, it supports
controlled repetition. The same task, cohort, and configuration can be rerun
after a system change to compare product versions. Third, it enables
cohort-level analysis by holding the system and task fixed while comparing
user groups. Fourth, it expands population coverage. A simulated-user pool can
contain millions or billions of profiles, while each task runs only the cohort
it needs \citep{ge2024personahub,yang2024oasis,piao2025agentsociety}.
The approach does not require a perfect model of human behavior to be useful.
Even an imperfect \citep{li2025far,zhou2026mind} but diverse simulated
population can expose corner cases, subgroup-specific friction, and failure
modes before deployment \citep{dou2025simulatorarena}.
Putting simulated-user evaluation into practice requires two foundations: a
structured persona population for sampling diverse cohorts
\citep{zhang2018personachat,mazare2018training,ge2024personahub}, and
environments in which those personas can interact with systems
\citep{park2023generativeagents,yang2024oasis,piao2025agentsociety}.

We introduce \emph{MatrAIx}, a population-scale simulated-user evaluation
infrastructure for testing AI systems and digital products with heterogeneous
users. Its first core component, \emph{Persona~8B}, represents human variation
through a shared categorical schema. The schema contains $1{,}290$ dimensions
spanning background, psychology, capability, behavior, and lifestyle.
Persona~8B contains $8.3$ billion records built using two complementary
approaches.
Synthetic records are sampled from a dependency graph that combines
source-informed distributions, cross-attribute correlations, and compatibility
rules. For example, English-proficiency probabilities are adjusted using
primary language and region, while a compatibility rule excludes a persona
whose primary language is English but whose English proficiency is
\texttt{None}. Human-grounded records draw from six sources: Wikipedia
biographies, Amazon Reviews histories, the Stack Overflow
Developer Survey, the General Social Survey (GSS), PRISM
Alignment profiles, and consented MatrAIx Persona Survey responses.
All are mapped into the same schema.
Each populated attribute includes a natural-language description, so the
result reads as a profile rather than a list of categorical values. To protect
privacy, human-grounded records are de-identified by removing direct
identifiers such as names and contact details while retaining only the
extracted attributes and descriptions. Together, these complementary paths
combine principles from synthetic-population modeling
\citep{borysov2019microagents,chapuis2022review} and grounded user profiling
\citep{wang2025know}. To
support research use, we apply contradiction checks, deduplication, and
calibration toward selected real-world demographic distributions. We then
release a coreset of approximately 1 million personas, comprising 599,847
human-grounded and 400,000 synthetic records.\footnote{\label{fn:persona1m}The
public dataset and its card are available at
\url{https://huggingface.co/datasets/MatrAIx2026/MatrAIx_Persona_1M}.}

The second core component is the \textit{MatrAIx Playground}, an interactive
interface for running simulated-user studies. It supports four types of environments:
Survey, AI Chatbot, Web, and App.
\textbf{Type I: Survey} asks persona agents to complete surveys and
questionnaires for concept testing, market prediction, and price-sensitivity
research. For example, one survey could ask how many consumers would still buy
a six-pack of soda after a \$2 price increase. \textbf{Type II: AI Chatbot} places persona
agents in conversations with AI assistants or customer-support chatbots. For
example, a study could test whether users continue a conversation after a
chatbot gives a hallucinated answer and then corrects it. Another could
measure how response latency affects satisfaction and willingness to
continue. \textbf{Type III:
Web} supports both browser automation and computer-use agents (CUAs) that
browse websites. A study could ask shoppers with different needs and budgets
to assess the recommendations and how easily they can find, compare, and
select an option. \textbf{Type IV: App} uses CUAs as simulated users of native
desktop and mobile applications. App tasks run in a Docker-based Linux desktop
sandbox or through a remote macOS desktop or iOS simulator. For example, a
study could test whether users can discover and use a feature, or find and
change privacy and security settings. Across all four types, MatrAIx evaluates
system behavior, feature usefulness, latency, user experience (UX), and privacy and security
controls. It records what each persona agent thinks, says, and does, including
task duration, completion status, and verifier results. These environments
build on advances in task-oriented user simulation
\citep{schatzmann2007agenda}, realistic web interaction
\citep{zhou2023webarena,koh2024visualwebarena}, and multimodal computer use
\citep{trivedi2024appworld,xie2024osworld}.

\begin{figure*}[t]
  \centering
  \includegraphics[width=0.99\linewidth]{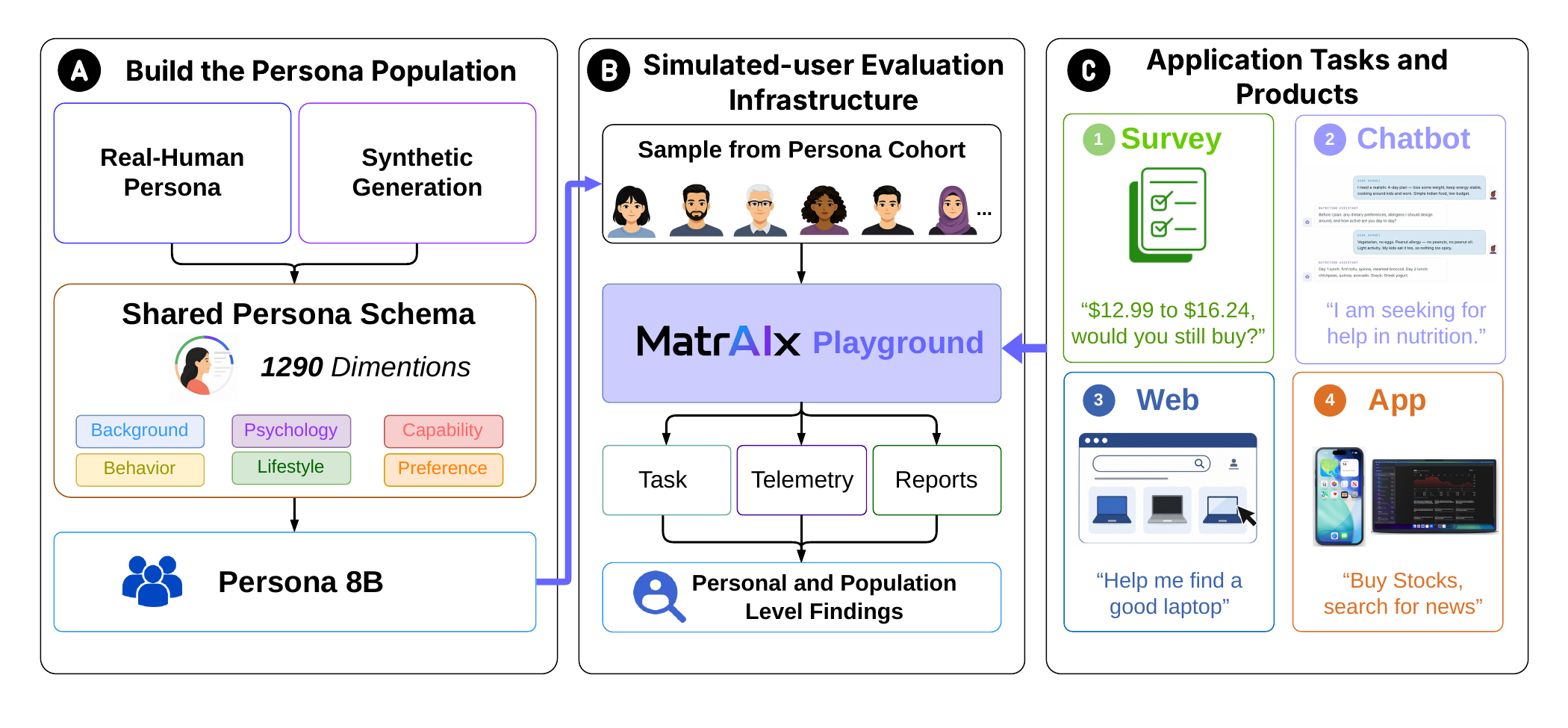}
  \caption{\textbf{Overview of the MatrAIx simulated-user evaluation
  framework.} Evaluators describe the target audience, and MatrAIx retrieves
  matching personas from Persona~8B to form an evaluation cohort. The four
  environments define how these persona agents interact, while MatrAIx
  Applications provides the task specifications they execute. Shared telemetry
  and task-owned verification preserve the evidence needed for subgroup and
  population-level reporting.}
  \label{fig:matraix-framework}
\end{figure*}

The third component is \textit{MatrAIx Applications}, a library of reusable tasks for
evaluating AI systems and digital products.
Each task specifies the target system or product, the persona cohort, and the
scenario and user goal. It also defines the outcome measures and the verifier
used to check each result. For example, a price-sensitivity task could present
the same soda price increase to personas from different income and economic
motivation groups. It would record whether they would still buy the product
and why, then verify that each response includes both purchase intent and a
rationale.
The current MatrAIx Applications release contains $1{,}010$ tasks: 621 Survey,
371 AI Chatbot, 12 Web, and 6 App. The library covers more than 25 domains,
including Commerce, Software, Finance, and Healthcare. Each task defines a
study that can be run with a selected persona cohort. The task count does not
mean that all $1{,}010$ studies have been executed. For this paper, we ran
18,189 evaluation trials across eight representative tasks. For each task, we
report the persona cohort and model configuration. We also report trial-level
verifier results and cohort-level analyses. Figure~\ref{fig:matraix-framework}
shows how the three components form the end-to-end evaluation pipeline.

Together, these components form an operational end-to-end pipeline across all
four environments. The pipeline connects population construction, cohort
retrieval, persona-agent execution, task-owned verification, and
population-level reporting. We evaluate three questions. Does the persona
population preserve plausible distributions and dependencies? Do agents
follow their assigned personas in what they say and do? Do simulated-user
studies reveal consistent differences across systems, models, and user groups?
Our results show that the pipeline operates across all four environments.
Assigned persona attributes are reflected in agent behavior at high rates, and
application studies reveal differences across persona groups and user needs.
These findings expose variation that aggregate scores alone can miss. We
assess population coherence, persona adherence, and evaluation sensitivity
separately because they capture distinct properties. This follows prior work
that treats persona fidelity, behavioral consistency, population
correspondence, and simulator-based system rankings as separate questions
\citep{wang2024incharacter,li2025far,dou2025simulatorarena,wang2025large,
zhou2026mind}. In addition, we conduct two validation studies. First, a
400-trial controlled study across ten behavioral attributes and all four
environments finds that agents express or correctly suppress the assigned
behavior in 366 trials (91.5\%). Second, two LLM judges evaluate all 1,000
extracted personas, and six humans rate a source-matched subset of 100. The
human mean is 4.135/5, and scores from GPT~5.5 and Claude Opus~4.8 are within
one point of the human mean in 79.2\% and 93.8\% of comparisons, respectively.

Our main contributions are:
\begin{itemize}[leftmargin=*]
  \item \textbf{Population-scale 8.3B persona data.}
  Persona~8B contains $8.3$ billion records under a shared
  $1{,}290$-dimensional schema, with a curated coreset of one
  million personas released for research.

  \item \textbf{Complementary persona construction methods.}
  Synthetic records are sampled from source-informed distributions with
  explicit dependencies and compatibility rules. Human-grounded records map
  information from biographies, reviews, surveys, and consented self-reports
  into the same schema.

  \item \textbf{Four evaluation environments.}
  Survey, AI Chatbot, Web, and App support studies ranging from questionnaires
  and conversations to browser and native-application use. Each environment
  records persona-agent interactions, applies task-specific verifiers, and
  aggregates results into cohort-level reports.

  \item \textbf{1K tasks across 25+ domains.}
  MatrAIx Applications contains $1{,}010$ reusable task specifications across
  more than 25 domains, including Commerce, Software, Finance, and Healthcare.

  \item \textbf{Validation of data, behavior, and task results.}
  We investigate population structure, extraction quality, persona adherence, and
  task outcomes.

  \item \textbf{End-to-end demonstrations at scale.}
  We demonstrate the complete pipeline through 18,189 trials across eight
  tasks and all four environments. A separate 400-trial controlled study finds
  that agents express or correctly suppress the assigned behavior in 91.5\%
  of trials.
\end{itemize}

%

\section{Related Work}
\label{sec:related-work}

\paragraph{Persona data and populations.}
Prior work represents personas as short dialogue-conditioning descriptions,
structured profiles, persistent user states, or large synthetic collections.
Some resources are crowd-authored \citep{zhang2018personachat} or generated at
scale \citep{mazare2018training,ge2024personahub}. Others are grounded in
population statistics \citep{nemotronpersonas}, reconstructed from documents
\citep{park2025charactergpt}, or updated from interaction histories
\citep{wang2025know,jiang2025personamem}. Free-text profiles are
flexible conditioning inputs, whereas typed schemas make missing values,
contradictions, population queries, and sampling decisions easier to inspect.
For a collection intended to represent a population, plausible individual
profiles are not enough. The collection must also preserve relevant marginals,
cross-attribute dependencies, and structural constraints. Synthetic-population
research studies these questions in terms of marginal and joint fidelity
\citep{borysov2019microagents,chapuis2022review}. Existing resources generally
focus on constructing persona datasets or conditioning models. They do not
cover the full evaluation process, from sampling a target population to running
interactive tasks and analyzing the outcomes.

\paragraph{User simulation.}
Population- and persona-conditioned language models have been studied as survey
respondents and task-oriented service users
\citep{argyle2023outofonemany,sun2024random,park2024generativeagents1000,
yoon2024evaluating,schatzmann2007agenda,kreyssig2018neural,gur2018user}.
This approach supports controlled comparisons across user profiles and can
reveal differences in preferences and failure modes. However, fluent or
plausible responses do not establish that a simulator behaves like a person.
Models may flatten within-group variation \citep{santurkar2023whoseopinions},
amplify stereotypes \citep{wang2025large}, ignore persona fields
\citep{wang2024incharacter}, or otherwise depart from human behavior
\citep{li2025far}.
Validation must therefore match the intended use. Survey simulations require
calibrated population sampling, while interactive simulations must also capture
behaviors such as disclosure, correction, refusal, and abandonment. Recent
work evaluates whether simulators adhere to assigned personas and preserve
behavioral chains \citep{li2025far}. It also measures human--simulation
agreement \citep{xie2024trustbehavior,chang2025chatbench}, ranking reliability
\citep{dou2025simulatorarena}, and sim-to-real transfer \citep{zhou2026mind}.

\paragraph{Agents and evaluation.}
Agent benchmarks evaluate systems that plan, use tools, and act in interactive
environments \citep{liu2023agentbench,mialon2023gaia,huang2026adkarena,
kim2026teambench}. Their tasks span APIs \citep{qin2023toolllm,yao2024taubench},
websites \citep{yao2022webshop,deng2023mind2web,zhou2023webarena,
koh2024visualwebarena}, applications \citep{trivedi2024appworld}, and operating
systems \citep{xie2024osworld}. Evaluation usually
combines executable task outcomes with rubric-based LLM judges for open-ended
outputs and interactions \citep{liu2023geval,zheng2023judging}. Because these
scores depend on the judge and prompt and can exhibit systematic biases, they
require calibration against human annotations
\citep{wang2024fair,angelopoulos2023prediction}. A separate
line of work uses agents as simulated participants rather than as the systems
under test. Generative Agents \citep{park2023generativeagents},
SOTOPIA \citep{zhou2023sotopia}, Concordia \citep{vezhnevets2023generative},
OASIS \citep{yang2024oasis}, and AgentSociety \citep{piao2025agentsociety}
give agents persistent identities, memories, and relationships to support
social interaction. MatrAIx adopts this participant role for
simulated users and uses them to evaluate a fixed target. Unlike an agent
benchmark, the target need not be an agent: it may be a website, application,
survey, or other digital product. MatrAIx varies the simulated-user population
under a declared sampling design while holding that target fixed.

\section{Persona 8B: A Population-Scale Persona Dataset}
\label{sec:persona8b}

\subsection{Representation and Schema}
\label{sec:persona-schema}

Persona~8B is the population-scale persona dataset that powers the MatrAIx
simulated-user evaluation infrastructure. It is not designed to reconstruct
identifiable people. Instead, it provides a shared schema for describing human
variation, querying records, and sampling evaluation cohorts. Persona~8B
contains $8.3$ billion persona records. A record becomes a \emph{persona agent}
when it is paired with a model. Evaluations then assign sampled persona agents
to an agent interface and task.

The representation is built around $d=1{,}290$ categorical dimensions. Let
\begin{equation}
  \mathcal{D}=\{X_1,\ldots,X_d\}, \qquad d=1{,}290,
\end{equation}
where dimension $X_i$ has a finite value set $\mathcal{X}_i$. A fully
specified synthetic persona is an assignment
\begin{equation}
  x=(x_1,\ldots,x_d)
  \in \mathcal{X}_1\times\cdots\times\mathcal{X}_d.
  \label{eq:persona-assignment}
\end{equation}
Human-grounded records use the same coordinate system but may be partial, with
$x_i=\texttt{null}$ when the available evidence does not support an assignment.
For example, age bracket includes values such as 18--24, 25--34, and 35--44.
English proficiency ranges from \texttt{Native} to \texttt{None}, with
intermediate levels such as \texttt{Fluent (C1-C2)}, \texttt{Intermediate
(B1-B2)}, and \texttt{Basic (A1-A2)}. Risk tolerance ranges from
\texttt{Risk-averse} to \texttt{Risk-seeking}.

The schema groups background, psychology, capability, behavior, and lifestyle
attributes under one typed interface (Table~\ref{tab:persona-schema}). It was
designed to support both population queries, such as selecting by age, region,
language, expertise, or accessibility needs, and model-facing persona
conditioning. Public sources inform both the schema and selected priors. These
sources cover demographics, economics, education, labor, health, values, and
technology use. Examples include UN World Population Prospects \citep{un_wpp}, World Bank
indicators \citep{world_bank_wdi}, ILOSTAT \citep{ilostat}, public surveys
\citep{wvs}, and developer-ecosystem statistics \citep{stackoverflow_survey}.
Table~\ref{tab:persona-schema} summarizes the schema at a high level. The
complete three-layer taxonomy appears in Appendix~\ref{app:schema-taxonomy}
(Figure~\ref{fig:persona-schema-taxonomy}). Appendix~\ref{app:schema-grounding}
maps the schema groups to their grounding sources and roles. The full
dependency graph appears in Appendix~\ref{app:dag-sampling}
(Figure~\ref{fig:full-dag}).

\begin{table*}[t]
  \centering\small
  \setlength{\tabcolsep}{5pt}\renewcommand{\arraystretch}{1.12}
  \begin{tabularx}{\textwidth}{@{}l r X@{\hspace{14pt}}X@{}}
    \toprule
    \textbf{Top-level group} & \textbf{Dims.} &
    \textbf{Representative attributes} & \textbf{Representative grounding} \\
    \midrule
    \textbf{Background} & 238
      & Age, region, language, education, family, career, industry
      & Population statistics, household surveys, education and
        labor taxonomies \\
    \addlinespace[2pt]
    \textbf{Psychology} & 210
      & Personality, values, worldview, motivation, risk
      & Validated instruments, values surveys, schema design
        priors \\
    \addlinespace[2pt]
    \textbf{Capability} & 331
      & Domain expertise, general skills, tools, programming,
        developer context
      & Occupational taxonomies, technology/developer surveys \\
    \addlinespace[2pt]
    \makecell[tl]{\textbf{Behavior and}\\\textbf{Interaction}} & 124
      & Preferences, habits, interaction state, work practices,
        technology adoption
      & Time-use, consumer, workplace, and technology-use evidence \\
    \addlinespace[2pt]
    \textbf{Lifestyle} & 387
      & Interests, media, culture, hobbies, sports, food, health,
        fitness
      & Health statistics, consumption surveys, cultural sources \\
    \midrule
    \textbf{Total} & \textbf{1,290} & & \\
    \bottomrule
  \end{tabularx}
  \caption{\textbf{Persona~8B schema overview.} Dimension counts refer to
  emitted categorical attributes. Sources provide different kinds and strengths
  of grounding; they do not imply direct population estimates for every value.}
  \label{tab:persona-schema}
\end{table*}

\subsection{Synthetic Persona Generation with DAG Sampling}
\label{sec:synthetic-generation}

Matching marginal distributions alone does not produce a coherent population.
Independent sampling can break age--education, region--language,
employment--seniority, and other dependencies, producing implausible profiles despite
accurate aggregate counts. We therefore use a dependency-aware probabilistic
model that combines source-informed correlations with explicit compatibility
constraints. Concretely, each of the 1,290 schema dimensions is a node. An edge
indicates that one dimension is sampled conditionally on another. For example,
a persona's education level is drawn given its age bracket, while English
proficiency is drawn given its primary language and region. We add an edge
when a source directly reports the conditional relationship. We do not infer
edges from a joint distribution that no available dataset provides. A persona
is then built one dimension at a time in an order that visits parents first,
so every draw uses the context on which it depends.
Appendix~\ref{app:dag-sampling} works through the full calculation for one
dimension and describes how the edges were recovered.

Let $G=(\mathcal{D},E)$ be a directed acyclic graph (DAG) over the
persona dimensions, and let $\mathrm{Pa}(i)$ denote the parents of $X_i$. The
proposal distribution factorizes as
\begin{equation}
  p_{\theta}(x)
  =\prod_{i=1}^{d}
  p_{\theta}\!\left(x_i\mid x_{\mathrm{Pa}(i)}\right).
  \label{eq:persona-dag-factorization}
\end{equation}
This factorization makes local dependency assumptions explicit and conditions
each dimension only on its relevant predecessors. For a root dimension, the local conditional probability distribution (CPD) is
a categorical prior $\pi_i(v)$, the population-wide probability of candidate
value $v$. For a non-root dimension, a candidate value
$v\in\mathcal{X}_i$ is scored by combining the prior with parent-dependent
adjustments and compatibility constraints:
\begin{equation}
  p_{\theta}(X_i=v\mid x_{\mathrm{Pa}(i)})
  \propto
  \pi_i(v)\,
  r_i(v;x_{\mathrm{Pa}(i)})\,
  m_i(v;x_{\mathrm{Pa}(i)}).
  \label{eq:persona-local-cpd}
\end{equation}
The source-informed adjustment $r_i$ combines parent-specific likelihood
ratios, while the binary mask $m_i\in\{0,1\}$ applies compatibility rules.
Consider
English proficiency when primary language is English and region is North
America. The prior $\pi_i(v)$ gives the population-wide probability of each
proficiency value before those parent attributes are known. The adjustment
$r_i$ increases the weight of values that are common in this context, such as
\texttt{Native}. It decreases the weight of values that are less common. The
mask $m_i$ handles a different question: whether a
combination is allowed at all. A
persona whose primary language is English cannot have English proficiency
\texttt{None}, so that candidate receives a zero mask and is removed. A less
typical but possible value, such as \texttt{Basic}, remains eligible rather
than being treated as a contradiction. Scores are then normalized over
$\mathcal{X}_i$. Separating dependency adjustment from compatibility filtering
preserves rare but valid profiles while enforcing hard constraints.
Appendix~\ref{app:dag-sampling} provides details on these definitions.

Base priors and local dependencies are derived from the sources summarized in
Table~\ref{tab:persona-schema}; graph construction and review procedures are
detailed in Appendix~\ref{app:dag-sampling}. Synthetic personas are generated
by forward sampling in a topological order
$\tau=(\tau_1,\ldots,\tau_d)$:
\begin{equation}
  x_{\tau_k}
  \sim
  p_{\theta}\!\left(
    X_{\tau_k}\mid x_{\mathrm{Pa}(\tau_k)}
  \right),
  \qquad k=1,\ldots,d.
  \label{eq:topological-sampling}
\end{equation}
Root attributes are drawn from their grounded categorical priors, while each
downstream attribute is drawn from its normalized local CPD. Each evaluation loads only its selected cohort rather than the full
population. Synthetic records still reflect the model's priors, dependencies,
and design choices. We therefore complement them with human-grounded records
from public profiles, coded surveys, and consented self-reports.

\subsection{Real-Persona Extraction and Volunteer Collection}
\label{sec:human-grounded-personas}

To complement synthetic coverage with observed evidence, we map biographies,
behavioral histories, coded surveys, and consented self-reports into the same
1,290-dimensional schema. Human-grounded records include only source-supported
assignments, while unsupported dimensions remain null.

The human-grounded population draws from six sources. \textit{Wikipedia}
provides biographies, while \textit{Amazon Reviews} are grouped by reviewer
to form review histories \citep{hou2024amazon}. The
\textit{Stack Overflow Developer Survey} is mapped through a deterministic
crosswalk that preserves coded responses \citep{stackoverflow_survey}. The
\textit{General Social Survey (GSS)} responses are also mapped directly from
the survey's coded answers into the persona schema \citep{gss_ncdf}.
\textit{PRISM Alignment} provides coded demographics and participant
self-descriptions \citep{kirk2024prism}. The \textit{MatrAIx
Persona Survey} contributes 355 consented self-reports collected through
social-media posts and university email lists. The survey collects no names,
contact details, or account identifiers, and the released records contain no
direct identifiers. We use LLM-based constrained extraction for the free-text
content from Wikipedia, Amazon Reviews, and PRISM Alignment.
Appendix~\ref{app:human-extraction} provides extraction and instrument details,
and Appendix~\ref{app:survey-cohort-figures} reports missingness and cohort
composition.

\subsection{Quality Control and the Public 1M Coreset}
\label{sec:persona-quality-control}

\paragraph{Quality filtering and deduplication.}
Synthetic records are checked for cross-attribute conflicts, while
human-grounded records are checked for unsupported assignments and provenance.
Human-grounded records are deduplicated using exact hashing and MinHash-based
fuzzy detection. Synthetic records are deduplicated using a
set of 14 high-information attributes: records with identical values across
all 14 attributes are treated as duplicates, and only one is retained.
Appendix~\ref{app:postprocessing} gives the full filtering rules, deduplication
procedure, and record counts.

\paragraph{Distribution calibration and the 1M coreset.}
The human-grounded sources are not population-representative. We therefore
select synthetic records so that the combined coreset approximates published
population statistics for age bracket, region, gender identity, and urbanicity
\citep{un_wpp,world_bank_wdi}. Missing fields are not imputed, and
the result is a best-effort match to these four marginal distributions rather than
representative joint coverage over all 1,290 dimensions. The final
deterministic coreset contains 599,847 human-grounded and 400,000 synthetic
records. This split is a release design choice rather than an estimate of a
real-world source ratio. Table~\ref{tab:persona-1m-composition} reports the
exact composition, and Appendix~\ref{app:coreset-calibration} provides the
sampling and calibration procedure.

\begin{table}[t]
  \centering
  \small
  \begin{tabular}{@{}lr@{}}
    \toprule
    \textbf{Source} & \textbf{Released records} \\
    \midrule
    Wikipedia extraction & 323,438 \\
    Amazon Review extraction & 97,915 \\
    Stack Overflow survey extraction & 113,120 \\
    PRISM Alignment & 1,487 \\
    General Social Survey & 63,532 \\
    MatrAIx volunteer survey & 355 \\
    \cmidrule(l){1-2}
    Human-grounded subtotal & \textbf{599,847} \\
    Full-DAG synthetic & \textbf{400,000} \\
    \midrule
    \textbf{Total} & \textbf{999,847} \\
    \bottomrule
  \end{tabular}
  \caption{\textbf{Composition of the public Persona 1M coreset.}
  ``Human-grounded'' identifies the origin of a record, not a guarantee that
  every extracted field is a verified fact. These counts mirror the
  \emph{Composition} table of the dataset card (footnote~\ref{fn:persona1m}),
  which is the authoritative record for the release.}
  \label{tab:persona-1m-composition}
\end{table}

\section{Evaluation Infrastructure}
\label{sec:execution-infrastructure}

\subsection{Simulation Configuration and Execution}

A simulation begins with a population query and application task submitted
through the MatrAIx Playground. The Playground records the eligible persona
pool, sampling procedure, task version, agent, and model in a run manifest. It
then launches one independent trial for each persona in the sampled cohort.
Each trial is represented as
$\tau=\langle\pi,\theta,\alpha,\mu,\sigma\rangle$,
where persona $\pi$ performs task $\theta$ through agent interface $\alpha$
using model $\mu$ and seed $\sigma$. Each trial produces a canonical artifact
bundle $A=\alpha(\pi,\theta;\mu)$ containing its submission and, when
applicable, its trajectory and environment state. A task-owned verifier maps
this bundle to typed findings $V_\theta(A)$. Trials share no state and are therefore
parallel by construction. The manifest retains both the requested population
and realized cohort, while typed findings keep persona fidelity, product
outcomes, and execution failures distinct.

\subsection{Four Evaluation Environments}

The four environments differ in how persona agents interact with the product
being evaluated and what evidence they record:
\begin{enumerate}
  \item \textbf{Type I: Survey.} Each persona agent completes a questionnaire.
  The environment records structured answers and rationales and checks that required
  questions have valid responses.
  \item \textbf{Type II: AI Chatbot.} Each persona agent converses with the
  chatbot being evaluated. The environment records the full conversation, tool
  and service calls, and whether the user's goal was resolved.
  \item \textbf{Type III: Web.} Each persona agent browses a live or task-hosted
  website. The environment records the pages viewed, actions taken, screenshots
  when used, and the agent's final submission.
  \item \textbf{Type IV: App.} Each persona agent operates a Linux, macOS, or
  iOS application with mouse, keyboard, or touch actions. The environment
  records the interaction sequence, final application state, and changes such
  as created files or updated settings.
\end{enumerate}
Implementation and deployment details for each environment are provided in
Appendix~\ref{app:runtime-environments}.

\subsection{Execution, Verification, and Reporting}

Trials can run on the local machine or on remote workers connected over HTTP.
Each worker can execute multiple trials in parallel, and adding workers
increases throughput without changing the task or output format. Only approved,
non-secret configuration fields are sent to remote workers, and model-provider
credentials remain on the worker.

For each trial, MatrAIx stores the persona, task, agent, model, and seed, along
with the interaction trajectory, final environment state, verifier results,
and environment-specific artifacts. Programmatic verifiers check observable
outcomes such as required states, constraints, and side effects. Human or LLM
judges are used when an outcome requires interpretation. Reports aggregate
these results by task, cohort, and subgroup while preserving links to the
underlying trials and evidence. This supports multidimensional evaluation
rather than reducing system performance to a single score
\citep{liang2022helm,xing2026curveshift}.

After reviewing a report, evaluators can change the cohort, scenario, or
verifier and rerun the study while keeping the task version and other settings
fixed. Appendix~\ref{app:runtime-environments} provides details on remote
execution, scaling, security, telemetry, and storage.

\section{Application Tasks}
\label{sec:application-tasks}

\subsection{Task Library}

MatrAIx Applications organizes tasks by evaluation environment and domain.
The environment determines how persona agents interact with a product, while
the task defines the AI system or digital product under test, persona
cohort, scenario, user objective, and outcome measures. The four environments
are Survey, AI Chatbot, Web, and App. The library uses Commerce, Software,
Finance, and Healthcare as anchor domains and covers more than 25 others,
including travel, legal services, insurance, education, entertainment, food,
real estate, and games. The library includes two categories of tasks. Grounded
tasks are based on public instruments, existing AI models or agents, real
products, or live interfaces. Synthetic tasks provide controlled coverage for
recurring study designs, including purchase intent, price sensitivity,
retention, support resolution, and recommendation. Our current release
contains 1,010 unique task specifications: 621 Survey, 371 AI Chatbot, 12 Web,
and 6 App tasks (Table~\ref{tab:task-coverage}).
Appendix~\ref{app:application-task-details} describes the inventory and
distinguishes available, implemented, executed, and reported tasks.

\begin{table}[t]
  \centering
  \small
  \setlength{\tabcolsep}{5pt}
  \begin{tabular}{@{}lrrrrrr@{}}
    \toprule
    & \textbf{Commerce} & \textbf{Software} & \textbf{Finance} &
    \textbf{Healthcare} & \textbf{Other} & \textbf{Total} \\
    \midrule
    Survey     & 202 & 138 & 141 & 139 & 1   & \textbf{621} \\
    AI Chatbot & 3   & 11  & 17  & 29  & 311 & \textbf{371} \\
    Web        & 2   & 2   & 2   & 0   & 6   & \textbf{12} \\
    App        & 0   & 5   & 1   & 0   & 0   & \textbf{6} \\
    \midrule
    \textbf{Total} & \textbf{207} & \textbf{156} & \textbf{161} &
    \textbf{168} & \textbf{318} & \textbf{1,010} \\
    \bottomrule
  \end{tabular}
  \caption{\textbf{Application-task coverage.} Counts are unique
  specifications on the repository's main branch
  together with the batch collections on its synthetic-task branches;
  individually contributed tasks still under review on open pull requests are
  not counted. ``Other'' aggregates more than 25 additional domains.}
  \label{tab:task-coverage}
\end{table}

\subsection{Application Task Specification}

Each task specifies four elements. It names the evaluation target, such as an
AI model, agent, chatbot, website, or native application. It also defines the
persona cohort, the user-facing scenario and objective, and the response,
behavior, outcome, or final state that counts as evidence. Keeping these fields
in the task contract makes tasks portable without placing product credentials
or scoring rules in the persona instructions. For example, a price-sensitivity
task can present the same price increase to shoppers with different economic
motivations and record their purchase intent and rationale. The stored cohort
query and seed support repetition with the same sample or comparison with a
different population while holding the product and scenario fixed.

Each task also specifies the evidence to retain and how it will be evaluated.
Survey tasks produce structured answers and rationales. AI Chatbot tasks record
the conversation and post-run feedback. Web tasks preserve the pages viewed,
actions taken, considered options, and final submission. App tasks can also
record exported files, permission changes, final application state, and
cross-application effects. These artifacts support measures such as the share
of simulated users who purchase or continue using a product, mean satisfaction
ratings, task-completion rates, and average completion times. For example, a
task can test whether more than 60\% of simulated users would continue using an AI chatbot or whether its mean satisfaction rating exceeds 4 out of 5. The interaction
trajectories show how personas reached, revised, or abandoned their decisions.

A task-specific verifier converts these artifacts into structured findings.
Programmatic checks evaluate directly observable outcomes, while human or LLM
judges assess interpretive properties using recorded prompts and rubrics.
Reports summarize completion, outcome distributions, uncertainty, and subgroup
differences while retaining links to the underlying traces and evidence.
Product outcomes, simulated-user behavior, and persona fidelity remain
separate, so task success is not treated as evidence of human validity.

\subsection{Case Study: Meal-Planning Chatbot}
\label{sec:task-example-mealplanning}

The meal-planning chatbot task shows how a task specification becomes an
executable study. In every trial, a persona agent asks the same GPT-4o mini assistant for a
multi-day meal plan tailored to its dietary needs and preferences. The
assistant remains fixed across all trials, while the model powering the persona
agent varies among Opus~4.8, GPT~5.5, and Haiku~4.5. This design allows us to
compare how different persona agents use the same persona information
when interacting with the same assistant. Figure~\ref{fig:task-case-mealplanning}
summarizes the study design and examines whether persona attributes are
associated with the agent's stated likelihood of following the resulting meal
plan. Appendix~\ref{app:val-case-mealplanning} reports the corresponding
conversation-path analysis. Together, the example shows how one task
specification connects the system under test, persona cohort, interaction
protocol, outcome measures, and supporting evidence.

\begin{casefeature}
  \casestudyrow{(A)\;\;AI Chatbot task\,\textperiodcentered\,Meal planning \& nutrition}%
    {}{%
  \caseheadingdark{Task and protocol}
  \caseopts{%
    \item A persona agent powered by Opus 4.8, GPT 5.5, or Haiku 4.5
      interacts with the same GPT-4o mini assistant
    \item Goal: obtain a multi-day meal plan tailored to the persona, including
      one substitution and one dining-out recommendation
    \item 7.1 turns per conversation on average (SD $=$ 1.8)}}{%
  \caseheadingdark{Outcome measures}
  \caseopts{%
    \item Eight self-report questions after the conversation; two examples:
    \item ``How likely are you to follow the meal plan from 1 to 10?''
      (1--10, primary outcome)
    \item ``How well did the meal plan satisfy your core dietary
      constraints and health goals?'': yes / partially / no}}{%
  \caseheadingdark{Persona cohort}
  \caseopts{%
    \item 1{,}000 personas per persona-agent model: 82 schema dimensions plus 42
      task-specific diet and cuisine attributes
    \item Ages 25--54; six diet-relevant attributes are stratified
    \item The rest are spread evenly across their levels, not weighted
      to any real population
    \item Each persona LLM draws its own 1{,}000 personas from the same
      sampling recipe}}
  \noindent{\casefont\scriptsize\bfseries (B)\;\;Persona deviation in the downstream outcome}\par
  \vspace{2pt}%
  \begin{center}
    \includegraphics[width=0.85\linewidth]{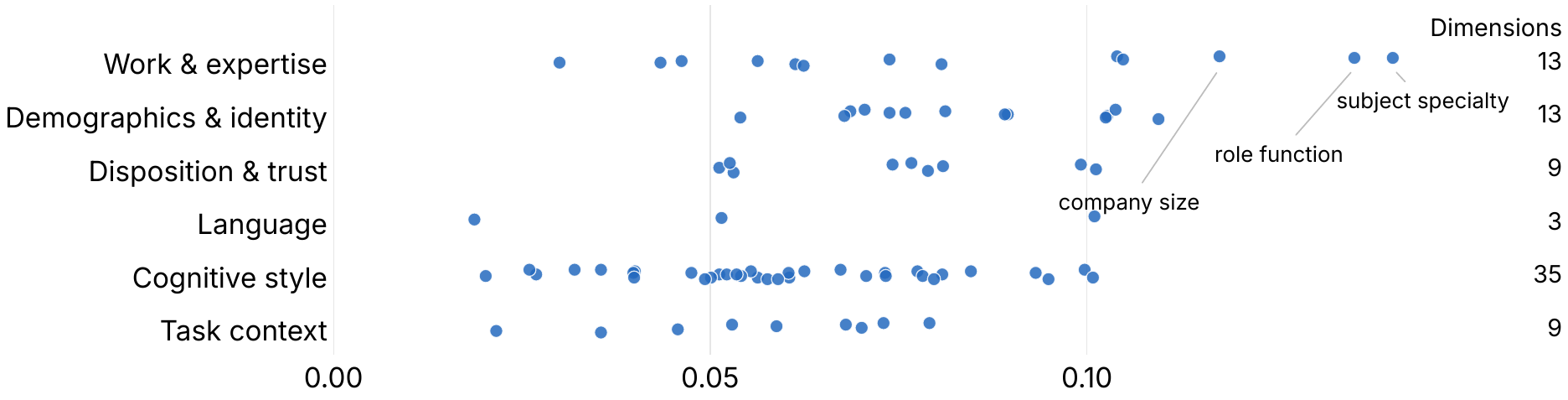}
  \end{center}
  \captionof{figure}{\textbf{Meal-planning chatbot task and persona-outcome analysis.}
  \textbf{(A)} The task protocol, outcome measures, and persona cohort. The
  cohort supports comparisons across its sampled subgroups and persona-agent
  models, but it is not weighted to represent a real population.
  \textbf{(B)} Association between each of 82 persona dimensions and whether
  the persona agent selected the modal adherence response in the GPT~5.5
  condition, measured using Cram\'er's $V$. The three largest associations are
  labeled. Empty nesters reported a higher likelihood of following the plan
  than career changers (66\% versus 46\%), but no association remained
  significant after Benjamini--Hochberg correction (best $q=0.51$). These
  differences are therefore descriptive rather than confirmed persona effects.
  Appendix~\ref{app:val-case-mealplanning} reports how the same subgroups
  differed in their conversation paths.}
  \label{fig:task-case-mealplanning}
\end{casefeature}

\section{Validation of Simulated-User Evaluation}
\label{sec:validation}

We evaluate four properties of the infrastructure: whether tasks execute as
specified, whether task-relevant persona effects can be recovered across
models, whether assigned persona attributes affect observable behavior, and
whether human-grounded records are supported by their source material.
Together, these studies test end-to-end execution, application-level
consistency, behavioral adherence, and source grounding.

\paragraph{Execution coverage.}
We ran eight representative tasks, two from each environment type, with OpenAI GPT~5.5,
Claude Opus~4.8, and Claude Haiku~4.5. Survey, AI Chatbot, and Web tasks used
approximately 1,000 personas per model. The two App tasks used 24 and 20 because
native interaction is substantially more expensive. All analyses report actual
completion denominators and apply Benjamini--Hochberg correction
\citep{benjamini1995fdr} across the
eight declared primary outcomes. Appendix~\ref{app:experimental-details}
provides complete run accounting, statistical procedures, and
artifact-integrity exceptions.

\paragraph{Application-level consistency.}
\label{sec:val-persona-effects}

Persona effects are clearest when the assigned attribute is directly relevant
to the task. In the OpenBB task, trust level separates subgroups under all three
persona-agent models
(Cram\'er's $V=0.228$--$0.363$, all $q<10^{-8}$), and all three models order
the four trust groups identically. This result shows that MatrAIx can recover a
consistent subgroup pattern across models when the task provides a clear
behavioral channel for the persona attribute. We report the persona-agent model
as part of every evaluation configuration, consistent with prior work showing
model-dependent differences in reflected opinions
\citep{santurkar2023whoseopinions}, user-simulation behavior
\citep{yoon2024evaluating}, and persona fidelity \citep{wang2024incharacter}.
Appendix~\ref{app:additional-results} reports the complete model- and task-level
comparisons.

\begin{figure}[!ht]
  \centering
  \begin{subfigure}[t]{0.61\textwidth}
    \centering
    \includegraphics[width=\linewidth]{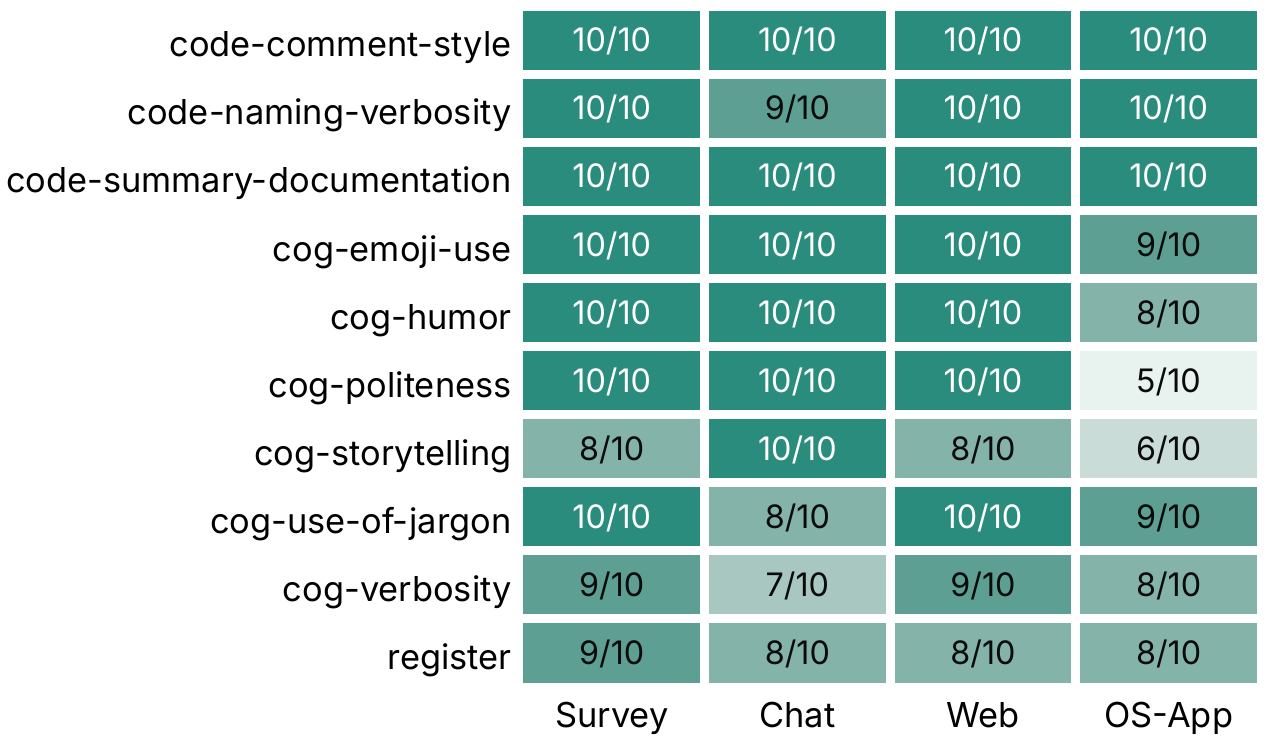}
    \caption{Attribute-level adherence.}
  \end{subfigure}\hfill
  \begin{subfigure}[t]{0.35\textwidth}
    \centering
    \includegraphics[width=\linewidth]{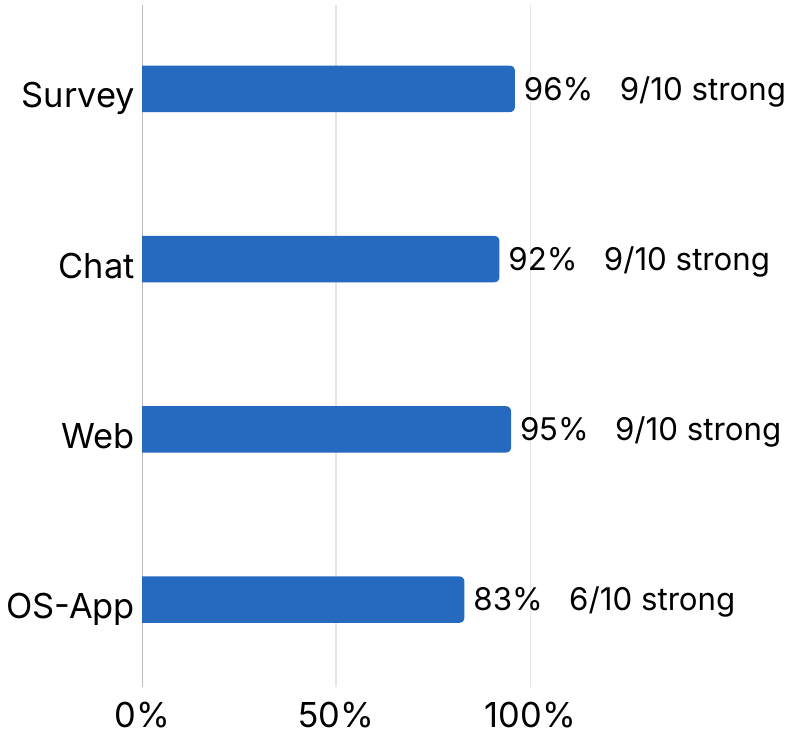}
    \caption{Environment summary.}
  \end{subfigure}
  \caption{\textbf{Controlled behavioral adherence across four environments.}
  \textbf{(a)}~Each cell pools five personas declaring one pole of an attribute
  and five declaring the opposite pole; $10/10$ means that all five positive
  personas expressed the target and all five negative personas expressed the
  opposite value. \textbf{(b)}~Overall successful-trial share by environment and
  the number of strong attributes, defined as at least four of five successes
  in both arms. An LLM judge evaluates the recorded trajectory or artifact.
  Overall, 366 of 400 trials succeed.}
  \label{fig:val-behavioral-adherence}
\end{figure}

\paragraph{Controlled behavioral adherence.}
Ten behavioral attributes are each evaluated in Survey, Chatbot, Web, and App,
with five personas declaring one pole and five declaring the opposite, for 400
trials. The declared behavior is expressed or correctly suppressed in 366
trials (91.5\%). Of the 40 attribute-by-environment cells, 33 achieve at least
four of five successes in both arms. Survey, AI Chatbot, and Web each meet this
threshold for 9 of 10 attributes, compared with 6 of 10 for App. Figure~\ref{fig:val-behavioral-adherence} summarizes the results;
Appendices~\ref{app:persona-adherence} and \ref{app:adherence-cells} provide the
design, judge evidence, and per-cell breakdown.

\paragraph{Extraction quality.}
We also test whether human-grounded persona records are supported by their
source material. Two LLM judges score 1,000 extracted personas, with 89.1\% of
paired metric scores within one point. On a
source-matched subset of 100 personas rated by six humans, the mean quality
score is 4.135/5. GPT~5.5 and Claude Opus 4.8 are within one point of the human mean in
79.2\% and 93.8\% of comparisons, respectively. These results support
source-grounded extraction quality. Appendix~\ref{app:extraction-quality} gives
the rubric, sampling procedure, and full results.

\paragraph{Interpretation.}
The persona-agent model is part of the evaluation configuration and should be
reported with each result. Important findings should be checked with more than
one model before they guide product decisions. The present studies validate
execution, persona adherence, and source-grounded extraction quality;
Appendix~\ref{app:responsible-use} discusses the remaining validation scope.

\section{Conclusion}
\label{sec:conclusion}

AI systems and digital products serve people with different goals,
capabilities, preferences, and constraints, yet many evaluations represent
only a generic user. MatrAIx provides an end-to-end infrastructure built from
three components: Persona~8B, with approximately 8.3 billion records under a
shared schema; the
MatrAIx Playground, which runs persona agents in Survey, AI Chatbot, Web, and
App environments; and MatrAIx Applications, a library of 1,010 versioned tasks
across more than 25 domains. In this paper, we completed 18,189 trials over eight representative application tasks using
three persona-agent models. In the controlled adherence study, assigned
behaviors were expressed or correctly suppressed in 366 of 400 trials
(91.5\%). The extraction study also found strong
support for human-grounded persona records: six human raters assigned a mean
quality score of 4.135 out of 5 on the source-matched subset. These results support using MatrAIx for pre-deployment screening, subgroup
analysis, stress testing, and comparison across product versions. Important
findings should be checked across persona-agent models and traced back to the
underlying interactions. Human studies remain necessary before applying
conclusions to real populations or consequential decisions. Before release,
simulate diverse users, then validate against reality.
\subsubsection*{Acknowledgments}

We thank our research funders and partners for making this work possible:
OpenAI, Anthropic, Microsoft Azure, Amazon Web Services, Meta, the MIT CSAIL
Alliance, and the MIT Sandbox Innovation Fund. Their support
through research funding, compute resources, model access, and program
mentorship enabled the construction of Persona~8B and the population-scale
evaluation runs reported here. The views and conclusions expressed in this
paper are those of the authors and do not necessarily reflect the positions of
the supporting organizations. We also thank Sky Ng, Ziwei Liu, Zhengyang Shan,
Jicheng Wang, Chiffon Nguyen, Qin Yang, Ruoxi Wu, Dr. Zhixu Tao, Yan Jiang,
and Cheng Cheng for helpful discussions and support for the project.

\bibliographystyle{plainnat}
\bibliography{references}

\appendix

\clearpage
\section*{Appendix Table of Contents}

\newcommand{\appoutline}[1]{%
  \noindent\textbf{\ref{#1}\quad \nameref{#1}}\dotfill\textbf{\pageref{#1}}\par\medskip}
\newcommand{\appoutlinesub}[1]{%
  \noindent\hspace{2em}\ref{#1}\quad \nameref{#1}\dotfill\pageref{#1}\par}

\appoutline{app:authors}
\appoutlinesub{app:author-groups}
\appoutlinesub{app:author-affiliations}
\appoutlinesub{app:author-contributions}
\appoutlinesub{app:competing-interests}
\appoutline{app:persona-schema}
\appoutlinesub{app:schema-taxonomy}
\appoutlinesub{app:schema-category-index}
\appoutlinesub{app:schema-grounding}
\appoutline{app:population-model}
\appoutlinesub{app:dag-sampling}
\appoutlinesub{app:postprocessing}
\appoutlinesub{app:coreset-calibration}
\appoutline{app:human-extraction}
\appoutlinesub{app:extraction-engine}
\appoutlinesub{app:source-preprocessing}
\appoutlinesub{app:extraction-modes}
\appoutlinesub{app:extraction-validation}
\appoutlinesub{app:survey-cohort-figures}
\appoutline{app:runtime-environments}
\appoutlinesub{app:trial-formalization}
\appoutlinesub{app:launch-surfaces}
\appoutlinesub{app:execution-lifecycle}
\appoutlinesub{app:scaling-reproducibility}
\appoutlinesub{app:telemetry-schema}
\appoutlinesub{app:playground}
\appoutline{app:application-task-details}
\appoutlinesub{app:task-library-snapshot}
\appoutlinesub{app:task-contract}
\appoutlinesub{app:task-cohort-contract}
\appoutlinesub{app:task-artifacts}
\appoutlinesub{app:task-verification}
\appoutlinesub{app:val-case-candyland}
\appoutlinesub{app:val-case-mealplanning}
\appoutlinesub{app:case-news}
\appoutline{app:experimental-details}
\appoutlinesub{app:validation-runs}
\appoutlinesub{app:validation-statistics}
\appoutline{app:additional-results}
\appoutlinesub{app:validation-outcomes}
\appoutlinesub{app:val-consistency}
\appoutlinesub{app:val-fidelity}
\appoutline{app:persona-adherence}
\appoutlinesub{app:adherence-design}
\appoutlinesub{app:adherence-cells}
\appoutlinesub{app:adherence-evidence}
\appoutline{app:extraction-quality}
\appoutlinesub{app:extraction-rubric}
\appoutlinesub{app:extraction-llm}
\appoutlinesub{app:extraction-human}
\appoutline{app:system-comparison}
\appoutline{app:future-directions}
\appoutline{app:responsible-use}

\clearpage
\section{Authors and Affiliations}
\label{app:authors}

\subsection{Author Groups}
\label{app:author-groups}

\paragraph{Organizers.}
\organizerauthors.

\paragraph{Contributors.}
\contributorauthors.

\paragraph{Advisory Committee.}
\advisoryauthors.

\noindent $^{\dagger}$Equal contribution.\\
\noindent Correspondence: Xiaomin Li
(\texttt{xiaominli@g.harvard.edu}); Yuexing Hao
(\texttt{yuexing@mit.edu}).

\subsection{Affiliations}
\label{app:author-affiliations}

\authoraffiliations

%
%
%
%

\subsection{Author Contributions}
\label{app:author-contributions}

X.L. and Y.H. contributed equally across all stages of the work. They conceived
and coordinated the project; designed the persona schema, population
construction methods, and simulated-user evaluation infrastructure; and
participated directly in implementation and execution across synthetic persona
generation and post-processing, human-grounded persona extraction and data
curation, the four Playground environments, application-task and verifier
development, evaluation runs, extraction-quality and persona-adherence
validation, statistical analysis, and figure preparation. They jointly
interpreted the results and wrote and revised the manuscript.

The contributor group built and operated the system reported here: the
dependency-aware generation pipeline and its post-processing, the
human-persona extraction engine and its source-specific adapters, the four
evaluation environments and the shared trial, telemetry, and reporting
interfaces, the application-task library and its verifiers, and the
statistical analyses and figures. Contributors also ran the evaluation jobs
behind the reported results and carried out the extraction-quality and
persona-adherence validation studies.

The advisory committee advised on research design, evaluation methodology,
measurement validity, and responsible release, and reviewed the manuscript.

All authors reviewed the manuscript and approved the submitted version.

\subsection{Competing Interests}
\label{app:competing-interests}

This work was supported by research funding, compute resources, model access,
and program mentorship from OpenAI, Anthropic, Microsoft Azure, Amazon Web
Services, Meta, the MIT CSAIL Alliance, and the MIT Sandbox Innovation Fund.

OpenAI and Anthropic, both listed above as supporters of this work, produce
agent models evaluated in \secref{sec:validation} and
Appendix~\ref{app:experimental-details}. We disclose this relationship for
transparency. All model conditions follow the reported evaluation procedures;
the per-trial records underlying each aggregate are retained; and the analysis
code and task definitions are released to support independent verification.

The supporting organizations had no role in the design of the study, the
selection of tasks or models, the analysis or interpretation of results, or the
decision to submit this work for publication. The views and conclusions
expressed here are those of the authors and do not necessarily reflect the
positions of the supporting organizations.

\section{Persona Schema}
\label{app:persona-schema}


\subsection{Three-Layer Persona Taxonomy}
\label{app:schema-taxonomy}

\begin{figure}[p]
  \centering
  \includegraphics[width=0.96\textwidth,height=0.88\textheight,keepaspectratio]{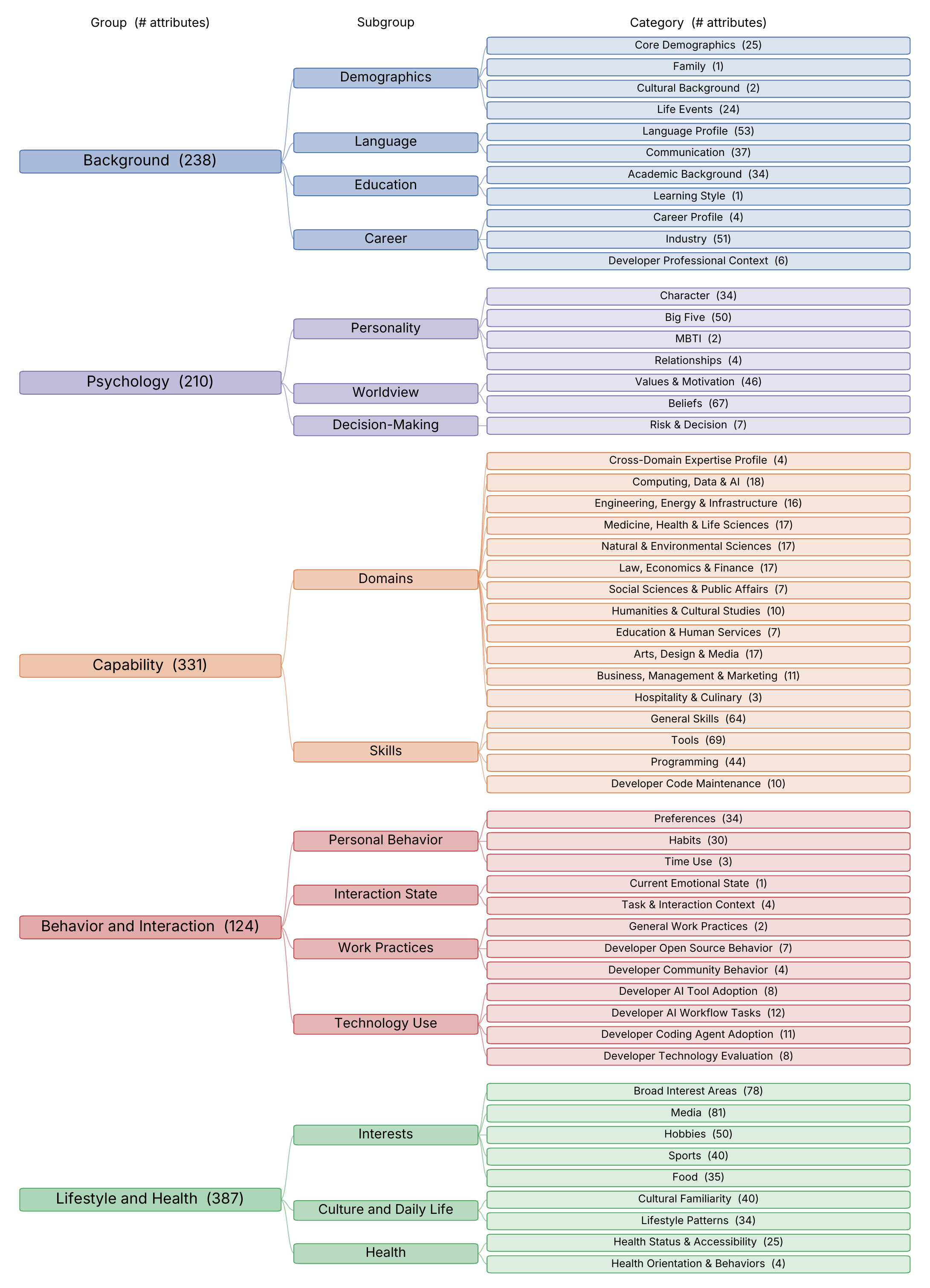}
  \caption{\textbf{Complete three-layer taxonomy of the Persona~8B schema.}
  The left column gives the five top-level groups and their attribute totals;
  the middle column gives 16 subgroups; and the right column gives all 55
  conceptual categories with counts. The hierarchy covers exactly 1,290
  attributes and excludes latent or helper variables used only by the
  generation graph.}
  \label{fig:persona-schema-taxonomy}
\end{figure}

The 1,290 attributes are organized at two related levels. The conceptual
taxonomy assigns every attribute to one of five groups, 16 subgroups, and 55
fine-grained categories. The storage and questionnaire schema uses 43
categories. Most conceptual categories correspond directly to one schema
category; selected broad schema categories, such as domain expertise and
cultural interests, are split into more informative conceptual categories.
Every attribute has exactly one assignment at each taxonomy layer, and all
counts in Figure~\ref{fig:persona-schema-taxonomy} sum to 1,290.

\paragraph{Provenance of the taxonomy.}
The two levels have different origins. Selected dimensions and value sets
follow measurement conventions from the source families catalogued in
Table~\ref{tab:schema-grounding-sources}. Census and labor instruments inform
categories for education, occupation, and household structure; public-opinion
surveys inform response scales for values and attitudes; and
developer-ecosystem surveys inform the technology-use vocabulary. These
sources play different roles and do not directly determine every schema field
or value.

The three-layer organization above them is ours. The five groups, 16
subgroups, 55 conceptual categories, and separate 43-category storage schema
are design choices made for this dataset. The conceptual layers allow a cohort
to be selected by an idea (\emph{accessibility needs}, \emph{domain expertise})
rather than by enumerating fields. The storage layer keeps the questionnaire
and record format stable when the conceptual grouping is revised. We are not aware of
a published persona taxonomy at this granularity that we could have adopted, and
we do not claim the grouping is the only defensible one. It is auditable
instead: every attribute has exactly one assignment at each layer, the counts in
Figure~\ref{fig:persona-schema-taxonomy} sum to 1,290, and the complete
attribute-level mapping ships with the release so that a reader who prefers a
different organization can regroup from the same fields.

\subsection{Complete Schema Category Index}
\label{app:schema-category-index}

Table~\ref{tab:persona-schema-category-index} indexes all 43 schema categories
used by the dimension catalog and questionnaire interface. The complete
attribute-level mapping, including stable index, field identifier, label,
schema category, and all three conceptual taxonomy assignments, is supplied as
\path{supp_material/persona_taxonomy_mapping.csv}. The 1,290-item questionnaire
supplement additionally records every field's prompt and allowed values.

\begingroup
\scriptsize
\setlength{\tabcolsep}{2.6pt}
\renewcommand{\arraystretch}{1.04}
\begin{longtable}{@{}l l l c p{0.335\textwidth}@{}}
  \caption{\textbf{Complete index of the 43 schema categories.} Counts sum to
  1,290 attributes. Examples are labels from the authoritative dimension
  catalog; the complete attribute-level mapping is supplied as
  \protect\path{supp_material/persona_taxonomy_mapping.csv}.}
  \label{tab:persona-schema-category-index}\\
  \toprule
  \textbf{Group} & \textbf{Subgroup} & \textbf{Schema category} &
  \textbf{Count} & \textbf{Representative attributes} \\
  \midrule
  \endfirsthead
  \toprule
  \textbf{Group} & \textbf{Subgroup} & \textbf{Schema category} &
  \textbf{Count} & \textbf{Representative attributes} \\
  \midrule
  \endhead
  \midrule
  \multicolumn{5}{r}{\textit{Continued on next page}} \\
  \endfoot
  \bottomrule
  \endlastfoot

  Background & Demographics & Demographic: Core & 25 & Age bracket; region; gender identity \\
  Background & Demographics & Demographic: Cultural & 2 & Cultural background; attitude toward immigration \\
  Background & Demographics & Demographic: Family & 1 & Household size \\
  Background & Demographics & Demographic: Life Events & 24 & Life stage; major life events; childhood environment \\
  Background & Language & Linguistic: Language & 53 & Primary language; English proficiency; multilingualism \\
  Background & Language & Linguistic: Communication & 37 & Expected tone; verbosity; communication preferences \\
  Background & Education & Learning: Academic & 34 & Highest education; academic field; institution tier \\
  Background & Education & Learning: Style & 1 & Learning style \\
  Background & Career & Professional: Career & 4 & Research output; seniority; years of experience \\
  Background & Career & Professional: Industry & 51 & Company size; role function; industry \\
  Background & Career & Developer: Professional Context & 6 & Professional status; role archetype; contribution context \\
  \addlinespace

  Psychology & Personality & Personality: Character & 34 & Domain stance; dominant trait; curiosity \\
  Psychology & Personality & Personality: Big Five & 50 & Imagination; artistic interest; emotionality \\
  Psychology & Personality & Personality: MBTI & 2 & Neurotype; Myers-Briggs type \\
  Psychology & Personality & Personality: Relationships & 4 & Attachment anxiety; attachment avoidance; interpersonal agency \\
  Psychology & Worldview & Values \& Motivation & 46 & Core value; religiosity; economic motivation \\
  Psychology & Worldview & Worldview: Beliefs & 67 & Political leaning; trust level; safety sensitivity \\
  Psychology & Decision-Making & Risk \& Decision & 7 & Risk tolerance; decision style; need for closure \\
  \addlinespace

  Capability & Domains & Expertise: Domains & 144 & Domain; subject specialty; technology savviness \\
  Capability & Skills & Expertise: Skills & 64 & Writing; copywriting; editing \\
  Capability & Skills & Skills: Tools & 69 & Excel; Google Sheets; Python \\
  Capability & Skills & Skills: Programming & 44 & Comment style; summary documentation; naming verbosity \\
  Capability & Skills & Developer: Code Maintenance & 10 & Complexity tolerance; modularity preference; type-system orientation \\
  \addlinespace

  Behavior and Interaction & Personal Behavior & Behavior: Preferences & 34 & Modality preference; accessibility needs; media diet \\
  Behavior and Interaction & Personal Behavior & Behavior: Habits & 30 & Journaling; meditation; use of to-do lists \\
  Behavior and Interaction & Personal Behavior & Behavior: Time & 3 & Time pressure; sleep schedule; micromanagement aversion \\
  Behavior and Interaction & Interaction State & State: Emotional & 5 & Emotional state; intent; query complexity \\
  Behavior and Interaction & Work Practices & Behavior: Work & 2 & Work schedule; office versus remote work \\
  Behavior and Interaction & Work Practices & Developer: Open Source Behavior & 7 & Open-source activity; GitHub contribution mode; pull-request style \\
  Behavior and Interaction & Work Practices & Developer: Community Behavior & 4 & Stack Overflow use; participation style; help-seeking preference \\
  Behavior and Interaction & Technology Use & Developer: AI Adoption & 8 & Coding-AI use frequency; sentiment; output trust \\
  Behavior and Interaction & Technology Use & Developer: AI Workflow Tasks & 12 & AI fit for code generation; debugging; testing \\
  Behavior and Interaction & Technology Use & Developer: Agent Adoption & 11 & Agent use frequency; autonomy preference; workflow impact \\
  Behavior and Interaction & Technology Use & Developer: Technology Evaluation & 8 & AI capability; API completeness; reliability and latency \\
  \addlinespace

  Lifestyle and Health & Interests & Interests: Topics & 78 & Politics; sports; travel \\
  Lifestyle and Health & Interests & Interests: Media & 81 & Pop music; rock music; hip-hop \\
  Lifestyle and Health & Interests & Interests: Hobbies & 50 & Knitting; crocheting; pottery \\
  Lifestyle and Health & Interests & Interests: Sports & 40 & Soccer; basketball; American football \\
  Lifestyle and Health & Interests & Interests: Food & 35 & Italian; French; Spanish cuisine \\
  Lifestyle and Health & Culture and Daily Life & Interests: Culture & 74 & Familiarity with national and regional cultures \\
  Lifestyle and Health & Health & Health: Physical & 25 & General health; chronic condition; mobility \\
  Lifestyle and Health & Health & Health: Fitness & 2 & Interest in fitness; exercise frequency \\
  Lifestyle and Health & Health & Health: Lifestyle & 2 & Diet type; alcohol use \\
\end{longtable}
\endgroup

\subsection{Detailed Schema-to-Source Grounding Map}
\label{app:schema-grounding}

Table~\ref{tab:schema-grounding-sources} expands the high-level schema summary
in Table~\ref{tab:persona-schema}. The sources play different roles. Some
define categories or measurement conventions, some provide population priors,
some support selected conditional dependencies, and others serve primarily as
validation references. Listing a source does not imply that every dimension in
the corresponding group is directly estimated from it.

\begingroup
\scriptsize
\setlength{\tabcolsep}{3pt}
\renewcommand{\arraystretch}{0.96}
\begin{longtable}{@{}p{0.13\textwidth} p{0.22\textwidth} p{0.20\textwidth} p{0.37\textwidth}@{}}
  \caption{\textbf{Detailed schema-to-source grounding map.} The mapping
  records source families and their roles in schema design, prior estimation,
  dependency construction, compatibility rules, or downstream validation.}
  \label{tab:schema-grounding-sources}\\
  \toprule
  \textbf{Schema group} & \textbf{Facets} & \textbf{Grounding roles} &
  \textbf{Sources} \\
  \midrule
  \endfirsthead
  \toprule
  \textbf{Schema group} & \textbf{Facets} & \textbf{Grounding roles} &
  \textbf{Sources} \\
  \midrule
  \endhead
  \midrule
  \multicolumn{4}{r}{\textit{Continued on next page}} \\
  \endfoot
  \bottomrule
  \endlastfoot

  Background
  & Demographics (52); language (90); education (35); career (61)
  & Category definitions; population priors; household, language, education,
    and labor dependencies
  & UN World Population Prospects and Population Data
    \citep{un_wpp,un_population_division}; World Bank WDI and WorldPop
    \citep{world_bank_wdi,worldpop}; Eurostat, ACS PUMS, and IPUMS
    \citep{eurostat,acs_pums,ipums}; DHS, UNICEF MICS, and OECD Family
    Database \citep{dhs_program,unicef_mics,oecd_family_database}; Pew and
    World Values Survey \citep{pew_research_center,wvs}; UNESCO UIS, World
    Bank Education Statistics, OECD INES, and OECD PISA
    \citep{unesco_uis,world_bank_education_statistics,oecd_ines,oecd_pisa};
    ILOSTAT, BLS OEWS, and O*NET \citep{ilostat,bls_oews,onet}; Stack Overflow
    Survey, GitHub Octoverse, and JetBrains Developer Ecosystem
    \citep{stackoverflow_survey,github_octoverse,jetbrains_developer_ecosystem}.
    \\
  \addlinespace

  Psychology
  & Personality (90); worldview (113); decision-making (7)
  & Instrument and value-set design; selected prevalence estimates; validation
  & IPIP and MIDUS \citep{ipip,midus}; Pew and World Values Survey
    \citep{pew_research_center,wvs}; GSS, European Social Survey, ISSP, and
    Gallup World Poll \citep{gss,european_social_survey,issp,gallup_world_poll};
    Afrobarometer, Arab Barometer, Asian Barometer, Latinobarometro, and
    Eurobarometer
    \citep{afrobarometer,arab_barometer,asian_barometer,latinobarometro,eurobarometer};
    ARDA \citep{arda}. \\
  \addlinespace

  Capability
  & Domain expertise (144); general skills (64); tools (69); programming (44);
    developer context (10)
  & Occupational and skill taxonomies; technology access and adoption;
    developer-tool prevalence
  & ITU Statistics, World Bank WDI, DataReportal, and Pew Internet
    \citep{itu_statistics,world_bank_wdi,datareportal,pew_internet}; Stack
    Overflow Survey, GitHub Octoverse, and JetBrains Developer Ecosystem
    \citep{stackoverflow_survey,github_octoverse,jetbrains_developer_ecosystem};
    O*NET \citep{onet}. \\
  \addlinespace

  Behavior and Interaction
  & Personal behavior (67); interaction state (5); work practices (13);
    technology use (39)
  & Time-use and consumer priors; workplace behavior; technology and AI adoption
  & American Time Use Survey, Consumer Expenditure Surveys, and OECD Time Use
    \citep{atus,consumer_expenditure_surveys,oecd_time_use_database}; ITU,
    DataReportal, and Pew Internet \citep{itu_statistics,datareportal,pew_internet};
    Stack Overflow Survey, GitHub Octoverse, JetBrains Developer Ecosystem, and
    O*NET \citep{stackoverflow_survey,github_octoverse,jetbrains_developer_ecosystem,onet}.
    \\
  \addlinespace

  Lifestyle
  & Interests (358); physical health (25); fitness (2); health lifestyle (2)
  & Health and disability priors; consumption and time use; cultural and
    interest category design
  & WHO GHO and IHME GBD \citep{who_gho,ihme_gbd}; ACS PUMS, NHIS, BRFSS,
    DHS, and UNICEF MICS
    \citep{acs_pums,cdc_nhis,cdc_brfss,dhs_program,unicef_mics}; ATUS, CEX,
    and OECD Time Use \citep{atus,consumer_expenditure_surveys,oecd_time_use_database};
    FAOSTAT and UNESCO Culture Statistics \citep{faostat,unesco_culture_statistics};
    Eurobarometer, Pew, DataReportal, WVS, and Gallup World Poll
    \citep{eurobarometer,pew_research_center,pew_internet,datareportal,wvs,gallup_world_poll}.
    \\
\end{longtable}
  \endgroup
\section{Population Model and Post-Processing}
\label{app:population-model}

%
%

\subsection{DAG Construction and Sampling Details}
\label{app:dag-sampling}


\begin{figure}[!ht]
  \centering
  \includegraphics[width=\textwidth]{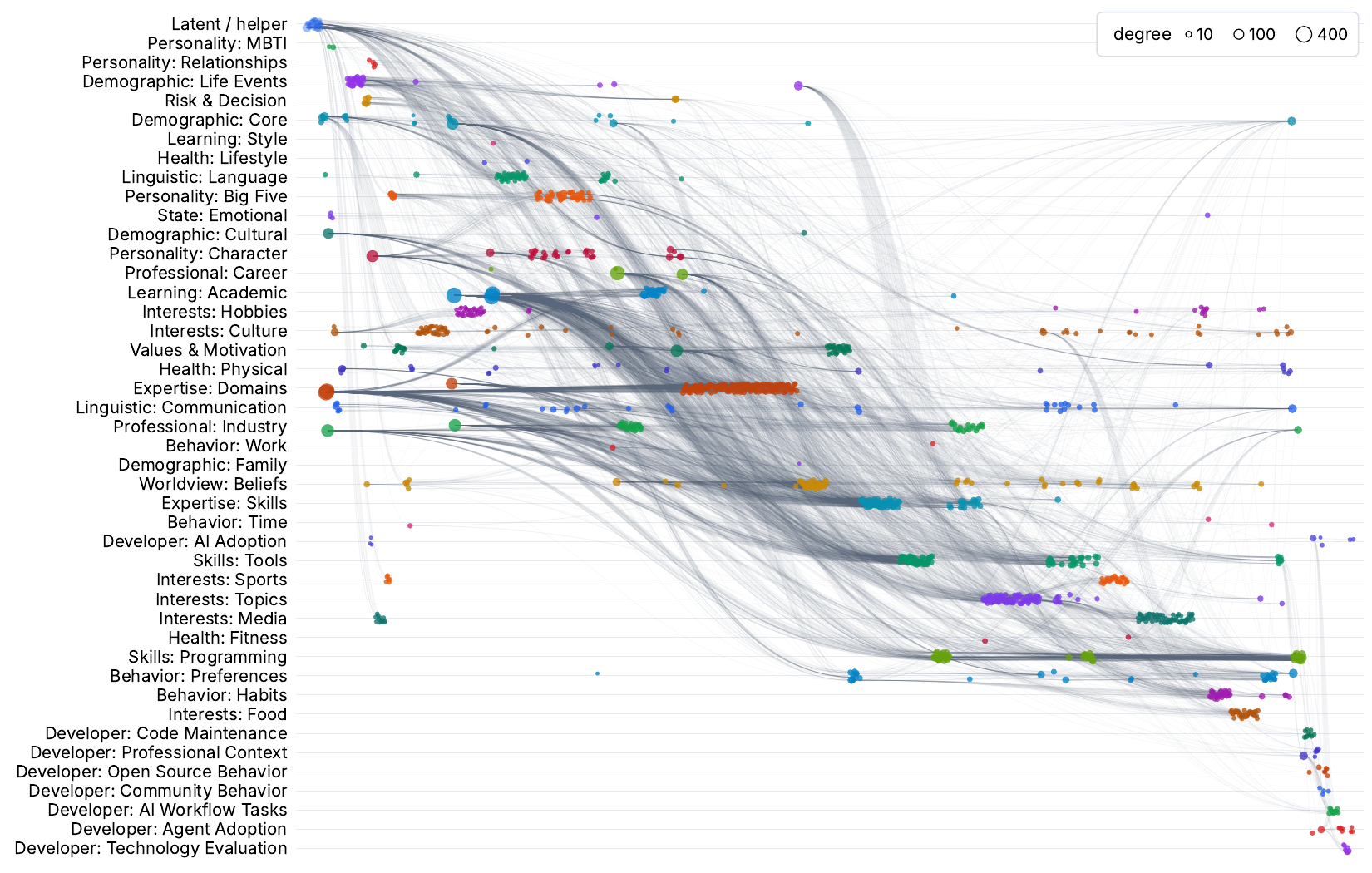}
  \caption{\textbf{The full persona DAG.} All 1,308 graph nodes and 6,999
  directed edges. Nodes are placed left to right by the topological order used
  in Equation~\ref{eq:topological-sampling} and grouped vertically into one
  lane per schema category, with lanes ordered by their mean topological
  position; marker area scales with a node's directed degree. Edges are drawn
  as translucent curves, so darker bands mark the dependency bundles that
  connect demographic and educational roots to downstream expertise, interest,
  and behavior dimensions.}
  \label{fig:full-dag}
\end{figure}

For each non-root dimension, the adjustment term in
Equation~\ref{eq:persona-local-cpd} aggregates source-informed dependency
factors:
\begin{equation}
  r_i(v;x_{\mathrm{Pa}(i)})
  =
  \prod_{a\in\mathcal{A}_i}
  \left(
    \frac{
      q_{a,i}(v\mid x_{S_a})+\epsilon
    }{
      \pi_i(v)+\epsilon
    }
  \right)^{\lambda_{a,i}},
  \qquad S_a\subseteq\mathrm{Pa}(i).
  \label{eq:persona-adjustment}
\end{equation}
Here $\mathcal{A}_i$ is the set of dependency factors for $X_i$,
$S_a$ is the parent subset used by factor $a$, and
$q_{a,i}(\cdot\mid x_{S_a})$ is the corresponding conditional distribution.
The ratio upweights values that become more likely in the parent context and
downweights those that become less likely. The exponent $\lambda_{a,i}$
controls factor strength, and $\epsilon>0$ provides smoothing.

The compatibility term combines local hard constraints:
\begin{equation}
  m_i(v;x_{\mathrm{Pa}(i)})
  =
  \prod_{b\in\mathcal{B}_i}
  \mu_{b,i}(v;x_{T_b}),
  \qquad T_b\subseteq\mathrm{Pa}(i),
  \label{eq:persona-compatibility}
\end{equation}
where $\mathcal{B}_i$ is the rule set for $X_i$ and each
$\mu_{b,i}(v;x_{T_b})\in\{0,1\}$ is a binary mask multiplier. Zero excludes an
invalid assignment and one leaves an admissible assignment unchanged.
Statistical rarity is represented only by $r_i$, so a rare but admissible
combination is not penalized again by the mask. Representative rules exclude
adult work histories for young children, reconcile primary language with
proficiency, and align accessibility states with relevant health attributes.

Graph construction begins with the demographic, socioeconomic, educational,
health, behavioral, and technology priors summarized in
Table~\ref{tab:persona-schema} and Appendix~\ref{app:schema-grounding}. Directed
edges and local CPDs identify the parent context for each child dimension.
Compatibility rules remain separate from empirical dependency factors so that
statistical association and logical validity can be audited independently.
Source review, schema review, and LLM-assisted coverage review identify missing
dependencies and candidate conflicts. The graph is checked for cycles before
its topological order is produced.

Forward sampling applies Equation~\ref{eq:topological-sampling} to each persona.
Repeating the procedure yields
\begin{equation}
  \mathcal{P}_N=\{x^{(1)},\ldots,x^{(N)}\},
  \qquad x^{(n)}\sim p_{\theta}(x).
  \label{eq:synthetic-population}
\end{equation}
Deterministic seeds make each shard reproducible. Independent shards and
append-only outputs permit parallel generation at population scale, while
downstream evaluation jobs instantiate only sampled cohorts.

A node is one schema dimension together with its finite value set, and an edge
records that a source supports conditioning one dimension on another. Of the
1,308 nodes in Figure~\ref{fig:full-dag}, 1,290 are exactly the attributes a
persona record contains. The remaining 18 are latent root factors such as
\texttt{latent\_digital\_engagement} and \texttt{latent\_financial\_security}:
each has a small ordinal value set and no parents, so it is sampled before
everything else, and its outgoing edges fan out to the observable dimensions it
coordinates. A family of attributes that should move together, for example the
dimensions that all reflect financial security, then inherits its correlation
from one shared cause rather than from many pairwise edges. Latent factors are
never emitted, so a persona record still contains exactly the 1,290 schema
attributes. Consider \texttt{english\_proficiency}, whose value set here is
\{\texttt{None}, \texttt{Basic}, \texttt{Fluent}, \texttt{Native}\}, with parents
\texttt{primary\_language} and \texttt{region}. Two of the terms in
Equation~\ref{eq:persona-local-cpd} attach to those parents. A dependency factor
$q_{a,i}$ says how the proficiency distribution shifts once the parent is known,
and it is estimated from a source. A compatibility rule $\mu_{b,i}$ says which
combinations are not admissible at all, and it is asserted rather than
estimated: a persona whose primary language is English cannot also have no
English proficiency.

\paragraph{A worked example.}
Table~\ref{tab:dag-worked-example} runs one child dimension through
Equation~\ref{eq:persona-local-cpd} for the parent context
$\texttt{primary\_language}=\texttt{English}$,
$\texttt{region}=\texttt{North America}$, with both factor exponents
$\lambda_{a,i}=1$ and $\epsilon$ small enough to ignore. \emph{The values are
illustrative and are chosen to show the arithmetic; they are not the deployed
parameters.} Reading across: the prior $\pi_i$ is the population-wide
distribution; each factor contributes the ratio $q_{a,i}/\pi_i$, which exceeds
one where the parent context makes a value more likely; the mask zeroes the one
combination a rule forbids; and the product is renormalized over the surviving
values.

\begin{table}[!ht]
  \centering
  \small
  \begin{tabular}{@{}lrrrrrrr@{}}
    \toprule
    $v$ & $\pi_i(v)$ & $q_{\text{lang}}$ & $q_{\text{region}}$ &
    $r_i(v)$ & $m_i(v)$ & $\pi_i r_i m_i$ & $p_\theta(v\mid x_{\mathrm{Pa}(i)})$ \\
    \midrule
    \texttt{None}   & 0.30 & 0.02 & 0.10 & 0.022 & 0 & 0.000 & 0.000 \\
    \texttt{Basic}  & 0.30 & 0.08 & 0.20 & 0.178 & 1 & 0.053 & 0.028 \\
    \texttt{Fluent} & 0.25 & 0.30 & 0.35 & 1.680 & 1 & 0.420 & 0.224 \\
    \texttt{Native} & 0.15 & 0.60 & 0.35 & 9.333 & 1 & 1.400 & 0.747 \\
    \midrule
    \textbf{sum} & 1.00 & 1.00 & 1.00 & & & 1.873 & 1.000 \\
    \bottomrule
  \end{tabular}
  \caption{\textbf{One child dimension through the local CPD (illustrative).}
  \texttt{english\_proficiency} conditioned on
  $\texttt{primary\_language}=\texttt{English}$ and
  $\texttt{region}=\texttt{North America}$. The prior alone would give
  \texttt{Native} a 15\% share; the two dependency factors raise it to 75\%,
  and the compatibility mask removes \texttt{None} outright rather than merely
  making it unlikely. Numbers illustrate Equation~\ref{eq:persona-local-cpd}
  and are not the deployed parameter values.}
  \label{tab:dag-worked-example}
\end{table}

Two things in the example generalize. First, a single factor with
$\lambda=1$ would reproduce that factor's conditional exactly, since
$\pi_i\cdot(q_{a,i}/\pi_i)=q_{a,i}$; the machinery earns its keep when several
factors condition on \emph{different} parent subsets $S_a$, as here, and no
single source supplies the joint. Second, separating $r_i$ from $m_i$ is what
lets a rare-but-valid profile survive. A low-probability combination is
downweighted by the factors and can still be sampled, whereas a combination that
is not admissible is removed by the mask and can never be, so statistical
rarity and logical invalidity never get confused with each other.

\paragraph{How the edges are recovered.}
Edges are not learned from a fitted joint over 1,290 dimensions, which no
available source would identify. They are added where a source reports a
conditional. Construction proceeds in three passes over the catalogue in
Table~\ref{tab:schema-grounding-sources}. First, a source review adds an edge
wherever a dataset reports one attribute broken down by another and records the
supporting table. Second, a schema review checks that each added parent is
meaningful for the child's value set rather than merely correlated in the
source population. Third, an LLM-assisted coverage review proposes pairs that
a human pass may have missed; each candidate is then accepted or rejected by
hand. Compatibility rules are collected separately in the same passes so that
an association and a prohibition are never entered as the same object. The
result is checked for cycles, and the topological order used by
Equation~\ref{eq:topological-sampling} is produced from the acyclic graph.

\subsection{Detailed Persona Post-Processing Pipeline}
\label{app:postprocessing}

Post-processing is non-destructive: each source shard produces a rejection
bitmap and a report with source, rule, and row-count provenance. The quality
filter evaluates 21 hard contradiction rules over packed synthetic codes and
populated human fields. Across $10{,}002{,}288{,}277$ original records, it
rejected $239{,}310$. Missing or unsupported human fields do not trigger a
conflict.

Human deduplication canonicalizes populated \texttt{field=value} tokens. Exact
128-bit hashes merge identical records; near-duplicates use 64-permutation
MinHash signatures, eight bands of eight rows, and an estimated Jaccard
threshold of $0.95$. Synthetic records instead use a 14-field projection chosen
by graph-prior entropy across schema categories. Records sharing a projection
signature form a bucket, and deterministic priorities select one survivor per
bucket. A final deterministic cutoff sets the desired synthetic corpus size.

\begin{table}[h]
  \tablestyle{4pt}{1.08}
  \begin{tabular}{@{}lrr@{}}
    \toprule
    \textbf{Stage} & \textbf{Rejected} & \textbf{Remaining} \\
    \midrule
    Original corpus & -- & 10,002,288,277 \\
    Contradiction filter & 239,310 & 10,002,048,967 \\
    Human exact/MinHash deduplication & 41,597 & 2,222,496 human \\
    Synthetic projection deduplication & 252,936,392 & 9,746,848,482 synthetic \\
    Synthetic deterministic cutoff & 1,349,070,978 & 8,397,777,504 synthetic \\
    \midrule
    \textbf{Audited baseline} & & \textbf{8,400,000,000} \\
    \bottomrule
  \end{tabular}
  \caption{Detailed post-processing accounting for the audited baseline. Human
  and synthetic remaining counts have different scopes until the final row.}
  \label{tab:persona-postprocess-accounting}
\end{table}

Accepted records are materialized into a unified Arrow/Parquet schema. The
1,290 attributes occupy a 645-byte vector with two four-bit codes per byte; a
162-byte null bitmap preserves missing values, and sparse overrides preserve
legacy values outside the current codebook. Descriptions, grounding,
confidence, assignment types, and source metadata remain attached where
available. Every output file is checked against the unified schema and listed
in a SHA-256 manifest. The accepted snapshot contains
$8{,}399{,}989{,}719$ rows, $10{,}281$ below the intended materialized target
because one incomplete Wikipedia conversion task was excluded.

\subsection{Coreset Candidate Selection and Calibration}
\label{app:coreset-calibration}

All retained records from Amazon, Stack Overflow, PRISM, GSS, and the volunteer
survey enter the human component; Wikipedia is calibrated to $323{,}438$ rows.
For the synthetic component, seed \texttt{20260720} selects 40 of 100 shards
and one Parquet file and row group from each, yielding $2{,}187{,}354$
candidates before selecting $400{,}000$.

Let $h_{dv}$ be the human count for value $v$ of dimension $d$, $H_d$ the
number of human records where $d$ is known, and $p_{dv}$ the target share. The
desired synthetic residual is
\begin{equation}
  r_{dv}=p_{dv}(H_d+400{,}000)-h_{dv}.
  \label{eq:coreset-residual}
\end{equation}
Missing fields are not imputed to satisfy a target, and infeasible residuals
are retained in the release audit.

Calibration iteratively updates a positive weight $w_i$ shared across the four
target dimensions. Fixed-size inclusion probabilities are
\begin{equation}
  \pi_i^{\mathrm{inc}}=1-e^{-t w_i},
  \qquad \sum_i \pi_i^{\mathrm{inc}}=n,
\end{equation}
where $t$ is solved for sample size $n$. Deterministic sampling without
replacement assigns
\begin{equation}
  q_i=\frac{-\log U_i}{w_i},
\end{equation}
with $U_i$ derived from the seed and stable row identifier, and selects the
$n$ smallest priorities. This yields an exact-size, order-independent sample.

The release consists of ten 100K-row Zstandard-compressed Parquet files.
\texttt{manifest.json} records source counts, file sizes, and hashes;
\texttt{audit.json} records targets, achieved shares, known and missing counts,
infeasible residuals, and synthetic candidate provenance; and
\texttt{RESULTS.md} summarizes the completed build.
\section{Human-Grounded Persona Construction}
\label{app:human-extraction}

\subsection{Extraction Engine and Output Contract}
\label{app:extraction-engine}

The current free-text extraction engine uses two parallel signals. A regex
matcher finds literal aliases and values, while regex and embedding retrievers
jointly propose semantically relevant dimensions to an LLM judge. The judge
sees only the retrieved dimensions, their questions, and closed allowed-value
sets; it may assign a value or abstain. Regex and judge outputs are retained
separately and merged by dimension, with disagreements preserved for audit.
Literal matches receive confidence $0.7$; judged assignments retain their
model confidence and quoted evidence.

Production source pipelines normalize extracted attributes to records of the
form
\begin{equation}
  \mathcal{E}(s)
  =\{(i,\hat{x}_i,c_i,e_i,a_i,d_i)\}_{i=1}^{1,290},
\end{equation}
where $\hat{x}_i$ is a schema value or null, $c_i$ is confidence, $e_i$ is
evidence, $a_i$ records assignment provenance, and $d_i$ is a field-level
description. Model-extracted fields distinguish direct evidence, structured
claims, summary inferences, and unsupported fields.

\subsection{Source-Specific Preprocessing}
\label{app:source-preprocessing}

One Wikipedia source row defines one persona and retains its global index,
Wikidata identifier, title, URL, text hash, and source-row provenance. The
source database contains $2{,}125{,}897$ profiles. Amazon instead defines one
persona per reviewer: reviews are sorted chronologically and rendered with
category, product, rating, verified-purchase status, title, and text. The
production cohort contains $100{,}000$ reviewers selected for sufficiently
long and repeated histories \citep{hou2024amazon}.

The production batch extractor partitions the 1,290 dimensions by semantic
category into 53 chunks of at most 50 dimensions. Wikipedia and Amazon use
Qwen3.6-35B-A3B through vLLM; PRISM uses Qwen3-235B-A22B through an
OpenAI-compatible endpoint. Wikipedia is divided into 200 contiguous index
shards and Amazon into 256 deterministic user buckets. Outputs are append-only,
and completed identifiers are skipped on restart, making runs resumable and
idempotent.

\subsection{Observed, Hybrid, and Direct Modes}
\label{app:extraction-modes}

The General Social Survey uses a deterministic crosswalk from 18 coded source
variables; all other dimensions remain null. Afrobarometer uses a comparable
rule-based crosswalk, and ConvAI2 uses conservative phrase matching. PRISM is
hybrid: nine coded demographic fields are mapped exactly and override the LLM,
while self-description and stated AI preferences support extraction of
additional fields. Direct volunteer submissions preserve self-reported values
and unanswered fields without inference.

The complete instrument is reproduced as supplementary material. That document,
\path{supp_material/matraix_persona_survey_instrument.pdf}, lists all 1,290
items grouped by their 43 interface categories and records, for each item,
the schema index, the field identifier used in the exported record, the prompt
text shown to the participant, the full closed set of selectable options, and
the neutral default applied when a participant skips the surrounding category.
The document is generated directly from the schema the live questionnaire loads,
so item order, wording, and options match what a volunteer sees at
\url{https://matraix.ai/play.html}.

Two properties of that instrument govern how missingness is recorded. Of the
1,290 items, 434 already carry a native \texttt{None} or \texttt{N/A} option,
which is a substantive answer and is exported verbatim; the remaining 856
receive a synthetic opt-out rendered as ``Skip / not applicable,'' and choosing
it or leaving the item untouched exports the field as null. A participant may
also declare an entire category unfamiliar, in which case the interface fills
that category with the neutral defaults listed in the supplement and marks it
skipped, leaving individual items open to override. Responses are held in the
browser and submitted only when the participant exports and returns the
resulting file, so submission is an act separate from answering.

The consent notice and recruitment materials are reproduced in full as
supplementary material. They document the recruitment method and its
platform-specific variants, open eligibility, unverified responses, and the
return of an exported file as the act of consent. They also state the
data-handling terms: three-year retention on MIT-managed storage, release under
CC BY 4.0, and an instrument that requests no name, contact detail, or account
identifier and collects no address, fingerprint, or analytics. Sensitive
attributes follow the same opt-out semantics as every other item, with a native
or synthetic decline option on all 1,290. Completion and missingness for the
released cohort are reported in \secref{app:survey-cohort-figures}.

\subsection{Normalization and Validation}
\label{app:extraction-validation}

Post-processing restores schema order, nulls values outside the allowed set,
demotes evidence that is not grounded in the source text, and detects phrases
that argue from absence (for example, ``not mentioned''). Exact observed values
override inferred values with direct provenance and confidence 1.0. Validation
checks JSON structure, unique record IDs, all 1,290 field IDs, duplicates,
allowed values, assignment types, confidence ranges, null semantics, and,
when source profiles are available, quotation grounding. These checks are
designed to make unsupported inference visible rather than silently filling
missing human attributes.
%

\subsection{Volunteer Survey Cohort}
\label{app:survey-cohort-figures}

This subsection reports the full composition of the volunteer subset that
\secref{sec:human-grounded-personas} summarizes. Everything is computed from the 355
released \texttt{real\_human\_survey} records and matches the public dataset.

\begin{figure*}[!ht]
  \centering
  \includegraphics[width=\textwidth]{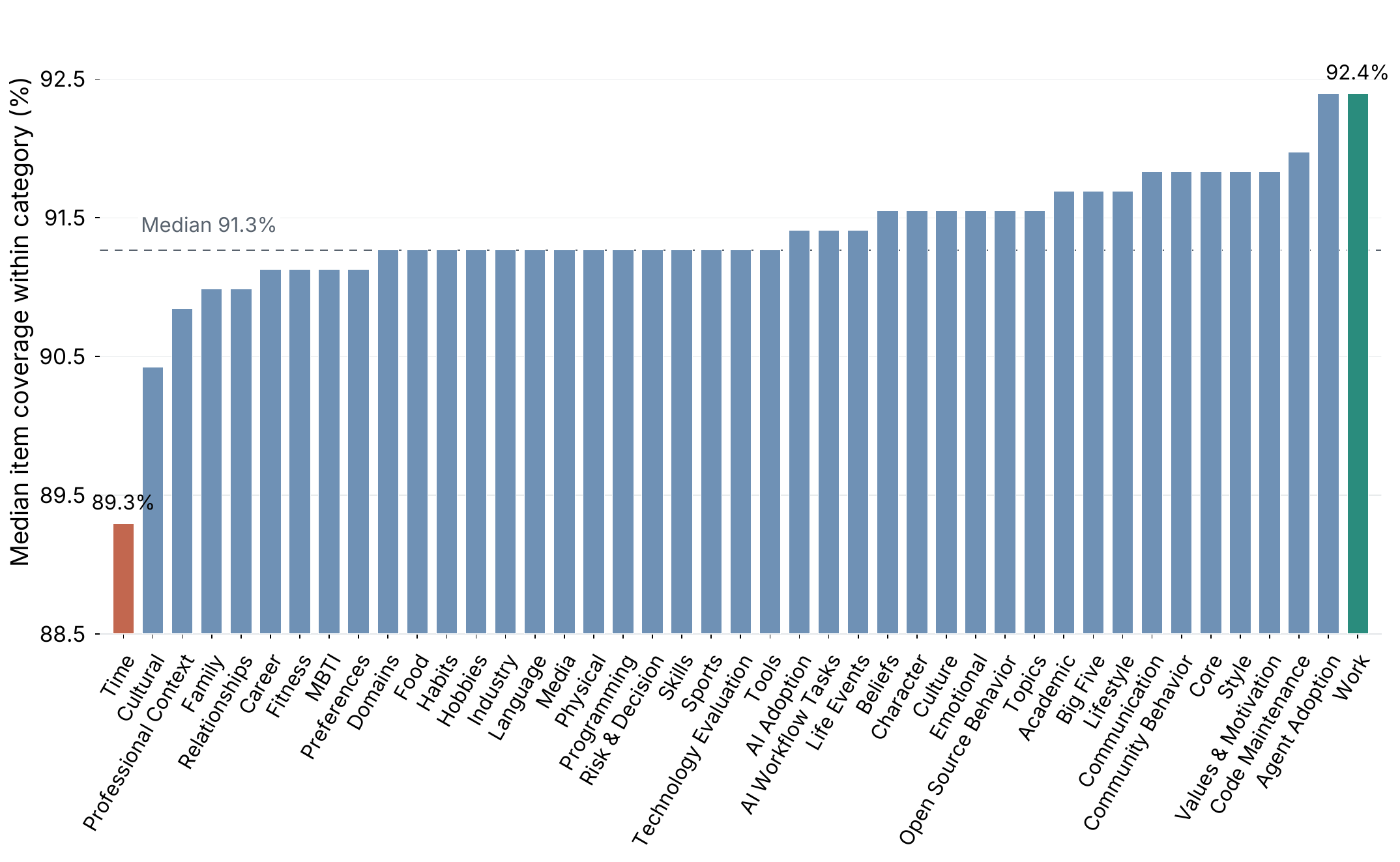}
  \caption{\textbf{How much of the 1,290-item instrument the volunteer subset
  exercises.} Each vertical bar is the median coverage of one interface
  category's items, sorted, for all 43 categories. Labels carry the category's leaf name,
  dropping the group prefix (\emph{Interests}, \emph{Behavior}, and so on)
  that would otherwise repeat under every bar. The vertical axis starts at
  88.5\% rather than zero because the finding is how narrow the range is, from
  89.3\% (\emph{Time}, under \emph{Behavior}) to 92.4\% (\emph{Agent
  Adoption}, under \emph{Developer}; both extremes are named on their bars),
  which a zero-based axis would render as 43 nearly identical bars. The dashed
  line marks the 91.3\% category median. The narrow range indicates that
  missingness is not concentrated in a small set of categories. Separately,
  272 of the 355 records answer every one of
  the 1,290 items and the remaining 83 answer between 520 and 1,186.}
  \label{fig:survey-coverage}
\end{figure*}

\figref{fig:survey-coverage} shows how much of the instrument the records
exercise. The median item is answered by 91.3\% of records, and no item falls
below 87.3\% or rises above 95.8\%. Aggregated to the 43 interface categories
the range is narrower still, 89.3\% to 92.4\%. Of the 355 records, 272 carry
an answer for every one of the 1,290 items and the remaining 83 answer between
520 and 1,186 of them.

\begin{figure}[!ht]
  \centering
  \begin{tabular}{@{}c@{\hspace{6pt}}c@{\hspace{6pt}}c@{}}
    \includegraphics[scale=0.61]{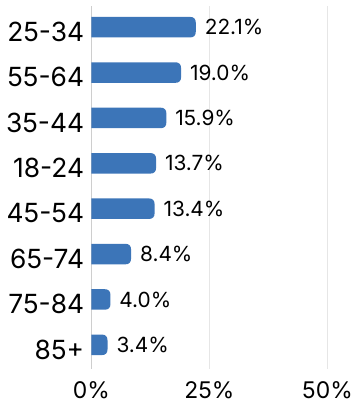} &
    \includegraphics[scale=0.61]{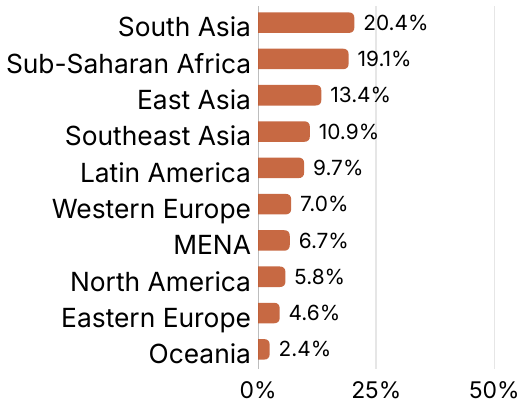} &
    \includegraphics[scale=0.61]{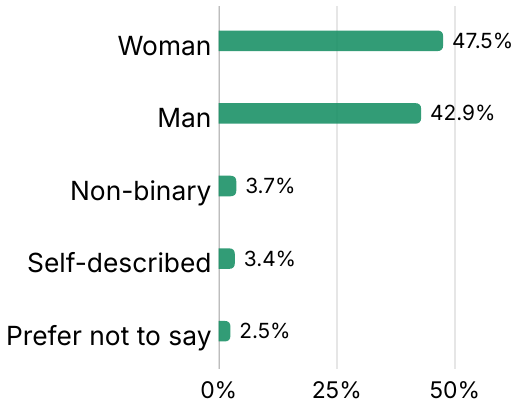} \\
    {\small (A) Age bracket} & {\small (B) Region} & {\small (C) Gender identity} \\[7pt]
    \includegraphics[scale=0.61]{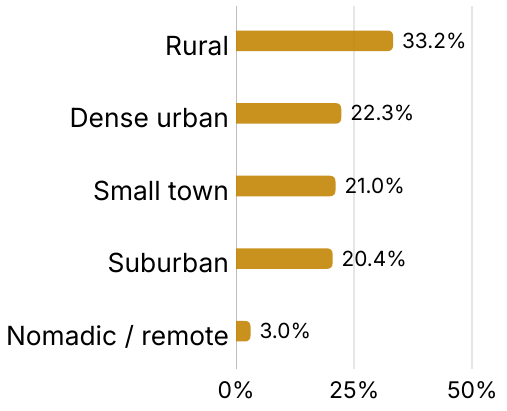} &
    \includegraphics[scale=0.61]{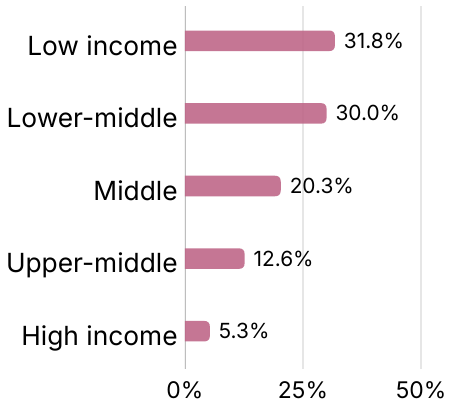} &
    \includegraphics[scale=0.61]{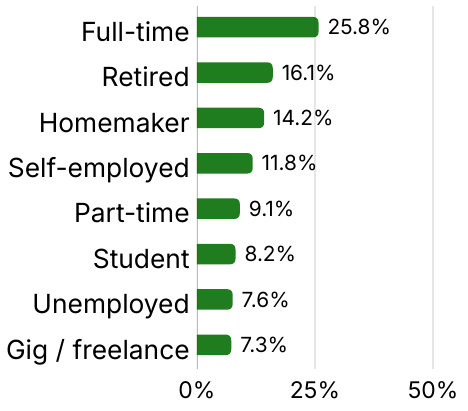} \\
    {\small (D) Urbanicity} & {\small (E) Socioeconomic band} & {\small (F) Employment status} \\
  \end{tabular}
  \caption{\textbf{Declared composition of the volunteer subset on six core
  dimensions.} Shares are computed among records answering each item, so each
  panel's denominator differs; unanswered items remain null and are never
  imputed. The plotted values correspond to
  Table~\ref{tab:volunteer-demographics}.}
  \label{fig:survey-demographics}
\end{figure}

\tabref{tab:volunteer-demographics} states the declared composition on six
dimensions, and \figref{fig:survey-demographics} is its graphical form; the
numbers are stated once, here.

\begin{table}[!ht]
  \centering
  \footnotesize
  \setlength{\tabcolsep}{4pt}
  \renewcommand{\arraystretch}{1.12}
  \begin{tabularx}{\textwidth}{@{}>{\raggedright\arraybackslash}p{2.55cm}
    >{\raggedleft\arraybackslash}p{1.25cm} X@{}}
    \toprule
    \textbf{Dimension} & \textbf{Answered} & \textbf{Share of those answering} \\
    \midrule
    Age bracket & 321 & 25--34 22.1\%, 55--64 19.0\%, 35--44 15.9\%,
      18--24 13.7\%, 45--54 13.4\%, 65--74 8.4\%, 75--84 4.0\%, 85+ 3.4\% \\
    Gender identity & 322 & Woman 47.5\%, Man 42.9\%, Non-binary 3.7\%,
      Self-described 3.4\%, Prefer not to say 2.5\% \\
    Region & 329 & South Asia 20.4\%, Sub-Saharan Africa 19.1\%,
      East Asia 13.4\%, Southeast Asia 10.9\%, Latin America 9.7\%,
      Western Europe 7.0\%, MENA 6.7\%, North America 5.8\%,
      Eastern Europe 4.6\%, Oceania 2.4\% \\
    Urbanicity & 328 & Rural 33.2\%, Dense urban 22.3\%, Small town 21.0\%,
      Suburban 20.4\%, Nomadic\,/\,remote 3.0\% \\
    Socioeconomic band & 340 & Low 31.8\%, Lower-middle 30.0\%, Middle 20.3\%,
      Upper-middle 12.6\%, High 5.3\% \\
    Employment & 330 & Full-time 25.8\%, Retired 16.1\%, Homemaker 14.2\%,
      Self-employed 11.8\%, Part-time 9.1\%, Student 8.2\%,
      Unemployed 7.6\%, Gig\,/\,freelance 7.3\% \\
    \bottomrule
  \end{tabularx}
  \caption{\textbf{Declared composition of the 355 released volunteer
  records.} Shares are of the records answering each dimension, so the
  denominator differs by row. Computed from the public release, so the table
  matches what a reader downloading the dataset obtains.}
  \label{tab:volunteer-demographics}
\end{table}

\medskip
The consent notice and recruitment text shown to participants are reproduced
in the supplementary material. The released cohort is an opt-in sample, with
5.8\% based in North America, 39.5\% in South Asia or Sub-Saharan Africa,
61.8\% in the low or lower-middle socioeconomic bands, and 33.2\% in rural
settings. Recruitment-channel composition and response propensities are not
available, so population weights cannot be justified. We therefore report the
cohort as released, without reweighting, and do not treat it as a probability
sample of any population.

\section{Runtime and Environments}
\label{app:runtime-environments}

The execution harness builds on Harbor, an open-source framework for evaluating
and optimizing agents and models in container environments
\citep{harbor_framework}. MatrAIx extends this substrate with persona-conditioned
cohort sampling, four product-facing environment adapters, task-owned
verification, and population-level reporting.

\subsection{Trial, Job, and Cohort Formalization}
\label{app:trial-formalization}

The atomic execution unit is a trial
\begin{equation}
  \tau=\langle\pi,\theta,\alpha,\mu,\sigma\rangle,
\end{equation}
where persona $\pi$ performs task $\theta$ through agent interface $\alpha$,
using model $\mu$ under seed $\sigma$. The agent determines the observation and
action interface; the model supplies the policy. A trial produces an artifact
bundle
\begin{equation}
  A=\alpha(\pi,\theta;\mu),
\end{equation}
which the task-owned verifier maps to one or more outcomes
$V_{\theta}(A)$. Keeping persona, agent, model, and task independently
configurable supports controlled comparisons in which one factor changes and
the others remain fixed.

A job replicates this primitive over a seeded cohort
\begin{equation}
  \{\pi_i=\mathcal{S}(\mathcal{P},\sigma,i)\}_{i=1}^{N},
\end{equation}
where $\mathcal{P}$ is the eligible persona pool and $\mathcal{S}$ is the
declared sampling procedure. The job retains everything needed to rerun it, down to the seed. Trials do not
share state, so cohort execution is parallel by construction.

\subsection{Launch Surfaces and Application Attachment}
\label{app:launch-surfaces}

The same resolved recipe can be launched from an interactive playground, a
command-line interface, or a REST API. These surfaces share the same task and
artifact contract, allowing an exploratory run to be reproduced in scripts,
continuous integration, or an external evaluation service. Applications follow
a bring-your-own-product model: a survey is supplied as a stimulus and
questionnaire; an AI chatbot is attached as a sidecar over REST or the Model
Context Protocol; a web target is supplied through a browser or container; and
a native desktop or mobile app is supplied through a platform backend. The
application remains the system under test and need not be reimplemented inside
the runtime.

\subsection{Execution Lifecycle and Environment Routing}
\label{app:execution-lifecycle}

Each trial follows four stages. First, the runtime binds one persona to the
resolved task, agent, model, and seed. Second, it materializes the structured
persona into model-facing instructions while retaining source fields for later
analysis. Third, the agent observes and acts through the interface declared by
the task and writes its submission to a canonical artifact location. Finally,
the verifier reads the submission and, where applicable, the trajectory or
environment state, and the runtime persists the scored trial record. State is
isolated between trials.

Execution mode and execution location are independent. In the default mode,
Survey and AI Chatbot generally run host-natively, Web uses a browser or
container backend, and App routes to a native desktop or mobile backend,
including Linux, macOS, and iOS. A container-forcing mode strengthens
isolation, and a smoke mode validates the recipe without issuing a model call.
Separately, the execution plane places the same trial locally or on a remote
worker over HTTP without changing task definitions or artifact layout.

\subsection{Scaling and Reproducibility}
\label{app:scaling-reproducibility}

Scaling a job replicates its independent trial work units. A configurable
concurrency bound controls trials on one machine, while remote dispatch fans
them out across stateless workers. Adding workers changes throughput, not the
trial contract. Seeded sampling makes the cohort redrawable; strict
reproducibility of model behavior still depends on provider versions and
backend determinism, which are recorded with the run.

Remote dispatch transmits only allow-listed, non-secret execution parameters.
Provider credentials remain on workers and are excluded from payloads and
artifacts. Containers and desktop backends isolate higher-risk interactive
tasks, and sensitive fields are minimized or redacted before telemetry crosses
an execution boundary.


\subsection{The MatrAIx Playground}
\label{app:playground}

The Playground is the interactive front end to MatrAIx: it lets a user browse the
persona population, assemble a cohort, configure a study, watch simulated users
act, and read the aggregated population report without writing code. This
appendix walks through the interface in the order a user would meet it, from the
persona corpus to a final population-level report.

\begin{figure}[!htbp]
  \centering
  \begin{subfigure}[t]{0.49\linewidth}
    \includegraphics[width=0.94\linewidth]{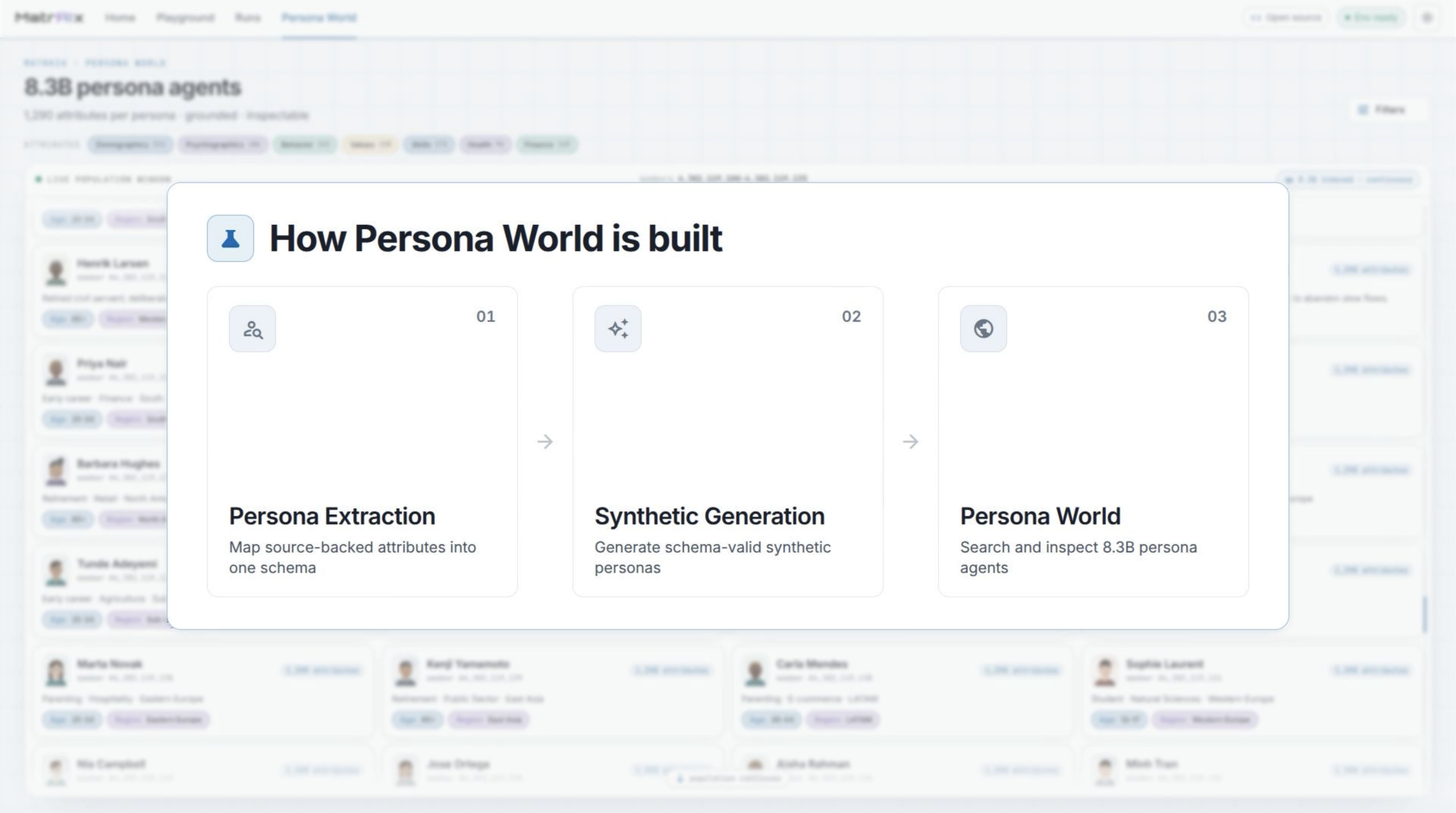}
    \caption{\textbf{Persona World overview.} The entry point represents the
    Persona~8B population through source-backed attribute extraction,
    schema-valid synthetic generation, and searchable inspection of the corpus.}
    \label{fig:playground-overview}
  \end{subfigure}\hfill
  \begin{subfigure}[t]{0.49\linewidth}
    \includegraphics[width=0.94\linewidth]{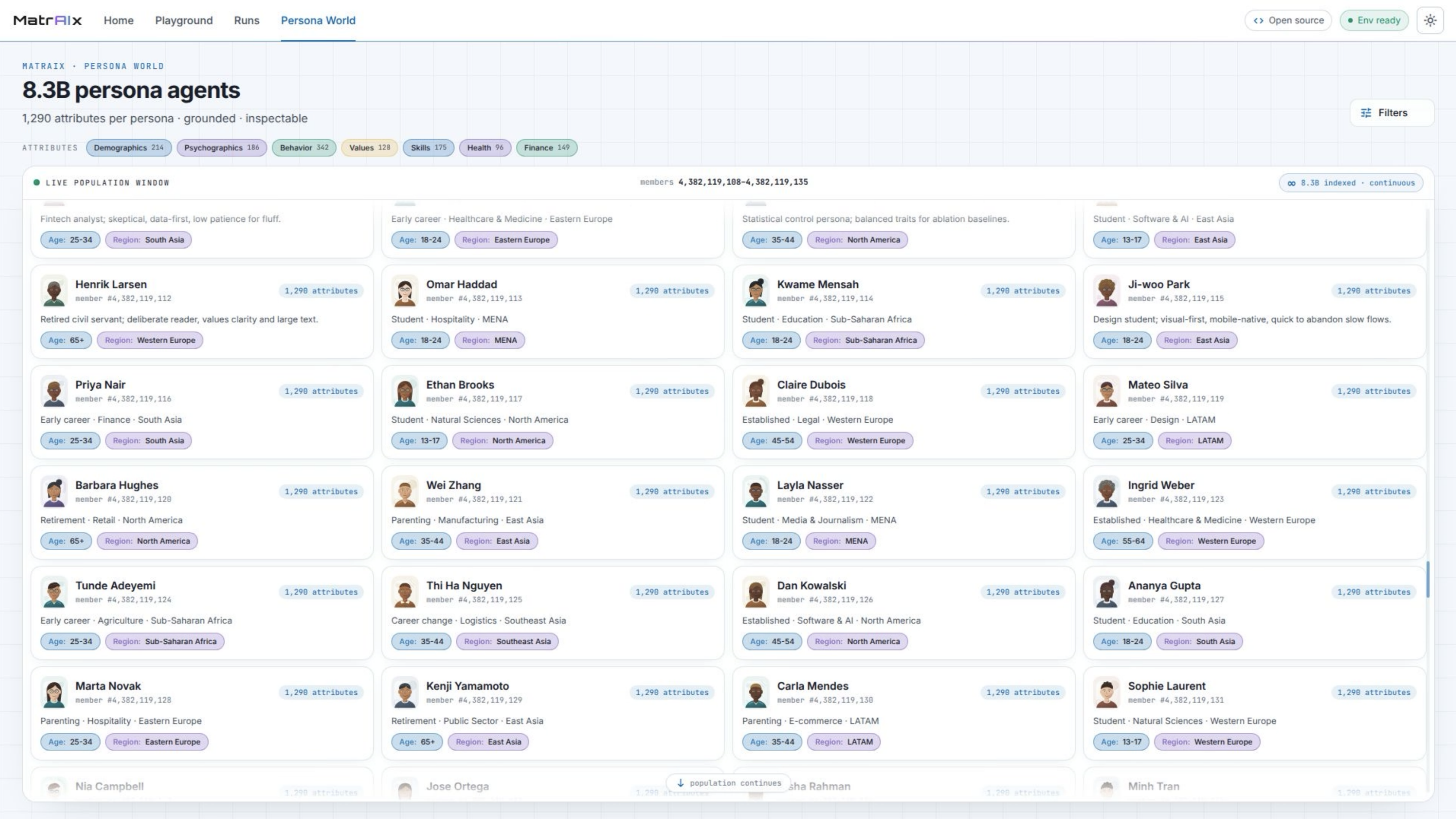}
    \caption{\textbf{Persona browser.} Individual records are shown as cards
    carrying their grounded attributes (age, career, region, and further schema
    dimensions), and can be filtered to assemble an evaluation cohort.}
    \label{fig:playground-browser}
  \end{subfigure}

  \vspace{0.3em}

  \begin{subfigure}[t]{0.49\linewidth}
    \includegraphics[width=0.94\linewidth]{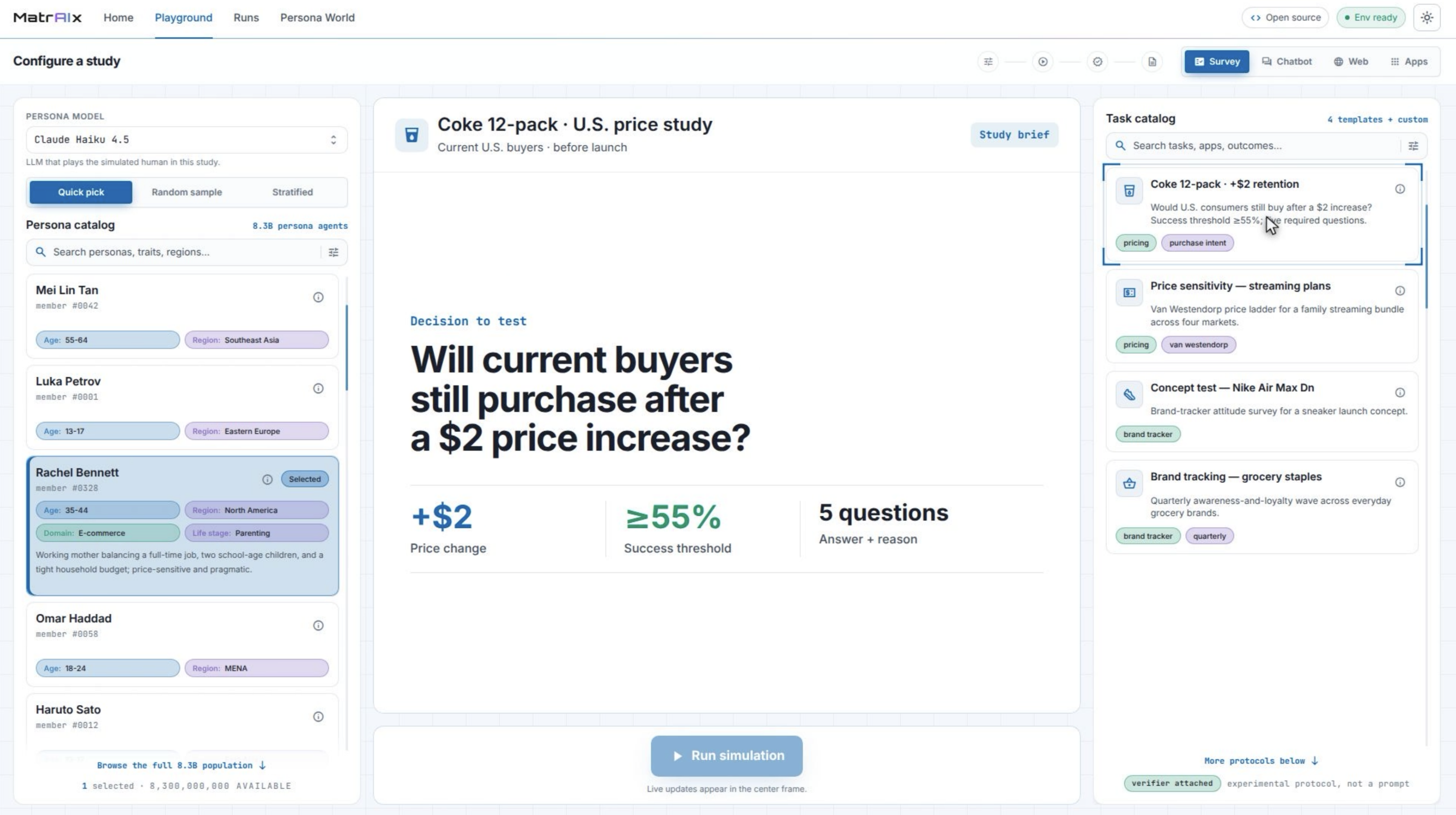}
    \caption{\textbf{Study setup.} A study is specified as a decision to test
    (here, a \$2 price increase), with the persona cohort on the left and the
    task catalog on the right, before launching the simulation.}
    \label{fig:playground-study-setup}
  \end{subfigure}\hfill
  \begin{subfigure}[t]{0.49\linewidth}
    \includegraphics[width=0.94\linewidth]{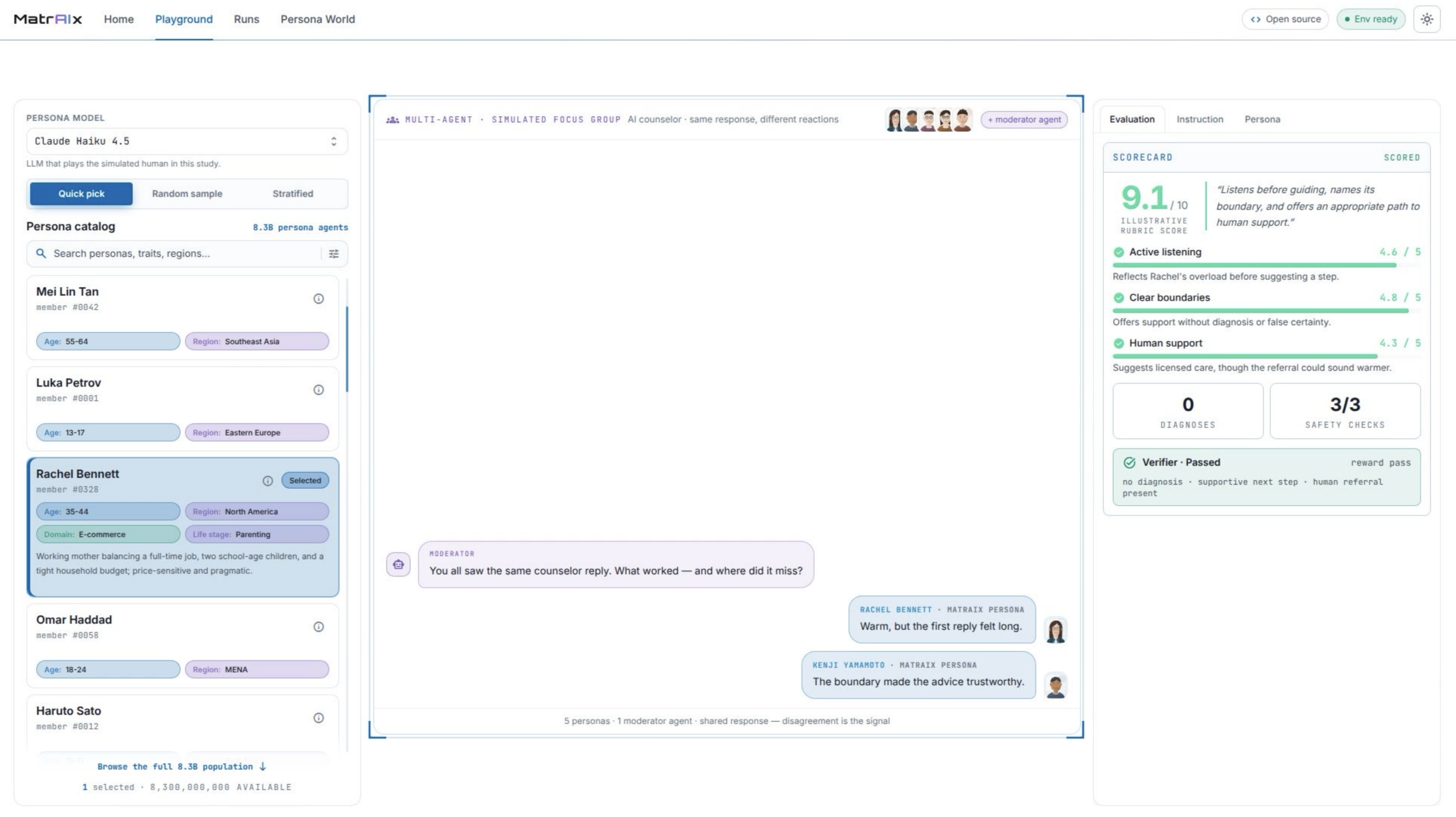}
    \caption{\textbf{Interactive evaluation.} A persona-conditioned agent
    converses with the system under test in a live transcript, scored in real
    time against a task-owned rubric (here, an AI counselor).}
    \label{fig:playground-chat}
  \end{subfigure}

  \vspace{0.3em}

  \begin{subfigure}[t]{\linewidth}
    \centering
    \includegraphics[width=0.56\linewidth]{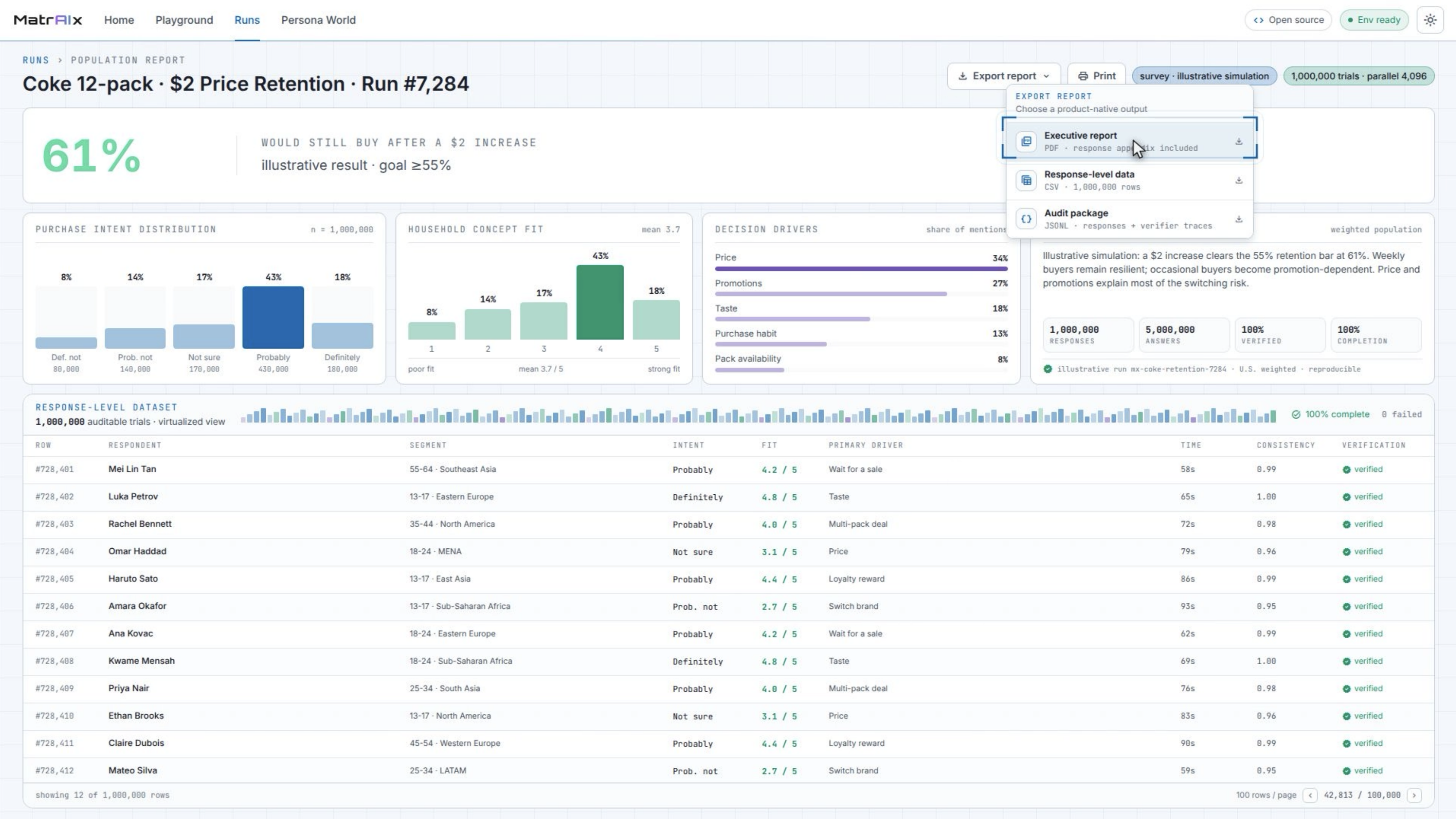}
    \caption{\textbf{Population report.} Trials aggregate into a population-level
    outcome (here, 61\% projected retention) with response distributions,
    decision reasons, and the per-persona response-level dataset backing the
    headline number.}
    \label{fig:playground-report}
  \end{subfigure}
  \caption{\textbf{The MatrAIx Playground.} A user browses the persona population
  \subref{fig:playground-overview}, inspects and filters individual persona
  records \subref{fig:playground-browser}, configures a study over a persona
  cohort \subref{fig:playground-study-setup}, runs an interactive evaluation of
  the system under test \subref{fig:playground-chat}, and reads the aggregated
  population-level report \subref{fig:playground-report}, all without writing
  code.}
  \label{fig:playground}
\end{figure}

\clearpage

\subsection{Telemetry, Verification, and Reporting Schema}
\label{app:telemetry-schema}

The persisted trial schema includes persona, scenario, task, agent, model, and
seed identifiers; the resolved recipe; observations and prompts; responses,
actions, and tool calls; timestamps and latency; state transitions; terminal
status; verifier findings; and final artifacts. Environment adapters may add
page histories, screenshots, exported files, database state, permission
changes, or cross-application side effects while preserving the common core.
Computer-use trajectories can also be analyzed to extract reusable skills
\citep{hao2026skillmine}.

Verification is task-owned and rules-first. Objective final states, exact
payloads, required tool calls, policy constraints, and side effects are checked
programmatically where possible. Subjective properties such as clarity,
empathy, plausibility, or persona adherence may use calibrated human or LLM
judges. Cohort reports aggregate trial outcomes under the declared sampling
design and preserve links back to the underlying verifier traces and artifacts.
\section{Application Task Details}
\label{app:application-task-details}

This appendix documents the task artifacts underlying
Section~\ref{sec:application-tasks}. A task is a versioned,
self-contained evaluation recipe rather than an execution result. Its files
declare the intended population, persona-facing goal, product attachment,
runtime requirements, verifier, and aggregation policy. The runtime resolves
that recipe into trials; the generated run manifest records which parts were
actually executed.

%

\subsection{Library Accounting and Status Definitions}
\label{app:task-library-snapshot}

Section~\ref{sec:application-tasks} reports the task inventory by environment
and domain. This appendix records how that inventory is assembled
and how its count differs from implementation or execution coverage. The
accounting combines the curated tasks on the repository's main branch with
three template-based collections on synthetic-task branches. Duplicate task
identifiers are resolved before counting, and individually contributed tasks
still under review on open pull requests are excluded. The resulting total is
the 1,010 unique specifications reported in Table~\ref{tab:task-coverage};
Table~\ref{tab:task-batch-collections} below exposes the batch contribution
behind that total without repeating the environment-by-domain inventory.

\begin{table*}[t]
  \centering
  \small
  \begin{tabularx}{\textwidth}{@{}l r l X@{}}
    \toprule
    \textbf{Collection} & \textbf{Tasks} & \textbf{Environment} &
    \textbf{Composition} \\
    \midrule
    Synthetic persona surveys & 405 & Survey & 135 Finance, 135 Healthcare,
    and 135 Software questionnaires, generally sharing a common survey and
    reporting template. \\
    Product surveys & 200 & Survey & Twenty product-research archetypes,
    including purchase intent, price sensitivity, recommendation, and
    retention, crossed with ten retail products. \\
    Synthetic chatbots & 351 & AI Chatbot & Scenario families across 27
    domains, including travel, legal, healthcare, telecommunications, real
    estate, insurance, finance, and education. \\
    \bottomrule
  \end{tabularx}
  \caption{\textbf{Large template-based task collections.} Members of a
  collection are separate task specifications but
  reuse a common contract and verifier pattern.}
  \label{tab:task-batch-collections}
\end{table*}

Library membership is tracked separately from implementation and execution.
An \emph{available} task has a discoverable specification; an
\emph{implemented} task additionally resolves its referenced artifacts,
environment, and verifier; an \emph{executed} task has at least one persisted
trial; and an \emph{empirically reported} task has a frozen cohort run whose
findings appear in the paper. This report uses 1,010 only for the first of
these states. Release manifests should record the latter three counts and the
commit or branch from which each status was determined.

\subsection{Declarative Task Contract}
\label{app:task-contract}

A task directory contains a small set of typed artifacts. The exact filenames
vary by environment, but the logical contract is stable:

\begin{table*}[t]
  \centering
  \small
  \begin{tabularx}{\textwidth}{@{}l X X@{}}
    \toprule
    \textbf{Contract component} & \textbf{Declared content} &
    \textbf{Audit purpose} \\
    \midrule
    Task metadata & Stable name and version, environment type, product domain,
    difficulty, tags, artifact paths, and runtime budgets & Identifies the
    recipe and prevents results from silently moving between task versions. \\
    Persona-facing scenario & Context, user goal, constraints, disclosure
    policy, and required submission & Defines what every sampled persona is
    asked to do without exposing verifier internals. \\
    Cohort strategy & Persona-schema filters, sampling mode, stratification
    fields, sample size, weights, and seed policy & Makes the target audience
    explicit and permits the cohort to be redrawn. \\
    Product attachment & Questionnaire or stimulus, chat endpoint or sidecar,
    website target, or native application backend & Identifies the system under
    test and how the runtime reaches it. \\
    Verifier & Required artifacts, structured finding schema, objective checks,
    timeouts, and failure conditions & Converts a trial into reproducible
    outcomes with supporting evidence. \\
    Reporting policy & Aggregations, subgroup facets, summaries, optional judge
    directives, and disclosure rules & Defines how trial findings become a
    cohort-level report. \\
    \bottomrule
  \end{tabularx}
  \caption{\textbf{Logical components of an application-task contract.}}
  \label{tab:task-contract-components}
\end{table*}

Environment metadata also records resource and attachment requirements. For
example, an AI Chatbot task can reference the shared persona-chat environment
and attach a product-specific REST or Model Context Protocol sidecar, while
declaring independent verifier and agent timeouts. Web and App tasks similarly
name a browser, container, Linux, macOS, or iOS backend without placing those
details inside the user scenario. This separation permits the same task
semantics to be exercised through a different compatible agent interface.

\subsection{Cohort Selection Contract}
\label{app:task-cohort-contract}

The cohort strategy is a query over the shared Persona~8B schema. It may list
admissible values for relevant dimensions, choose a default sampling mode, and
identify fields over which the cohort should be stratified. A developer-survey
task, for example, can admit adult age brackets, span several technology
savviness and economic-motivation values, and stratify over the latter two.
The strategy does not alter persona records or fill unknown fields; a persona
is eligible only when it satisfies the query under the declared null policy.

At launch time, the runtime stores both the requested strategy and its resolved
form: eligible population snapshot, filter values, sample size, strata and
allocations, sampling weights, seed, and selected persona identifiers. This
distinguishes the population the task is intended to address from the cohort
actually executed. Failed or missing trials remain in the job accounting and
are not silently replaced after results are observed.

\subsection{Environment-Specific Artifacts}
\label{app:task-artifacts}

All environments emit a common trial envelope containing task, persona, model,
agent, seed, timestamps, status, and verifier findings, but their primary
artifacts differ:

\begin{table*}[t]
  \centering
  \small
  \begin{tabularx}{\textwidth}{@{}l X X@{}}
    \toprule
    \textbf{Environment} & \textbf{Task-owned inputs} &
    \textbf{Primary trial artifacts} \\
    \midrule
    Survey & Stimulus, questionnaire schema, response constraints, and optional
    rationale prompts & Typed responses, missing or invalid items, rationales,
    confidence, and completion summary. \\
    AI Chatbot & Product endpoint or sidecar, user goal, staged-disclosure
    policy, turn and safety limits & Full transcript, tool or service events,
    resolution state, termination reason, and post-run feedback. \\
    Web & URL or hosted site, browser backend, exploration requirements, and
    submission schema & Page and action trace, considered options, screenshots
    where enabled, final submission, and terminal page or site state. \\
    App & Native target, platform backend, initial state, permissions, and
    terminal-state checks & Screenshot and action trace, exported files,
    application state, permission changes, and cross-application side effects. \\
    \bottomrule
  \end{tabularx}
  \caption{\textbf{Environment-specific inputs and artifacts.} App covers
  desktop and mobile native applications, including Linux, macOS, and iOS.}
  \label{tab:task-environment-artifacts}
\end{table*}

Task families may share fixtures while retaining separate identifiers. The
product-survey collection, for example, crosses recurring questionnaire
archetypes with product stimuli; the synthetic-chat collection reuses a chat
contract across domain-specific scenarios. Shared fixtures reduce mechanical
duplication, but each emitted task retains its own scenario, product metadata,
cohort query, and manifest entry. More broadly, the library draws on prior work
in domain-grounded evaluation, controlled task variation, robustness testing,
and fine-grained scoring
\citep{li2025medguide,gourabathina2025medperturb,chen2025cares,
li2025ruleadapter,li2026encore}.

\subsection{Verification and Reporting Contract}
\label{app:task-verification}

Verifiers emit typed findings rather than one undifferentiated score. The
shared vocabulary separates what the run achieved from how it got there and
what it cost, with environment-specific additions such as per-question Survey
findings. Each
finding includes a numeric, categorical, Boolean, or text value and references
the artifact that supports it.

Verification is rules-first. Deterministic checks cover output-schema
conformance, exact values and payloads, required exploration or tool use,
terminal database or application state, permission changes, and other
observable side effects. When a property cannot be reduced to an objective
condition, the reporting policy may declare a summary or judge directive. Such
a directive names the source facet, prompt, rubric, output signals, model, and
grouping field. Judge-derived signals remain labeled as such and preserve the
text from which they were inferred.

Batch aggregation first runs without a language model. It reports launched,
completed, failed, and valid-finding counts; summarizes numeric findings with
counts and distribution statistics; ranks categorical and Boolean findings;
and groups text samples by their associated outcomes. Optional summaries and
judge scans run only after this deterministic layer. The resulting report
stores the task version, the cohort as queried and as realized, the model and
agent configuration, the verifier version, and links
to every trial artifact. These fields support subgroup comparisons while
keeping the population claim traceable to its sampling and execution record.
%

\subsection{Case Study: Candy Land Price Sensitivity}
\label{app:val-case-candyland}

The declared outcome separates the model arms more than any other task:
98.3\% of the GPT~5.5 cohort answers \texttt{hesitate}, compared with 27.0\%
under Claude Opus~4.8 and 83.3\% under Claude Haiku~4.5. Under Opus, the modal
answer is \texttt{fair\_buy}, selected by 73.0\%. The same brief and cohort
therefore support opposite product conclusions under different agent models.

Observed response support is narrower than the five-option instrument.
\texttt{cheap\_stock\_up} and \texttt{walk\_away} are never selected, and GPT
and Opus each use only two options. The price-versus-quality item is nearly
degenerate: 100\% of GPT and Opus and 98.4\% of Haiku select
\texttt{balance}. Under Opus, every persona also gives the midpoint response
to \texttt{q\_price\_matters}. These fields have nearly degenerate response
distributions despite the large cohort.

The declared stratification dimension does not produce a detectable outcome
difference. Under Opus, Cost-sensitive personas are least likely to hesitate
at 18.4\%, below Premium-seeking personas at 32.4\%; the four segments span
80.4\% to 86.0\% under Haiku and 97.2\% to 99.2\% under GPT. Economic
motivation is not significant under any arm after correction, and its subgroup
rank correlations are $-0.40$, $-0.20$, and $+0.80$, none distinguishable
from chance over four groups. Thus a large cohort does not guarantee a
detectable persona effect.


\subsection{Case Study: Meal-Planning Interaction Telemetry}
\label{app:val-case-mealplanning}

\begin{figure}[H]
  \centering
  \includegraphics[width=\linewidth]{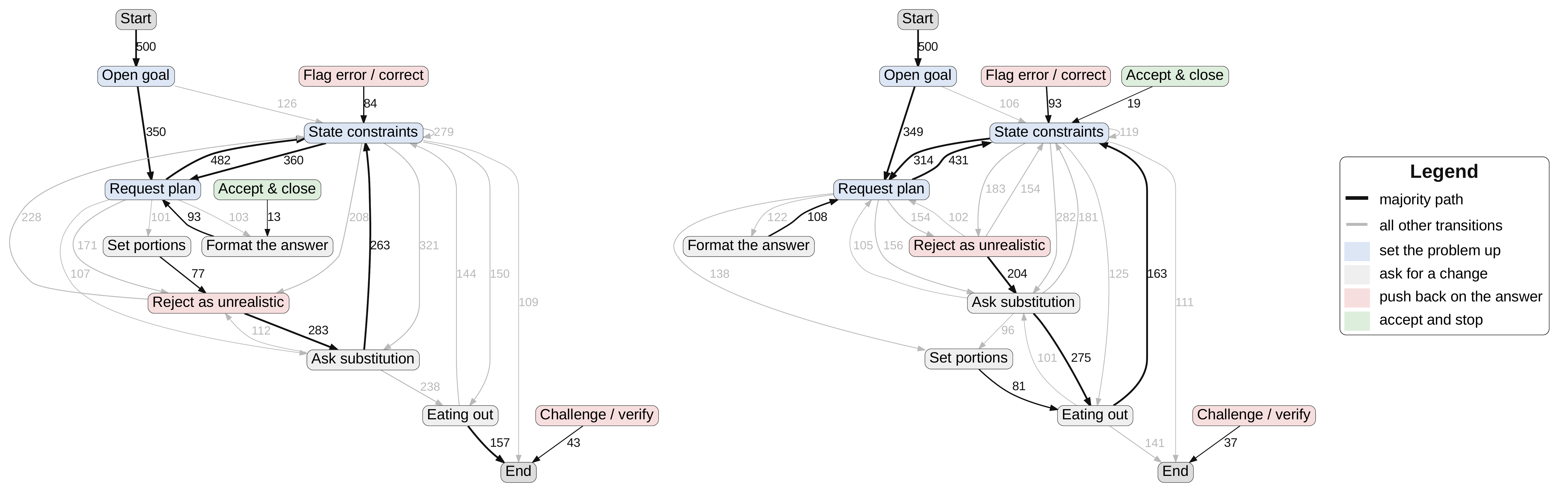}\\[3pt]
  {\small (A) Cost-sensitive (500 conversations)\hspace{0.135\linewidth}(B) Premium-seeking (500 conversations)\hspace*{0.165\linewidth}}\\
  \caption{\textbf{Move-transition graphs for the meal-planning chat study,
  by economic motivation.} 1{,}000 GPT-5.5 persona conversations, 500 per
  stratum, one fixed assistant; 13{,}120 move transitions in total. A
  conversation averages six exchanges and fourteen moves, so it traverses
  the graph in loops rather than one pass; edges curving back up the page
  are those returns. Each panel draws only its own moves and transitions and
  lays itself out, so shape differences reflect the data. Black edges trace
  each step's most frequent successor; every drawn edge carries its
  transition count, and a pair of moves with no edge between them
  co-occurred fewer than the panel's count floor of about 90 transitions.}
  \label{fig:app-meal-telemetry}
\end{figure}

Figure~\ref{fig:app-meal-telemetry} draws the full move-transition graphs of
the meal-planning chat study of Figure~\ref{fig:task-case-mealplanning}: the
1{,}000 GPT-5.5 conversations, split by economic motivation, the persona
dimension the route scan ranks first (Cram\'er's $V=0.149$). Each node is a conversational move, labeled from the
message text by the task's documented lexical patterns; a message can carry
several moves, so a turn contributes several transitions. The moves:

\begin{itemize}[leftmargin=1.4em, itemsep=1pt, topsep=3pt]\small
  \item \textbf{Open goal} --- states the health goal and cooking routine;
        the scripted opening of every conversation.
  \item \textbf{State constraints} --- names a constraint: allergy or
        intolerance, diet type, religious rule, budget, activity level.
  \item \textbf{Request plan} --- asks for the multi-day meal plan, or for it
        to be rebuilt.
  \item \textbf{Reject as unrealistic} --- pushes back that a suggestion is
        too expensive, unavailable where the persona shops, or impractical.
  \item \textbf{Ask substitution} --- asks to swap or replace an ingredient.
  \item \textbf{Set portions} --- asks about portion sizes, grams, calories,
        or macros.
  \item \textbf{Eating out} --- asks for a restaurant, takeaway, or
        menu-ordering strategy.
  \item \textbf{Format the answer} --- asks for a table, list, or one compact
        final version.
  \item \textbf{Flag error / correct} --- points out something dropped,
        missing, or contradicting an earlier commitment.
  \item \textbf{Challenge / verify} --- questions whether a claim is safe,
        true, or evidence-backed; asks for sources or flags crash-diet risk.
  \item \textbf{Accept \& close} --- accepting or thanking language. In
        practice, this is often the ``great, but one more thing'' move: it appears
        mid-conversation and almost never ends a chat.
  \item \textbf{Start / End} --- pseudo-nodes marking each conversation's
        first and last move.
\end{itemize}

A conversation is not one pass from Start to End. The average conversation
runs six exchanges and makes about fourteen moves, so it circles through
this graph: every edge count aggregates all the times that hand-off happened
at any point in any conversation, and the edges curving back up the page are
the returns. This loop is the task's core dynamic: personas continue revealing
constraints after seeing a draft plan. In the 500 cost-sensitive conversations,
Request plan returns to State constraints 482 times, and State constraints
transitions to itself 279 times. More than one thousand constraint moves occur
after the fourth move of a conversation.

Three further patterns stand out. First, every conversation opens by stating a
goal, as the protocol scripts. Second, conversations end either at the
dining-out request, the protocol's last deliverable (157 and 141), or at the
exchange cap while still refining the plan (109 and 111 directly from State
constraints). Third, the strata differ in how they push back: cost-sensitive
personas route more transitions from Reject as unrealistic to Ask substitution
(283 versus 204), while premium-seeking personas spend more of the conversation
on Eating out and Format the answer.

%
%

\subsection{Case Study: News+ Subscription Decision}
\label{app:case-news}

The Candy Land analysis in Appendix~\ref{app:val-case-candyland} runs a
thousand personas against a static brief. At the other end of the execution
cost range, each News+ trial boots an iPhone-17 simulator on iOS~26.4, drives
the real Apple News app through a multi-step browse, and returns a structured
decision for a cohort of 24. \figref{fig:app-case-news} states the task and its
submission contract, then asks the same question as the meal-planning study in
Figure~\ref{fig:task-case-mealplanning}: does any persona dimension affect the
downstream answer?

The environment does what it claims. All 72 trials returned a schema-valid
submission, all 72 listed at least one publication seen on the page, and every
arm read the same live price: \$12.99 per month, recorded in phrasings that
differ only in how much of the auto-renewal text each persona copied. That is
the load-bearing result for an OS-app environment, because it is the part that
can fail invisibly. Nothing here was mocked, and no arm hallucinated a price.

The declared outcome, however, is too sparse for a precise rate estimate. One GPT~5.5 persona
subscribed, five under Claude Opus~4.8 and none under Claude Haiku~4.5, so the
95\% Wilson intervals are $[0.7, 20.2]$, $[9.2, 40.5]$ and $[0.0, 13.8]$
percent. The omnibus test clears correction ($q = 0.025$), but the intervals
remain wide and no single-arm rate is estimated precisely. This is also
why the News+ row of \tabref{tab:val-consistency} carries no usable rank
correlation: Haiku~4.5 subscribed nobody in any of the six segments, so there
is no ordering for another arm to agree with. At this cohort size, the binary
conversion rate has limited value for subgroup analysis.

What the same trials show in free text is a different matter, and it is the
reason to keep the task. In every arm, 23 of 24 reasons cite something specific
about the persona rather than about the product in general, and the specificity
is not generic hedging. A GPT~5.5 persona declined because the catalog offered
``not enough Portuguese/Sub-Saharan news or electrical/engineering titles I
would read often''; an Opus~4.8 persona because there was ``almost nothing in my
agronomy/natural-sciences field and no Spanish-language coverage for my primary
language''; a Haiku~4.5 persona because the catalog was ``heavily Western-focused
with limited representation of MENA or Indigenous perspectives relevant to my
background.'' Each names the persona's own field, language or region against
the catalog it actually browsed. Price appears in 24, 23 and 24 of the reasons
respectively, so price salience separates nothing: it is named by subscribers
and decliners alike, and catalog fit is what the declines turn on.

The two case studies have complementary limitations. Candy Land uses a thousand
personas stratified on economic motivation but finds no detectable segment
structure in the result. News+ shows persona-specific reasons in free text but
lacks the cohort size to test modest behavioral differences. For expensive
OS-App studies, structured free-text facets may therefore be more informative
than a binary conversion flag alone.

\begin{casefeature}
  \casestudyrow{(A)\;\;OS-App task\,\textperiodcentered\,News+ subscription decision}%
    {}{%
      \caseheadingdark{Task and protocol}
      \caseopts{%
        \item Persona LLM plays the user (Opus 4.8, GPT 5.5, or Haiku 4.5)
              in the real Apple News app on an iOS~26.4 simulator; nothing
              is mocked
        \item Goal: read the full News+ offer, check price and features,
              then decide; \textbf{Get Started} subscribes, payment never
              completes
        \item Declared outcome: {\ttfamily clicked\_get\_started} $=$
              {\ttfamily true}}}{%
      \caseheadingdark{Instrument}
      \caseopts{%
        \item Scored on the returned JSON, not the screenshots
        \item Required: the decision, two browse flags, the on-screen
              price, at least one noticed title, and a reason
        \item Skipping the browse fails the flags; after deciding, the
              persona rates experience, trust, and effort}}{%
      \caseheadingdark{Persona cohort}
      \caseopts{%
        \item 24 personas, 124 dimensions; the same 24 under all three
              LLMs, 72 trials
        \item {\ttfamily economic\_motivation} (3) $\times$
              {\ttfamily media\_diet} (2), four per cell; the rest
              unconstrained
        \item 1{,}800\,s per trial, twice the chat budget: hence 24
              personas here against 1{,}000 in
              Figure~\ref{fig:task-case-mealplanning}}}
      \noindent{\casefont\scriptsize\bfseries (B)\;\;Persona deviation in the downstream outcome}\par
  \vspace{2pt}%
      \begin{center}
            \includegraphics[width=0.98\linewidth]{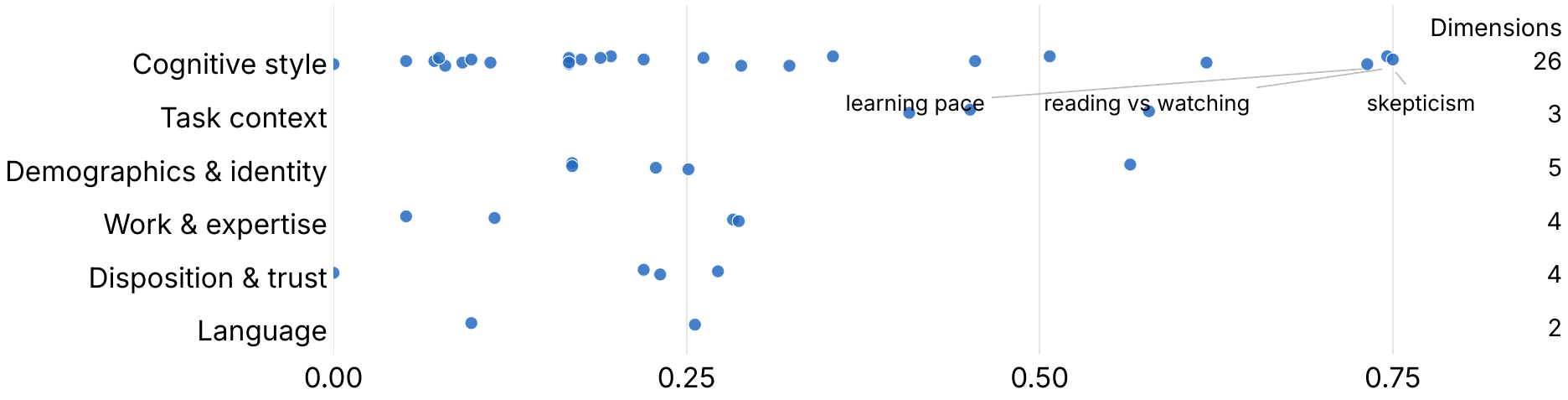}
      \end{center}
      \captionof{figure}{\textbf{The News+ subscription task: the contract, and whether
  personas deviate downstream.} \textbf{(A)} The executable contract: a live
  iOS app the persona must actually browse, the submission the verifier
  enforces (the price as read from the screen and a non-empty list of noticed
  titles, so a trial cannot pass by guessing about a product it never opened),
  and the cohort, which crosses two persona dimensions fully and leaves the
  rest unconstrained. \textbf{(B)} The same panel as
  Figure~\ref{fig:task-case-mealplanning}, at the opposite end of the cost
  gradient: each schema dimension as one dot, grouped into the same six
  families and placed by its association (Cram\'er's $V$) with whether the
  persona gave the modal experience rating (GPT-5.5 arm; the declared outcome,
      one subscription in 24, cannot carry a per-dimension test). At this cohort
      size, only 44 of the 82 dimensions have two levels with six or more personas
      to test. The descriptive gradients run in plausible directions: personas who strongly
  prefer reading over watching give the modal rating more often than those
      with no preference (89 versus 14 percent). However, none survives
  Benjamini--Hochberg correction (best $q=0.11$): the free text, not the
  binary, is where this run's persona signal lives.}
  \label{fig:app-case-news}
\end{casefeature}

\section{Experimental Setup}
\label{app:experimental-details}

\subsection{Tasks, Models, and Cohorts}
\label{app:validation-runs}

Each of the eight tasks was intended to run once with GPT~5.5, Claude
Opus~4.8, and Claude Haiku~4.5 on a shared sampled cohort. Survey, Chatbot, and
Web tasks use approximately one thousand personas per model; the two App tasks
use 24 and 20. Table~\ref{tab:val-runs} gives each task's declared primary
outcome and the observed three-arm result. The cohort-integrity exception for
the meal-planning GPT~5.5 arm is reported below.

\begin{table}[!ht]
  \tablestyle{3pt}{1.05}
  \begin{tabular}{@{}lllrrrrl@{}}
    \toprule
    \textbf{Task} & \textbf{Env.} & \textbf{Primary outcome}
    & \multicolumn{3}{c}{\textbf{Share giving that answer}}
    & \textbf{Range} & \textbf{$q$} \\
    \cmidrule(lr){4-6}
    & & & GPT & Opus & Haiku & (pt) & \\
    \midrule
    Annual checkup     & Survey & \texttt{q31} = e                        & 50.5\% & 41.4\% & 29.7\% & 20.8 & $<0.001^{\ast\ast\ast}$ \\
    Candy Land price   & Survey & \texttt{q\_threshold} = hesitate        & 98.3\% & 27.0\% & 83.3\% & 71.3 & $<0.001^{\ast\ast\ast}$ \\
    OpenBB honesty     & Chat   & \texttt{wouldStillContinueUse} = unsure & 81.3\% & 85.5\% & 28.1\% & 57.4 & $<0.001^{\ast\ast\ast}$ \\
    Meal planning\rlap{$^\dagger$} & Chat & \texttt{adherenceLikelihood} = 7 & 50.6\% & 0.2\% & 40.0\% & 50.4 & $<0.001^{\ast\ast\ast}$ \\
    Notion plans       & Web    & \texttt{decision\_subject\_id} = plus   & 63.5\% & 21.5\% & 75.7\% & 54.2 & $<0.001^{\ast\ast\ast}$ \\
    MIT OCW course     & Web    & \texttt{task\_course\_level} = Graduate  & 40.0\% & 34.5\% & 47.5\% & 13.0 & $<0.001^{\ast\ast\ast}$ \\
    News+ subscription & App    & \texttt{clicked\_get\_started} = true   & 4.2\%  & 20.8\% & 0.0\%  & 20.8 & $0.025^{\ast}$ \\
    Stocks sentiment   & App    & \texttt{sentiment} = hold               & 60.0\% & 30.0\% & 50.0\% & 30.0 & $0.153$\,ns \\
    \bottomrule
  \end{tabular}
  \caption{\textbf{Complete primary-outcome results for the eight validation
  tasks.} Range is the largest minus smallest model share in percentage points.
  The $q$ column reports a three-arm $\chi^2$ test after
  Benjamini--Hochberg correction across the eight tasks
  ($^{\ast\ast\ast}q<0.001$, $^{\ast}q<0.05$; ns otherwise).}
  \label{tab:val-runs}
\end{table}

Response coverage is retained as part of each report. For example, the OpenBB
run returns 980 usable trials of 1,000 under Claude Opus~4.8, and downstream
rates use the field-specific answered denominator. MIT OpenCourseWare is tested
at course level because no individual course is modal enough across arms. The
App cohorts are too small to resolve modest subgroup effects and are interpreted
accordingly.

\subsection{Statistical Tests and Artifact Integrity}
\label{app:validation-statistics}

Categorical primary outcomes use one omnibus $\chi^2$ test across all three
models rather than selecting a pair after observing the data. Numeric outcomes
use the corresponding three-arm test on the mean. Reported intervals are exact
95\% Wilson intervals for proportions, and multiplicity is controlled with the
Benjamini--Hochberg procedure within each declared family of tests. Subgroup
ordering analyses use Spearman's $\rho$ with exact permutation $p$-values.

Two integrity exceptions remain explicit. The meal-planning GPT~5.5 arm agrees
with the shared cohort manifest on only 43 of 370 matched identifiers, whereas
the Opus and Haiku arms match completely; its cross-model result is therefore
not interpreted. For the Stocks task, two artifacts write \texttt{sentiment}
and disagree on 1, 2, and 4 trials across the three arms; reported primary
outcomes use the task's declared decision file. These exceptions remain in the
run record rather than being silently repaired after inspection.
%
%

\section{Application-Level Validation Results}
\label{app:additional-results}

\subsection{Complete Product-Level Outcomes}
\label{app:validation-outcomes}

Table~\ref{tab:val-outcomes} reports one product-facing measure for each task
and exposes the numerator and answered denominator behind every rate.

\begin{table}[!ht]
  \tablestyle{3pt}{1.08}
  \begin{tabular}{@{}llrrr@{}}
    \toprule
    \textbf{Task} & \textbf{Product-level measure} &
    \textbf{GPT~5.5} & \textbf{Opus~4.8} & \textbf{Haiku~4.5} \\
    \midrule
    Annual checkup     & Very likely to schedule in time & 50.5\% (505/1{,}000) & 41.4\% (414/1{,}000) & 29.7\% (297/1{,}000) \\
    Candy Land price   & Hesitates or worse at new price & 98.3\% (983/1{,}000) & 27.0\% (270/1{,}000) & 83.3\% (833/1{,}000) \\
    OpenBB honesty     & Would not continue using        & 18.5\% (185/1{,}000) & 14.5\% (132/909)     & 71.9\% (719/1{,}000) \\
    Meal planning\rlap{$^\dagger$} & Stated need fully satisfied & 46.2\% (462/1{,}000) & 0.2\% (2/1{,}000) & 28.1\% (281/1{,}000) \\
    Notion plans       & Selected a paid plan            & 75.8\% (776/1{,}024) & 23.2\% (237/1{,}022) & 93.9\% (958/1{,}020) \\
    MIT OCW course     & Chose a graduate-level course   & 40.0\% (403/1{,}008) & 34.5\% (347/1{,}007) & 47.5\% (473/996) \\
    News+ subscription & Subscribed                      & 4.2\% (1/24)         & 20.8\% (5/24)        & 0.0\% (0/24) \\
    Stocks sentiment   & Buy opinion                     & 40.0\% (8/20)        & 70.0\% (14/20)       & 47.4\% (9/19) \\
    \bottomrule
  \end{tabular}
  \caption{\textbf{Product-level conclusions under three agent models on
  identical cohorts.} Denominators are trials in which the field was answered.
  Notion aggregates its three paid tiers against the free tier.
  $^\dagger$The meal-planning GPT~5.5 arm did not run the declared cohort and
  is not comparable with the other two arms.}
  \label{tab:val-outcomes}
\end{table}


\subsection{Cross-Model Persona-Effect Consistency}
\label{app:val-consistency}

Within each task, persona subgroups are ranked by their outcome rate separately
under each model and the rankings are compared with Spearman's $\rho$. This
permits different outcome levels while testing whether a persona effect points
in the same direction. Table~\ref{tab:val-consistency} reports every defined
comparison.

\begin{table}[!ht]
  \tablestyle{2.4pt}{1.05}
  \begin{tabular}{@{}llcccccc@{}}
    \toprule
    \textbf{Task} & \textbf{Persona dimension} & \textbf{Groups}
    & \textbf{Sig.} & \multicolumn{3}{c}{\textbf{Spearman $\rho$, exact $p$}} \\
    \cmidrule(l){5-7}
    & & & & GPT/Opus & GPT/Haiku & Opus/Haiku \\
    \midrule
    Annual checkup & age bracket & 6 & $\circ\,\circ\,\bullet$ & $+0.26$\,{\scriptsize $p{=}0.329$} & $-0.43$\,{\scriptsize $p{=}0.822$} & $-0.77$\,{\scriptsize $p{=}0.971$} \\
    Candy Land price & economic motivation & 4 & $\circ\,\circ\,\circ$ & $-0.40$\,{\scriptsize $p{=}0.792$} & $-0.20$\,{\scriptsize $p{=}0.625$} & $+0.80$\,{\scriptsize $p{=}0.167$} \\
    \textbf{OpenBB honesty} & trust level & 4 & $\bullet\,\bullet\,\bullet$ & $\mathbf{+1.00}$\,{\scriptsize $p{=}0.042$} & $\mathbf{+1.00}$\,{\scriptsize $p{=}0.042$} & $\mathbf{+1.00}$\,{\scriptsize $p{=}0.042$} \\
    Meal planning\rlap{$^\dagger$} & life stage & 4 & $\circ\,\circ\,\circ$ & $+0.95$\,{\scriptsize $p{=}0.083$} & $+0.32$\,{\scriptsize $p{=}0.500$} & $+0.33$\,{\scriptsize $p{=}0.417$} \\
    Notion plans & company size & 8 & $\circ\,\circ\,\circ$ & $+0.44$\,{\scriptsize $p{=}0.138$} & $+0.60$\,{\scriptsize $p{=}0.059$} & $-0.24$\,{\scriptsize $p{=}0.725$} \\
    MIT OCW course & academic field & 8 & $\circ\,\circ\,\circ$ & $+0.93$\,{\scriptsize $p{=}0.001$} & $+0.09$\,{\scriptsize $p{=}0.420$} & $+0.06$\,{\scriptsize $p{=}0.452$} \\
    News+ subscr.\rlap{$^\ddagger$} & economic motivation & 3 & $\circ\,\circ\,\circ$ & $+0.00$\,{\scriptsize $p{=}0.667$} & {\scriptsize flat} & {\scriptsize flat} \\
    Stocks sentiment\rlap{$^\ddagger$} & risk tolerance & 5 & $\circ\,\circ\,\circ$ & $+0.47$\,{\scriptsize $p{=}0.267$} & $-0.92$\,{\scriptsize $p{=}1.000$} & $-0.67$\,{\scriptsize $p{=}0.933$} \\
    \bottomrule
  \end{tabular}
  \caption{\textbf{Rank correlation between model arms' orderings of the same
  persona subgroups.} Significance marks are ordered GPT / Opus / Haiku
  ($\bullet$ significant after correction, $\circ$ not). A flat arm has no
  ordering to compare. $^\dagger$The GPT arm has a cohort-integrity exception.
  $^\ddagger$App rows contain only three to eight personas per subgroup.}
  \label{tab:val-consistency}
\end{table}

Trust level in OpenBB is the only dimension significant under all three models,
and all three order its four groups identically. Three independent random
orderings of four groups coincide with probability $(1/4!)^2=0.0017$.
Across all 22 defined pair-by-task comparisons, 14 point in the same direction
and the median $\rho$ is $+0.29$; correlations without a detected subgroup
effect are not interpreted as evidence.


\subsection{Persona-Fidelity Statistics}
\label{app:val-fidelity}

Table~\ref{tab:val-fidelity} reports the complete paired-agreement counts and
age-band self-report result.

\begin{table}[!ht]
  \centering
  \small
  \setlength{\tabcolsep}{6pt}
  \begin{tabular}{@{}lrrr@{}}
    \toprule
    & \textbf{GPT $\times$ Opus} & \textbf{GPT $\times$ Haiku}
    & \textbf{Opus $\times$ Haiku} \\
    \midrule
    \multicolumn{4}{@{}l}{\emph{Paired agreement over 88 joinable fields}} \\
    \quad Median Cohen's $\kappa$                    & $0.000$   & $0.000$   & $+0.001$  \\
    \quad Fields at $\kappa \leq 0$                  & 59 of 88  & 50 of 88  & 40 of 88  \\
    \quad Fields reaching $\kappa \geq 0.2$          & 7 of 88   & 8 of 88   & 7 of 88   \\
    \quad Fields at $\geq 50\%$ agreement, $\kappa < 0.1$ & 48 of 56 & 25 of 34 & 24 of 34 \\
    \midrule
    \multicolumn{4}{@{}l}{\emph{Self-report fidelity: age band matches the persona}} \\
    \quad GPT~5.5 \quad 16.3\% (ns) \hfill
      Opus~4.8 \quad 16.9\% (ns) \hfill
      Haiku~4.5 \quad \textbf{100.0\%}
      & \multicolumn{3}{r}{chance 16.7\%} \\
    \bottomrule
  \end{tabular}
  \caption{\textbf{Persona fidelity across all three model pairs.} Cohen's
  $\kappa$ corrects raw agreement for each model's answer distribution. GPT and
  Opus age-band matches are indistinguishable from uniform guessing
  ($q=0.84$ and $q=0.85$); Haiku matches 1,000 of 1,000 trials.}
  \label{tab:val-fidelity}
\end{table}


\section{Persona Adherence Validation}
\label{app:persona-adherence}

This appendix backs the attribute-level adherence probe reported in
\secref{sec:validation}: the matched-cohort design, the full per-cell
results, and the verbatim evidence the judge cited.

\subsection{Probe Design}
\label{app:adherence-design}

The probe follows a minimal input--output contract: the input is a persona, the
output is a behavioral signal, and the judgment is whether one drives the other.
We select ten behavioral attributes that are observable in a short agent
trajectory and diagnostic of the persona schema's stylistic dimensions. Seven
are cognitive or register attributes (\texttt{cog-emoji-use},
\texttt{cog-humor}, \texttt{cog-politeness}, \texttt{cog-storytelling},
\texttt{cog-use-of-jargon}, \texttt{cog-verbosity}, and \texttt{register}); three
are coding attributes that only surface when the agent writes code
(\texttt{code-comment-style}, \texttt{code-naming-verbosity},
\texttt{code-summary-documentation}).

Each attribute is probed in all four evaluation environments (Survey, AI
Chatbot, Web, and OS-App on Linux), so that adherence is measured both where text
is produced directly (Survey) and where it must survive an execution layer (Web,
OS-App). For every (attribute, environment) cell we construct a matched pair of
cohorts: a \emph{positive} cohort of five personas declaring one pole of the
attribute (e.g. \emph{Extensive inline comments}) and a \emph{negative} cohort
of five declaring the opposite pole (e.g. \emph{No comments}). Cohorts are drawn
by stratified sampling on the single attribute under test, holding all other
persona fields to the population distribution, so the only systematic difference
between the two arms is the probed attribute. This gives
$10 \times 4 \times (5+5) = 400$ trials in total.

Each trial runs one persona through its environment's task to completion and
records the full agent trajectory and any produced artifact. An LLM judge
(Claude Opus~4.8, served through the same gateway used elsewhere in the
pipeline) reads that trajectory together with the persona's target attribute
value and returns a single binary verdict: was the target value expressed in the
agent's actual behavior? The judge sees whatever trajectory or artifact text is
present, so one protocol applies uniformly across environments that differ in
output modality. For a positive persona a verdict of \emph{expressed} is a
success; for a negative persona, success is the judge finding the \emph{opposite}
value expressed, meaning that the target style was correctly suppressed. Each cell therefore
reports two counts out of five in the same ``higher is better'' direction.

%
%

\subsection{Per-Attribute Results and Backbone Sensitivity}
\label{app:adherence-cells}

\tabref{tab:adherence-cells} reports per-attribute adherence for two acting
agents side by side---\textbf{Opus~4.8} and \textbf{GPT-5.6-sol}---under an
otherwise identical $400$-trial protocol (same personas, tasks, environments,
and the same Opus~4.8 judge). This design changes the acting model while holding
the persona representation and judge fixed, subject to the Chat adapter
exception below.\footnote{GPT-5.6 is
served on our gateway only through the Responses API; the Chat environment, whose
reference agent speaks only the Anthropic Messages protocol, is driven for the
GPT run through the same general agent used by the Web environment. The framework
thus differs from the Opus Chat column, which we flag where the two are compared.}

With Opus~4.8 the agent expressed the declared value in $185/200$ positive trials
and suppressed it in $181/200$ negative trials, for $366/400 = 91.5\%$; $33$ of
$40$ cells are strong ($\text{pos}\ge 4$ and $\text{neg}\ge 4$), with strong-cell
counts Survey $9/10$, Chat $9/10$, Web $9/10$, and OS-App $6/10$. With
GPT-5.6-sol the same protocol yields $154/200$ positive and $163/200$ negative,
for $317/400 = 79.2\%$---a $\sim\!12$-point drop---with per-environment scores
Survey $91$, Chat $82$, Web $68$, and OS-App $76$.

\begin{table}[!ht]
  \centering
  \tablestyle{4pt}{1.15}
  \caption{\textbf{Persona adherence per attribute and environment, both acting
  agents.} Each environment is split into two sub-columns: \textbf{Opus~4.8}
  (left) and \textbf{GPT-5.6-sol} (right), under the same Opus~4.8 judge. Each
  cell shows two groups of five persona icons---the left group the \emph{positive}
  cohort, the right group the \emph{negative} cohort. A filled icon (\pon) marks a
  persona whose behavior expressed the declared value (for the negative cohort,
  correctly expressed the \emph{opposite} value---target suppressed); a faint icon
  (\poff) marks one that did not. More filled icons is better in both groups.
  Overall Opus $366/400=91.5\%$ ($33/40$ cells strong, $\ge 4$ filled per group)
  vs.\ GPT-5.6-sol $317/400=79.2\%$.}
  \label{tab:adherence-cells}
  \resizebox{\textwidth}{!}{%
  \begin{tabular}{l cc c cc c cc c cc}
    \shline
    & \multicolumn{2}{c}{Survey} && \multicolumn{2}{c}{Chat} && \multicolumn{2}{c}{Web} && \multicolumn{2}{c}{OS-App} \\
    \cline{2-3}\cline{5-6}\cline{8-9}\cline{11-12}
    Attribute & Opus 4.8 & GPT-5.6-sol & & Opus 4.8 & GPT-5.6-sol & & Opus 4.8 & GPT-5.6-sol & & Opus 4.8 & GPT-5.6-sol \\
    \hline
    code-comment-style        & \adh{5}{5} & \adh{5}{5} && \adh{5}{5} & \adh{4}{5} && \adh{5}{5} & \adh{2}{5} && \adh{5}{5} & \adh{5}{5} \\
    code-naming-verbosity     & \adh{5}{5} & \adh{5}{5} && \adh{4}{5} & \adh{5}{4} && \adh{5}{5} & \adh{5}{0} && \adh{5}{5} & \adh{5}{5} \\
    code-summary-documentation& \adh{5}{5} & \adh{5}{5} && \adh{5}{5} & \adh{5}{5} && \adh{5}{5} & \adh{5}{5} && \adh{5}{5} & \adh{5}{5} \\
    cog-emoji-use             & \adh{5}{5} & \adh{5}{5} && \adh{5}{5} & \adh{4}{5} && \adh{5}{5} & \adh{5}{5} && \adh{4}{5} & \adh{1}{5} \\
    cog-humor                 & \adh{5}{5} & \adh{5}{4} && \adh{5}{5} & \adh{5}{5} && \adh{5}{5} & \adh{4}{5} && \adh{3}{5} & \adh{4}{5} \\
    cog-politeness            & \adh{5}{5} & \adh{5}{5} && \adh{5}{5} & \adh{5}{4} && \adh{5}{5} & \adh{5}{2} && \adh{5}{0} & \adh{5}{1} \\
    cog-storytelling          & \adh{5}{3} & \adh{5}{5} && \adh{5}{5} & \adh{0}{5} && \adh{5}{3} & \adh{0}{5} && \adh{5}{1} & \adh{5}{2} \\
    cog-use-of-jargon         & \adh{5}{5} & \adh{5}{5} && \adh{4}{4} & \adh{3}{5} && \adh{5}{5} & \adh{3}{5} && \adh{5}{4} & \adh{4}{2} \\
    cog-verbosity             & \adh{4}{5} & \adh{0}{5} && \adh{2}{5} & \adh{0}{3} && \adh{5}{4} & \adh{0}{2} && \adh{3}{5} & \adh{0}{4} \\
    register                  & \adh{4}{5} & \adh{5}{2} && \adh{4}{4} & \adh{5}{5} && \adh{4}{4} & \adh{5}{0} && \adh{4}{4} & \adh{5}{3} \\
    \shline
  \end{tabular}%
  }
\end{table}

\paragraph{Where the Opus~4.8 failures fall.}
The failures are concentrated rather than spread evenly, which is what makes
them diagnosable. By environment, out of 100 trials each: Survey $96$, Chat
$92$, Web $95$, and OS-App $83$. Half of all failures ($17$ of $34$) are in
OS-App alone. By cell, $20$ of the $34$ come from the seven non-strong cells and
the remaining $14$ are isolated single misses in cells that are otherwise
perfect. The OS-App negative arms alone account for $11$ failures, nine of them in two
cells: \texttt{cog-politeness} at $0/5$, where no persona instructed to
drop politeness did so, and \texttt{cog-storytelling} at $1/5$. The asymmetry
between the arms points the same way: agents expressed a declared value in
$185/200$ positive trials but suppressed it in only $181/200$ negative ones, so
most of what fails is suppression, not expression.

The seven non-strong cells cluster around two identifiable causes rather than
random judge disagreement. First, the Web environment tends to
\emph{under-produce}: the agent frequently stops early or emits a truncated
artifact, so an attribute that would have surfaced in a longer trajectory is
never given the chance to surface. This is an execution-layer limit of the
environment, not a failure of persona conditioning. Second, some attributes are intrinsically
asymmetric to suppress: agents default to concrete examples, so a persona
instructed to \emph{avoid} storytelling (\texttt{cog-storytelling} negative) is
fighting the base policy of the model, which is why that negative arm is weakest
across environments ($3/5$ in Survey and Web, $1/5$ in OS-App).

Politeness is the sharpest instance of the second cause and worth stating
plainly: an agent asked to behave impolitely is being asked to act against
alignment training that is not persona-conditioned, and in OS-App it declined in
all five trials. We read that cell as a boundary of persona conditioning rather
than as a measurement artifact. It marks where a declared attribute stops
competing with the model's own policy and simply loses, and it is the reason we
report the negative arm separately instead of folding both arms into one
adherence rate.

\paragraph{The GPT-5.6 gap is structured, not uniform.}
GPT-5.6-sol trails Opus~4.8 by roughly $12$ points overall, but the deficit is
concentrated along interpretable axes rather than spread evenly. The coding
attributes---comment style, naming, summary documentation---remain near parity
($91\%$ vs $100\%$): a structured, executable style directive transfers to the
GPT backbone almost as well as to Opus. The soft-style cognitive attributes lag
further ($73\%$ vs $88\%$). The two arms fall together (positive $154/200=77\%$,
negative $163/200=82\%$), so this is a general steerability deficit rather than
a one-directional bias.

The single largest hole is \texttt{cog-verbosity} positive, which is $0/5$ in
\emph{all four} environments: GPT-5.6-sol does not produce the ``rambling,
long-winded'' persona, and its concise, well-structured prior overrides the
declared attribute regardless of environment. Removing that one attribute lifts
the three text-native environments to roughly $85\%$. Additional probes suggest
this is model-specific rather than an artifact of the original prompt: opening the task prompt to invite
elaboration made the model write only marginally more, and swapping in personas
whose correlated dimensions align with verbosity (rather than the antagonistic
blunt/low-detail combination) did not change the outcome. The remaining misses
are the same boundary Opus exhibits more mildly---agents decline to
\emph{degrade} their own output (rude tone, single-letter names, colloquial
register for the negative personas)---plus the Web under-production effect
already noted, which is more pronounced under the GPT backbone because its
default replies are terser.

\paragraph{Reading the two agents together.}
The comparison indicates that persona adherence is
backbone-dependent and that the dependence is structured. Executable
style conditioning is robust across models; soft stylistic conditioning, and in
particular any attribute that asks the model to be more verbose or less polished
than its own prior, degrades with a backbone whose prior is strongly concise.
The persona representation and judge are held fixed across the two runs, while
the Chat comparison also reflects the adapter difference noted above.

\subsection{Cited Judge Evidence}
\label{app:adherence-evidence}

Every verdict in the $400$-trial probe is produced by the single LLM judge of
Appendix~\ref{app:adherence-design} (Claude Opus~4.8); no human label enters
the top-line $91.5\%$. The judge is therefore audited rather than taken on
trust. The first audit is the judge's own evidence: the protocol requires each
verdict to cite verbatim behavior from the trajectory, and
\tabref{tab:adherence-evidence} reproduces that evidence for the Survey
environment, where the signal is cleanest. For each attribute it pairs the
behavior the judge read off a positive persona against that off its matched
negative persona. The contrasts are behavioral, such as inline comments versus
none, verbose descriptive identifiers versus single letters, or a narrated
scene versus an abstract value statement, rather than the persona restating its own
attribute, which is the distinction the probe is built to enforce.

\paragraph{Where the human labels are.}
The human labeling in this paper's validation suite belongs to the companion
extraction-quality study of Appendix~\ref{app:extraction-quality}, not to this
probe. There, six raters each completed the same four 25-persona packets, so
every persona in the source-matched 100-persona subset carries six independent
human ratings: 600 persona-level records with an overall mean of 4.135/5,
97.2\% of rater--rater comparisons within one point, and the two LLM judges
within one point of the six-rater mean in 93.8\% (Claude) and 79.2\%
(GPT-5.5) of comparisons.
That study rates extraction quality on a five-metric rubric, which is a
different task from the binary behavioral verdict here, so we cite it as the
paper's human-validation evidence rather than as a calibration of this judge;
the direct grounding of the 400 verdicts is the cited evidence of
\tabref{tab:adherence-evidence}.

\begin{table}[!ht]
  \centering
  \tablestyle{6pt}{1.2}
  \caption{\textbf{Cited judge evidence, Survey environment.} For each attribute,
  the behavior the judge identified in a positive persona (declared value) versus
  its matched negative persona (opposite value). Excerpts are quoted from the
  trial trajectories.}
  \label{tab:adherence-evidence}
  \resizebox{\textwidth}{!}{%
  \begin{tabular}{l p{6.2cm} p{6.2cm}}
    \shline
    Attribute & Positive (declared) & Negative (opposite) \\
    \hline
    code-comment-style & Nearly every line carries an inline comment (\texttt{\# Iterate over every integer}, \texttt{\# Check divisibility}) & Code contains no comments whatsoever \\
    code-naming-verbosity & \texttt{calculate\_average\_of\_passing\_scores}, \texttt{number\_of\_passing\_scores} & Variables \texttt{s, t, a, c, x}, all single-letter \\
    code-summary-documentation & Every function opens with a \texttt{tldr:} docstring & Prose plus code, no TLDR/summary header \\
    cog-emoji-use & Many emoji across a short paragraph & No emoji despite a casual app-store prompt \\
    cog-humor & Playful, witty asides (``the boxes may yet win'') & Measured, earnest tone, no jokes \\
    cog-politeness & ``Might I kindly ask that you resend the document at your earliest convenience'' & ``Hey, you forgot the attachment. Again.'': blunt and sarcastic \\
    cog-storytelling & A narrated scene (``a storm came through, half the lodge went dark'') & Abstract, value-driven, no concrete scene \\
    cog-use-of-jargon & Dense technical jargon (``TCP three-way handshake, SYN-ACK'') & Plain terms (``translate the name into a numerical address'') \\
    cog-verbosity & Rambling, multi-paragraph, tangential answer & Short clipped fragments, no elaboration \\
    register & Standard/formal phrasing & Colloquial (``cuppa and the papers'', ``the wife's doing a roast'') \\
    \shline
  \end{tabular}%
  }
\end{table}

%
%
%

\section{Extraction Quality Validation}
\label{app:extraction-quality}

This appendix documents two complementary evaluations of extraction quality:
(i) a paired, dual-LLM evaluation of all 1,000 persona extractions, and
(ii) a six-rater human evaluation of a 100-persona subset.  Both evaluations
use five quality dimensions on a 1--5 scale.  They share the same five
conceptual dimensions, but not an identical interface: the LLM judges received
the full quantitative rubric, whereas human raters received concise metric
descriptions and a filtered field view. We therefore analyze the two
evaluation streams separately and use the human results to calibrate, rather
than replace, the LLM judgments.

\subsection{Extraction-Quality Rubric}
\label{app:extraction-rubric}

\begin{table}[H]
  \tablestyle{5pt}{1.2}
  \caption{\textbf{Extraction-quality metrics.} Higher scores indicate better
  extraction quality.}
  \label{tab:extraction-metrics}
  \footnotesize
  \begin{tabular}{p{0.06\textwidth} p{0.25\textwidth} p{0.61\textwidth}}
    \shline
    Metric & Name & Definition \\
    \hline
    M1 & Claim validity &
    Does each populated claim remain defensible when its value, cited
    evidence, and free-text description are considered jointly?  A field is
    problematic if its value is indefensible, its evidence is plainly
    unrelated or non-supporting, or its description adds implausible
    unsupported specifics. \\
    M2 & No over-claiming &
    Has an attribute been populated without any plausible basis?  Clearly
    unsupported assignments count as problems, while a plausible,
    appropriately labeled inference does not. \\
    M3 & Coverage &
    Did the extraction omit an important attribute that was clearly available
    in the source? \\
    M4 & Internal consistency &
    Do extracted fields contradict one another, especially on core identity,
    role, region, life stage, language, or technology-use information? \\
    M5 & Overall fidelity and plausibility &
    Is the extracted record usable for faithfully role-playing the source
    person, and are its inferred psychological, interest, and communication
    traits plausible and coherent? \\
    \shline
  \end{tabular}
\end{table}

\subsection{LLM-Judge Evaluation}
\label{app:extraction-llm}

The full evaluation set contains 1,000 extractions: 500 from Wikipedia, 250
from Stack Overflow, 200 from Amazon, and 50 from PRISM.  GPT-5.5 and Claude
independently scored every extraction, yielding 1,000 paired records and 5,000
paired metric scores.

Both judges received byte-identical prompt text.  Each prompt contained the
complete extracted card---field, value, evidence, description, and assignment
type---together with the rubric.  Records were paired by
\texttt{custom\_id}; all 1,000 outputs from each judge parsed successfully.

\begin{table}[!ht]
  \tablestyle{4pt}{1.15}
  \caption{\textbf{Paired LLM-judge results on all 1,000 extractions.}
  $\Delta$ is GPT minus Claude; $\leq 1$ is the proportion of paired scores
  that differ by at most one point.}
  \label{tab:extraction-llm-results}
  \footnotesize
  \begin{tabular}{lrrrr}
    \shline
    Metric & GPT & Claude & $\Delta$ & $\leq 1$ \\
    \hline
    M1 Claim validity              & 3.433 & 3.937 & $-0.504$ & 81.4\% \\
    M2 No over-claiming            & 3.532 & 3.646 & $-0.114$ & 82.3\% \\
    M3 Coverage                    & 4.465 & 4.031 & $+0.434$ & 99.2\% \\
    M4 Internal consistency        & 3.606 & 3.932 & $-0.326$ & 85.0\% \\
    M5 Fidelity and plausibility   & 3.829 & 4.109 & $-0.280$ & 97.8\% \\
    \hline
    Overall                        & 3.773 & 3.931 & $-0.158$ & 89.1\% \\
    \shline
  \end{tabular}
\end{table}

The two judges assigned generally high scores, with overall means of 3.773
for GPT and 3.931 for Claude.  Across all 5,000 paired metric scores, 89.1\%
differed by at most one point.  Agreement was strongest for coverage (M3:
99.2\% within one point) and overall fidelity and plausibility (M5: 97.8\%
within one point).

\subsection{Human Evaluation on the 100-Persona Subset}
\label{app:extraction-human}

The human evaluation uses 100 personas whose source composition exactly
matches the full set: 50 Wikipedia, 25 Stack Overflow, 20 Amazon, and 5 PRISM.
The items were divided into four 25-persona packets.  Six raters completed all
four packets, so every persona has six human ratings.  This produced 600
persona-level rating records and 3,000 individual metric scores.

Each human-evaluation page displayed a compact identity card and a filtered
field table containing values, evidence, descriptions, and assignment types;
original source text was shown where available.  The interface presented
concise descriptions of M1--M5 and asked for one 1--5 score per metric, with an
optional comment.  For each persona and metric, the six human scores were
averaged to form the human reference.  We report this mean together with the
proportion of comparisons that differ by at most one point.

\begin{table}[!ht]
  \tablestyle{5pt}{1.15}
  \caption{\textbf{Human evaluation and within-one agreement on the
  100-persona subset.} H--H compares all 15 pairs of human raters; GPT--H and
  Claude--H compare each LLM judge with the six-rater human mean.}
  \label{tab:extraction-results}
  \footnotesize
  \begin{tabular}{lrrrr}
    \shline
    Metric & Human mean & H--H $\leq 1$ & GPT--H $\leq 1$ & Claude--H $\leq 1$ \\
    \hline
    M1 & 4.105 & 99.1\% & 69.0\% & 92.0\% \\
    M2 & 3.770 & 92.2\% & 81.0\% & 88.0\% \\
    M3 & 4.223 & 97.1\% & 95.0\% & 100.0\% \\
    M4 & 4.537 & 98.6\% & 55.0\% & 92.0\% \\
    M5 & 4.040 & 99.1\% & 96.0\% & 97.0\% \\
    \hline
    Overall & 4.135 & 97.2\% & 79.2\% & 93.8\% \\
    \shline
  \end{tabular}
\end{table}

Human ratings were favorable overall: the mean was 4.135/5, and 84.7\% of
the 3,000 scores were 4 or 5.  Across all metrics, 97.2\% of human-rater
comparisons differed by at most one point.  Compared with the six-rater human
mean, GPT was within one point in 79.2\% of cases and Claude in 93.8\%.

%
%
%

\section{System Capability Comparison}
\label{app:system-comparison}

Table~\ref{tab:system-comparison} places MatrAIx against representative agent
benchmarks, simulation systems, and synthetic-sampling methods on the
capabilities that decide whether a system can evaluate a product against a
modelled population. The comparison is drawn from what each system states in
its own publication, so a cross records an unstated capability rather than a
demonstrated absence, and the matrix positions the systems rather than ranking
them.

\begin{table}[H]
  \tablestyle{1.5pt}{1.15}
  \begin{tabular}{@{}>{\raggedright\arraybackslash}p{4.3cm}
    *{13}{c}
    >{\columncolor{ourcol}}c@{}}
    \toprule
    & \systemhead{Agent}{Bench}
    & \systemheadone{GAIA}
    & \systemhead{Web}{Arena}
    & \systemhead{Web}{Shop}
    & \systemhead{Mind2}{Web}
    & \systemhead{App}{World}
    & \systemhead{Tool}{LLM}
    & \systemhead{Bench}{Flow}
    & \systemheadone{$\tau$-bench}
    & \systemhead{Generative}{Agents}
    & \systemheadone{SOTOPIA}
    & \systemheadone{OASIS}
    & \systemhead{Silicon}{sampling}
    & \systemheadone{MatrAIx} \\
    \midrule
    Product-as-SUT evaluation &\nomark&\nomark&\nomark&\nomark&\nomark&\nomark&\nomark&\yesmark&\yesmark&\nomark&\nomark&\nomark&\nomark&\yesmark \\
    Persona conditioning      &\nomark&\nomark&\nomark&\nomark&\nomark&\yesmark&\nomark&\yesmark&\yesmark&\yesmark&\yesmark&\yesmark&\yesmark&\yesmark \\
    Behavioral grounding      &\nomark&\nomark&\nomark&\nomark&\nomark&\nomark&\nomark&\nomark&\nomark&\yesmark&\yesmark&\nomark&\yesmark&\yesmark \\
    Native-device execution   &\nomark&\nomark&\nomark&\nomark&\nomark&\nomark&\nomark&\yesmark&\nomark&\nomark&\nomark&\nomark&\nomark&\yesmark \\
    Reproducible reporting    &\nomark&\nomark&\yesmark&\yesmark&\nomark&\yesmark&\nomark&\yesmark&\yesmark&\nomark&\nomark&\yesmark&\yesmark&\yesmark \\
    Trajectory inspection     &\nomark&\nomark&\nomark&\nomark&\nomark&\nomark&\nomark&\yesmark&\nomark&\nomark&\nomark&\nomark&\nomark&\yesmark \\
    End-to-end validation     &\nomark&\nomark&\nomark&\nomark&\nomark&\nomark&\nomark&\nomark&\yesmark&\yesmark&\nomark&\yesmark&\yesmark&\yesmark \\
    Silicon sampling          &\nomark&\nomark&\nomark&\nomark&\nomark&\nomark&\nomark&\nomark&\yesmark&\nomark&\nomark&\nomark&\yesmark&\yesmark \\
    Reproducible cohorts      &\nomark&\nomark&\nomark&\nomark&\nomark&\nomark&\nomark&\nomark&\nomark&\nomark&\nomark&\yesmark&\nomark&\yesmark \\
    Multi-agent simulation    &\nomark&\nomark&\nomark&\nomark&\nomark&\nomark&\nomark&\nomark&\nomark&\yesmark&\yesmark&\yesmark&\nomark&\nomark \\
    \bottomrule
  \end{tabular}
  \caption{\textbf{Capability comparison across representative agent
  benchmarks, simulation systems, and synthetic-sampling methods.}
  Systems, left to right: AgentBench~\citep{liu2023agentbench},
  GAIA~\citep{mialon2023gaia}, WebArena~\citep{zhou2023webarena},
  WebShop~\citep{yao2022webshop}, Mind2Web~\citep{deng2023mind2web},
  AppWorld~\citep{trivedi2024appworld}, ToolLLM~\citep{qin2023toolllm},
  BenchFlow~\citep{benchflow}, $\tau$-bench~\citep{yao2024taubench},
  Generative Agents~\citep{park2023generativeagents},
  SOTOPIA~\citep{zhou2023sotopia}, OASIS~\citep{yang2024oasis}, and
  Silicon sampling~\citep{argyle2023outofonemany}.
  A check (\yesmark) indicates explicit support in the published system;
  a cross (\nomark) indicates the capability is not stated. Dimensions read as
  follows: \emph{product-as-SUT evaluation} treats the product, not the agent,
  as the subject under test; \emph{persona conditioning} instantiates a user
  from a sampled population profile; \emph{behavioral grounding} verifies that
  behavior tracks a target attribute rather than confounders; \emph{native-device
  execution} covers real mobile/desktop apps beyond API demos;
  \emph{reproducible reporting} separates verifier facts from reporting policy;
  \emph{trajectory inspection} exposes per-trial traces; \emph{end-to-end
  validation} closes the persona--task--product loop; \emph{silicon sampling}
  substitutes simulated respondents for survey/UX pre-studies;
  \emph{reproducible cohorts} deterministically re-instantiate the same cohort;
  and \emph{multi-agent simulation} denotes persistent multi-agent social worlds
  (a deliberate non-goal for MatrAIx). The matrix compares capability coverage
  rather than providing an overall ranking.}
  \label{tab:system-comparison}
\end{table}


\section{Future Directions}
\label{app:future-directions}

Future work can extend both the evaluation framework and the persona models.
User-centered benchmarks could evaluate satisfaction, trust, retention, and
recovery from failure alongside task completion in interactive and long-horizon
settings. Persona~8B could also support automated generation of diverse,
realistic tasks and robustness tests in which users express the same goal in
different ways. Expanding the open-source Survey, AI Chatbot, Web, and App
environments would make these evaluations easier to reproduce and extend across
products and domains.

Methodological work could improve persona selection, fidelity, and consistency
across models and repeated runs; develop dynamic personas with long-term memory;
and compare prompting with steering, parameter-efficient adaptation, or
training-based simulation. Interpretability tools could help identify how
persona attributes affect model behavior. Broader applications include
user-simulation-based red teaming, user-controlled personalized assistants,
and studies in psychology, economics, policy, social science, multimodal
interaction, and embodied systems. These extensions require stronger
validation against longitudinal human data, especially before simulated
behavior is used for personalization, safety assessment, or decisions about
real populations.

%

\section{Responsible Use, Release, and Limitations}
\label{app:responsible-use}

MatrAIx personas are simulation instruments rather than people, and a cohort of
them is not a probability sample of a real population. The population conditions on attributes
including age, region, income, employment, and health, so it can be turned to
uses the release does not sanction: impersonating a named individual,
attributing opinions or behavior to a real person or an identifiable community,
assembling profiles that target individuals, or scripting persuasion, exclusion,
or price discrimination aimed at a protected group. None of these are supported
uses. Nor does running a simulated cohort discharge the obligation to consult
the people a product actually affects, particularly where a decision carries
consequences for them in health, finance, employment, or other regulated
settings. The volunteer instrument that grounds part of the population asks for
no name, contact detail, or account identifier and carries a decline option on
all 1,290 items. Work built on the release is expected to preserve that posture
rather than attempt re-identification. Human-referenced deployment studies remain the appropriate
standard for consequential claims, including evaluations of patient-facing,
EHR-integrated agents \citep{hao2025prostate}.

The public artifact is the Persona~1M coreset rather than the full internal
population. Records enter the coreset only after the filtering, provenance,
and deduplication checks of \secref{sec:persona-quality-control}. Its synthetic
component is calibrated toward four supported marginals, and the release ships
with a manifest, an audit of achieved shares and infeasible residuals, and
per-file hashes. Corpus size is distinct from execution scale: a record becomes
a persona agent only when paired with a model, interface, and task, and an
evaluation instantiates only its sampled cohort.

Every reported result is therefore a persona-agent result, not a direct claim
about how a person would behave. Results can also depend substantially on the
persona-agent model; for example, the paid-plan share on one product page spans
23.2\% to 93.9\% across three models on identical cohorts. MatrAIx runs should
therefore be treated as hypothesis-generating when the target claim concerns
people. Such claims require human validation on the same task and instrument,
following a design such as the 100-record study in
Appendix~\ref{app:extraction-quality}.

One configuration deserves its own warning, because it is the configuration a
user of this infrastructure is most likely to reach for. When the model playing
the persona and the model behind the system under test share a backbone, a
favorable result is ambiguous. Agreement may mean the system served the user
well, or it may mean a model recognized and preferred its own output, and the
run cannot tell those apart. The direction of the bias is not obviously benign
either: a persona may accept an answer a person would have pushed back on, which
inflates satisfaction and suppresses exactly the friction the evaluation exists
to surface.

The present experiments do not isolate that effect. They vary the persona
model across three frontier systems, but each task holds one application fixed,
so persona model and system backbone are never crossed and the self-preference
term cannot be separated from ordinary model-to-model variation. What the
results do establish is that the persona model is a first-order factor rather
than an implementation detail: the paid-plan share spans 23.2\% to 93.9\% across
three models on identical cohorts, and median pairwise Cohen's $\kappa$ across
88 joinable fields is approximately zero. An outcome that moves that much with
the acting persona model could also be sensitive to a shared backbone.

Two practices follow, and MatrAIx supports both. The agent model is recorded as
part of the evaluation configuration rather than left implicit, so a shared
backbone is visible in the report instead of hidden in it. And an evaluation
whose conclusion matters should also run the same cohort under at least one persona
model that does not share a backbone with the system under test, treating
agreement between matched models as a hypothesis to check rather than a result
to report. Isolating the effect properly requires crossing persona model with
system model on one task, which we have not run and consider the natural next
experiment.

A second gap is one of scope rather than configuration. \secref{sec:related-work}
argues that support-style simulation requires realistic disclosure, correction,
refusal, and abandonment. The present experiments do not directly measure these
behaviors.
The adherence probe asks whether a declared style appears in a trajectory, which
is a question about conditioning, not about whether the resulting user resembles
a real one. Nothing here establishes that a MatrAIx persona withholds context,
pushes back, or gives up the way people do.

Two designs could address this gap using logs that already exist
\citep{zhao2024wildchat,zheng2024lmsyschat,kirk2024prism,paruchuri2025s} and
telemetry MatrAIx already stores. At the population level, simulated user turns
can be compared against matched slices of real logs on turn length, question
type, volunteered versus withheld context, correction and abandonment rates, and
response-option entropy. A real-versus-simulated classifier can summarize these
differences, with an AUC near $0.5$ as the target. At the individual level, a persona can be extracted
from the first half of a held-out real conversation, the next user turns
simulated, and the result scored against what the person actually wrote, with
shuffled-persona and no-persona baselines separating persona information from
conversational momentum. The second is the behavior-chain test of
\citet{li2025far} applied to in-the-wild data rather than to a constructed
benchmark. Both would have to be reported per agent model given the model
dependence in \secref{sec:validation}, and public logs would need screening for
memorization, since a model may have seen the very conversation it is asked to
continue. We consider this a priority for future validation.

\end{document}